\documentclass[twoside]{article}

\usepackage[accepted]{aistats2022}
\usepackage[round]{natbib}

\usepackage{import}
\usepackage{microtype}
\usepackage{graphicx}
\usepackage{xspace} % need it for the etal/etc/eg macros. CVPR includes, ICML 
\usepackage{amsmath,amssymb,amsfonts}
\usepackage{etoolbox}

\usepackage[thmmarks, amsmath, thref]{ntheorem}
\usepackage{bm}
\usepackage{bbold}
\usepackage{mathtools}

\usepackage[normalem]{ulem} % for \sout w/o the annoying underline in \emph

\usepackage{epsfig}
\usepackage{graphicx}
\usepackage{subcaption}
\usepackage{booktabs}
\usepackage{algorithm}
\usepackage[noend]{algpseudocode}
\usepackage[backref, colorlinks,citecolor=blue]{hyperref}
\hypersetup{
  colorlinks   = true, %bColours links instead of ugly boxes
  urlcolor     = blue, %Colour for external hyperlinks
  linkcolor    = blue, %Colour of internal links
  citecolor   = blue, %Colour of citations
}

\usepackage{diagbox}

\usepackage{lipsum}  

\usepackage{threeparttable}
\usepackage[singlelinecheck=true,justification=centering]{caption}
\usepackage{makecell}
\usepackage{multirow}
\usepackage{color,soul}
\usepackage{xcolor}  

\definecolor{dimgray}{rgb}{0.35, 0.35, 0.35}

\subimport{./}{macros}
\theoremstyle{definition}  
\newtheorem{Remark}{Remark}

\theoremstyle{definition}

\newcommand{\footnotestandalone}[1]{\let\thefootnote\relax\footnotetext{#1}}
\begin{document}

% % START  (TEMP)
% \pagenumbering{gobble}
% \tableofcontents
% \clearpage
% \pagenumbering{arabic}
% %  END 

\runningtitle{Common Failure Modes of Subcluster-based Sampling in DPGMMs -- and a DL Solution}

\runningauthor{Winter*, Dinari*, and Freifeld $\qquad$ (* = equal contribution)}

\twocolumn[

% \aistatstitle{Characterization of SubC sampler failure modes and a deep learning solution}
\aistatstitle{Common Failure Modes of Subcluster-based Sampling in Dirichlet Process Gaussian Mixture Models -- and a Deep-learning Solution}

\aistatsauthor{Vlad Winter* \And Or Dinari* \And Oren Freifeld
\newline }
% \aistatsaddress{\And The Department of Computer Science, Ben-Gurion University, Israel \And  }  ]
\aistatsaddress{ winterv@post.bgu.ac.il \\ Ben-Gurion University 
\And dinari@post.bgu.ac.il\\ Ben-Gurion University 
\And  orenfr@cs.bgu.ac.il \\ Ben-Gurion University } ]

% \subimport*{./}{abstract.tex}

\begin{abstract}
    The Dirichlet Process Gaussian Mixture Model (DPGMM) is often used to cluster data when the number of clusters is unknown.  One main DPGMM inference paradigm relies on sampling. Here we consider a known state-of-art sampler (proposed by
Chang and Fisher III (2013) and improved by~Dinari \etal (2019)), analyze its failure modes, and show how to improve it, often drastically. Concretely, in that sampler, whenever a new cluster is formed it is augmented with two subclusters whose labels are initialized at random. Upon their evolution, the subclusters serve to propose a split of the parent cluster. We show that the random initialization is often problematic and hurts the otherwise-effective sampler. Specifically, we demonstrate that this initialization tends to lead to poor split proposals and/or too many iterations before a desired split is accepted. This slows convergence and can damage the clustering. As a remedy, we propose two drop-in-replacement options for the subcluster-initialization subroutine. The first is an intuitive heuristic while the second is based on deep learning. We show that the proposed approach yields better splits, which in turn translate to substantial improvements in performance, results, and stability.
Our code is publicly available. 

\end{abstract}
\section{INTRODUCTION}\label{Sec:Intro}
\begin{figure}[ht!]
  \subcaptionbox{\vspace{-.1cm}\label{fig:intro:theirs}}[0.9\linewidth]{\includegraphics[width=0.80\linewidth,trim=1.1cm 0.9cm 0.3cm 0.2cm, clip]{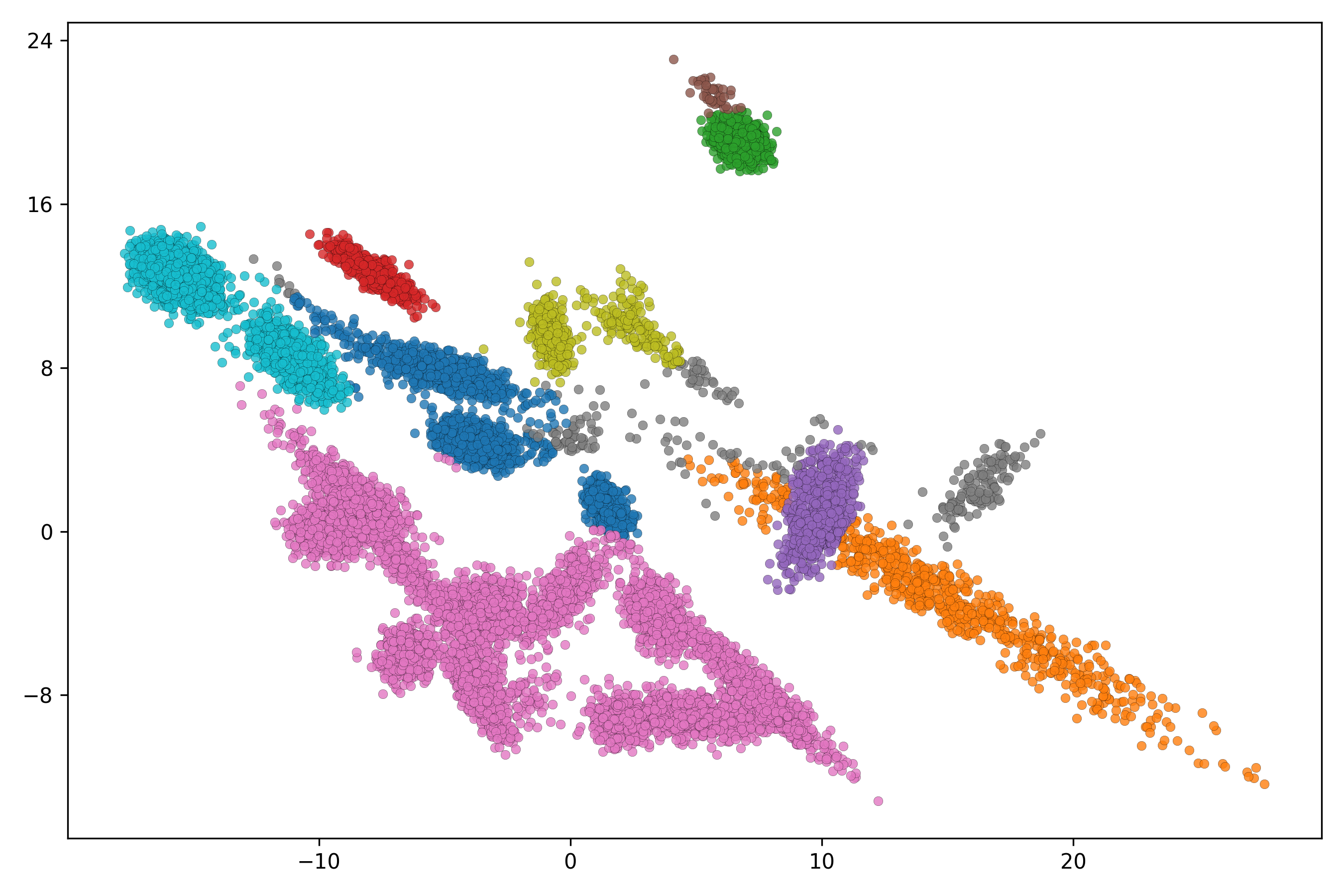}} \\ 
  \subcaptionbox{\vspace{-.1cm}\label{fig:intro:ours}}[0.9\linewidth]{\includegraphics[width=0.80\linewidth,trim=1.1cm 0.9cm 0.3cm 0.2cm, clip]{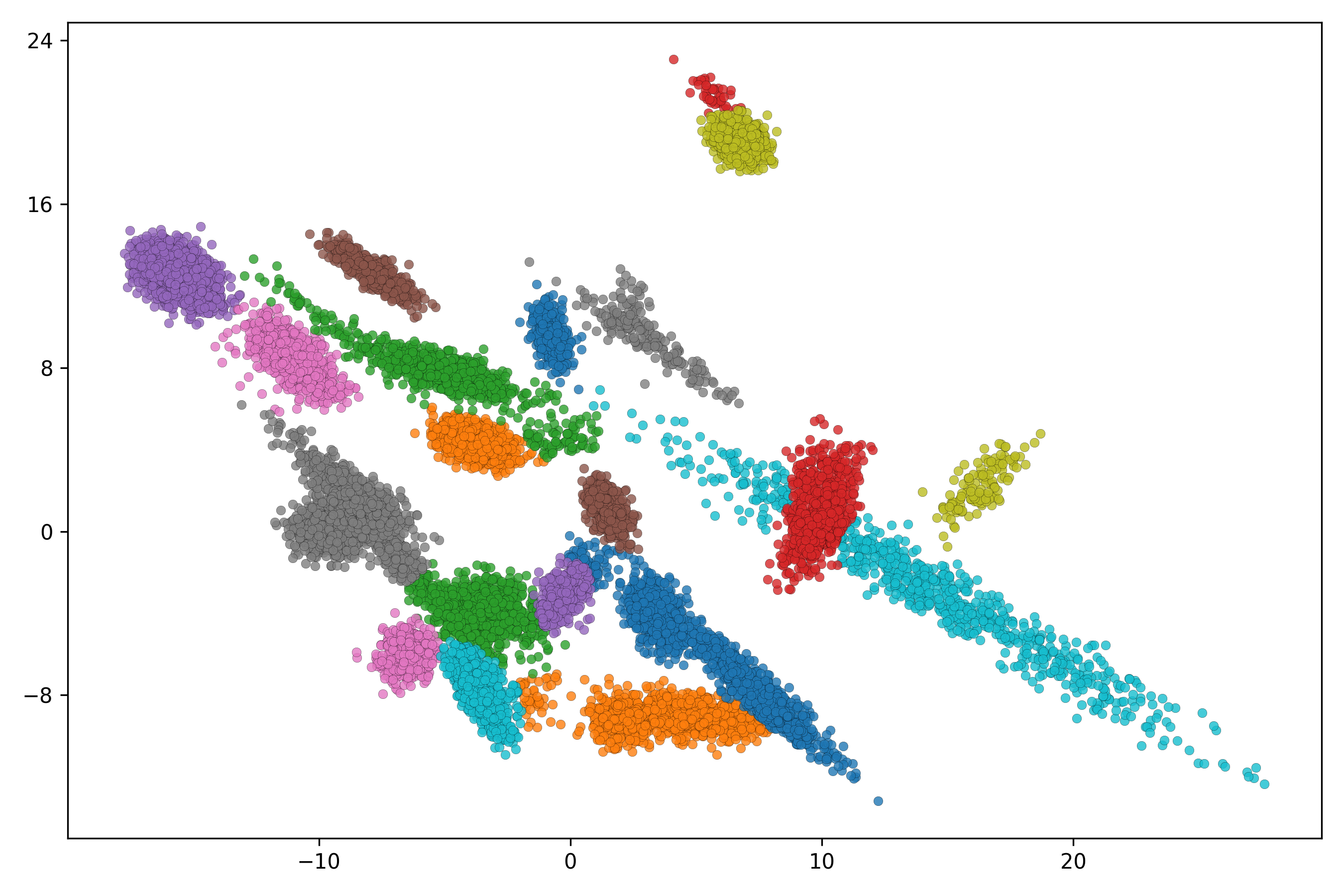}}
  % \captionsetup{aboveskip=0pt}
  % \subcaptionbox{\vspace{-.1cm}\label{fig:intro:theirs}}[0.9\linewidth]{\includegraphics[width=0.8\linewidth]{../figures/intro_random.png}} \\ 
  % \vspace{-1.5\baselineskip}
  % \subcaptionbox{\vspace{-.1cm}\label{fig:intro:ours}}[0.9\linewidth]{\includegraphics[width=0.8\linewidth]{../figures/intro_splitnet.png}}
  % \captionsetup{justification=justified, singlelinecheck=false, aboveskip=0pt}
\vspace{-0.8\baselineskip}
\captionsetup{justification=justified, singlelinecheck=false}
% \caption[Random Initialization vs. Smart Initialization]{ \TODO{Rewrite}
%     \textbf{Top}: a post-convergence result of the DPGMM sampler from~\cite{Chang:NIPS:2013:ParallelSamplerDP} 
%     % Using their random initializations of the subclusters of each newly-formed cluster 
%     in a difficult dataset of many partially-overlapping clusters.
%     Their initializations make the sampler propose bad splits that are mostly rejected while most of the desired splits are never proposed.
%     Thus, the sampler gets stuck in a clearly-sub-optimal point, erroneously labeling many small clusters as one big cluster. 
%     \textbf{Bottom}: powered with our proposed SplitNet initializations, the same sampler proposes much better splits which let it quickly converge to the correct clusters. 
% } 
\caption[Random Initialization vs. Smart Initialization]{(a) A post-convergence clustering result of the SubC sampler~\citep{Chang:NIPS:2013:ParallelSamplerDP} 
    % Using their random initializations of the subclusters of each newly-formed cluster 
    in a relatively-difficult dataset. Note the significant underestimation of the $\#$ of clusters. 
%     Their initializations make the sampler propose bad splits that are mostly rejected while most of the desired splits are never proposed.
%     Thus, the sampler gets stuck in a clearly-sub-optimal point, erroneously labeling many small clusters as one big cluster. 
    (b) With our proposed SplitNet for subcluster initializations, the same sampler quickly converges to a much better result. 
}
\label{fig:intro}
\end{figure}

The Dirichlet Process Gaussian Mixture Model (DPGMM), a   
Bayesian Nonparametric (BNP) extension 
 of the Gaussian Mixture Model (GMM), provides a flexible and principled approach to clustering when $K$, the number of clusters, is unknown. One appeal of BNP clustering is that, as $K$ itself is inferred, the models adapt to the complexity of the data.
 \footnotestandalone{\small {\textbf{Acknowledgements.}
This work was supported by the Lynn and William
Frankel Center at BGU CS, by
the Israeli Council for Higher Education via the BGU Data Science Research Center, and by Israel Science Foundation Personal Grant \#360/21. O.D.~was also funded 
by the Jabotinsky Scholarship from Israel's Ministry of Technology and Science, and by BGU's Hi-Tech Scholarship.}
}

Exact DPGMM inference being infeasible, the main inference paradigms are the variational approach and the Markov Chain Monte Carlo (MCMC,~\cite{Robert:Book:2013:Monte}) sampling-based approach. 
\cite{Chang:NIPS:2013:ParallelSamplerDP} proposed a sampler, henceforth referred to as the SubC sampler
(due to is usage of~\emph{subclusters}; see~\autoref{Sec:Background}) which is one of the best and fastest DPGMM samplers (especially via its recent and more scalable reimplementation by~\cite{Dinari:CCGRID:2019:distributed}).
A key component of the SubC sampler is the splitting and merging of clusters, which allows changing $K$. For reasons to become clear shortly, our paper focuses on their splits. During the sampling iterations, whenever a new cluster is formed it is augmented with two subclusters. Next, in the subsequent iterations, these subclusters evolve with the rest of the model where, every once in a while, the sampler \emph{proposes} to split the cluster into its two subclusters. 
Note well: \emph{subclusters are created not only before running the sampler as part of the overall initialization but also many times during the inference process itself}; \ie, whenever a cluster is split (during the sampler's run), its subclusters become clusters, and each of which must be augmented with their own pair of new subclusters. 

A practical question arises: upon their creation, how should the subsclusters be initialized? \cite{Chang:NIPS:2013:ParallelSamplerDP} did so by randomly partitioning the cluster into its subclusters. Seemingly, this is a natural and benign choice.  Another question, perhaps less obvious, then follows: \emph{how much, if at all, does the choice of the initialization method matter}?

As we will show, \emph{the surprising answer is that it can matter a lot and have drastic effects} on the SubC sampler's  performance (\eg, it can be the difference between poor clustering~(\autoref{fig:intro:theirs}) and a success~(\autoref{fig:intro:ours})), convergence speed, and stability.  

We note here that the crust of the matter is that the random initializations tend to lead to poor split proposals (which are mostly rejected) and/or too many iterations before a desired split is accepted.
Upon discovering that the subcluster initializations play such a crucial role, and upon characterizing the failure modes caused by the random initialization, we set out to investigate better
subcluster-initialization methods. 
Of note, since subcluster initialization often occurs multiple times during the sampler's run, such methods must be fast (prohibiting the usage of relatively-expensive clustering methods). 
This leads us to propose two drop-in-replacement alternatives for subcluster initialization: 1) a simple-yet-effective heuristic based on 2-means (\ie, the classical $K$-means~\citep{Lloyd:PCM:1982:kmeans} whose ``$K$'' is 2 and should not be confused with $K$, that in this paper denotes the total number of inferred clusters according to the model);
2) an even better alternative, a new Deep Neural Net, self-coined ``SplitNet''. 
The utility of those solutions, SplitNet in particular, is especially noticeable in the context of the aforementioned failure modes.  
In our experimental study, we compare the baseline SubC sampler, its two variants 
based on our proposed solutions, as well as several other popular DPGMM methods, and test them all using a comprehensive suite of synthetic simulations and real datasets. We show that with our better initializations, the sampler outperforms the other methods in inference time, stability, and various clustering evaluation metrics.

To summarize, our two main contributions are: 
1) we characterize failure modes of the SubC sampler~\citep{Chang:NIPS:2013:ParallelSamplerDP};
2) we propose two subcluster initialization methods that improve that sampler, the important among the two being a novel Deep Learning (DL) method. 
Finally, our code is publicly available at \url{https://github.com/BGU-CS-VIL/dpgmm_splitnet}.

\section{RELATED WORK}\label{Sec:RelatedWork}
Bayesian non-parametric (BNP) mixture models are commonly used in unsupervised tasks such as clustering, topic modeling, and density estimation~\citep{Muller:Book:2015:BNP}. 
A common formulation is the Dirichlet Process Mixture Model (DPMM)~\citep{Ferguson:AoS:1973:Bayesian,Antoniak:AoS:1974:DPMM}, exemplified by the DPGMM. 
% However, exact inference in DPMM is often intractable, and so, as it is usually the case with Bayesian inference, there are two main approaches to allow inference: a sampling-based approach with Monte Carlo Markov Chain (MCMC) and a variational approach. On top of that, the distribution of the computations across CPU cores and/or machines is another avenue that has been explored.

\textbf{Sequential and Parallel Samplers for DPMM.}
Earlier MCMC methods for DPMM inference were primarily sequential. Two of the chief examples are the Collapsed Gibbs Sampler~\citep{Neal:JCGS:2000:markov} and a sampler based on the Chinese Restaurant Process~\citep{aldous1985exchangeability}.
See also~\cite{walker2007sampling} for a technique based on slice sampling~\citep{Damlen:JRSS:1999:Gibbs}. 
Due to the slowness of the sequential samplers, and as an alternative, in recent years several parallel samplers have been proposed. For example,
\cite{papaspiliopoulos2008retrospective} relied on retrospective sampling while~\cite{Williamson:ICML:2013:parallel} proposed a parallel sampler which treats the DPMM as a mixture of DPMMs, allowing to divide the workload between several processes.
The SubC sampler~\citep{Chang:NIPS:2013:ParallelSamplerDP} is another type of a parallel Gibbs sampler, based on a split/merge MCMC framework (the latter was first introduced by~\cite{Jain:Jounral:2004:split}). The SubC sampler, which will be discussed in more depth in~\autoref{Sec:Background}, alternates between a restricted Gibbs sampler, which handles a fixed amount of components, and the split/merge framework.
The latter can modify the number of instantiated components. That work was accompanied with a single-machine, multithreaded implementation (in MATLAB/C++) which uses a shared memory model. For a discussion on parallel samplers and possible associated  pitfalls, see~\cite{Gal:ICML:2014:Pitfalls}. 

\textbf{Distributed computations in DPMM samplers.}
% \RED{DO WE WANT TO SAY ANYTHING ABOUT THIS?}
%
In addition to parallelism, efforts have been made to distribute computations (\eg:~\cite{Ge:ICML:2015:distributed,Wang:JCAI:2017:Scalable}). Of particular interest to us is~\cite{Dinari:CCGRID:2019:distributed}, a more scalable and faster implementation
%  (in Julia, with an optional Python wrapper) 
of the SubC sampler~\citep{Chang:NIPS:2013:ParallelSamplerDP}. It is based on a distributed memory model and supports not only multiple cores but also multiple machines. In fact, it turns out that 
even when both the original and the new implementations of the SubC sampler are run
on a \emph{single} multi-core machine, the one from~\cite{Dinari:CCGRID:2019:distributed} is faster.
Thus, we use it as our baseline. 

% Examples for distributed DPMM samplers include~\cite{Ge:ICML:2015:distributed} that uses a map-reduce scheme, and the sampler from~\cite{Wang:JCAI:2017:Scalable} which maintains consistency via a consolidation scheme. Another example was proposed by~\cite{dinari2019distributed} who expanded the sampler from~\cite{Chang:NIPS:2013:ParallelSamplerDP} by proposing a scalable distributed implementation (in Julia) which uses a distributed-memory model via an explicit partitioning of the data. This facilitated the use of Chang and Fisher’s DPMM sampler to high dimensions and large-scale datasets

\textbf{Variational inference for the DPMM.}
While we focus on sampling, an important alternative is variational inference, first proposed in the DPMM context by~\cite{Blei:Journal:2006:variational}. That approach was also explored for DPMMs by others, \eg,~\cite{Hoffman:JMLR:2013:sovb,Hughes:NIPS:2013:memoizedDP,Kurihara:NIPS:2007:accelerated,Wang:AISTATS:2011:online,Wang:NIPS:2012:truncation,Bryant:NIPS:2012truly,fei2019dirichlet,huynh2016streaming}.

\section{BACKGROUND}\label{Sec:Background}
Let $D$ be the dimensionality of the data. 
One known DPGMM construction (out of several) is based on the following sampling procedure, where we also assume a Normal Inverse Wishart (NIW) prior: 
\begin{align}
    & \bpi|\alpha \sim \mathrm{GEM}(\bpi;\alpha),  
    \label{Eqn:DPLatentVar:pi}
    \\
    & \theta_k |\lambda  \overset{\iid}{\sim} \mathrm{NIW}(\theta_k;\lambda), \qquad \forall k\in \set{1,2,\ldots},\\
    & z_i|\bpi \overset{\iid}{\sim} \mathrm{Cat}(z_i;\bpi), \qquad \forall i\in \set{1,2,\ldots,N}, \\
    & \bx_i|z_i,\theta_{z_i} \sim \Ncal(\bx_i;\theta_{z_i}), \qquad\forall i\in \set{1,2,\ldots,N}\ .
\label{Eqn:DPLatentVar:xi}
\end{align}
Here $\bpi=(\pi_k)_{k=1}^\infty$ is a weight vector of infinite length (\ie, $\pi_k>0$ for every $k$ and $\sum_{k=1}^{\infty}\pi_k=1$) drawn from the Griffiths-Engen-McCloskey stick-breaking process (GEM)~\citep{Pitman:Book:2002:Combinatorial} with a concentration parameter $\alpha>0$ 
while $\theta_k = (\bmu_k, \bSigma_k)$ stands for the mean vector and covariance matrix of Gaussian $k$ (so $\bmu_k \in \RD$ and $\bSigma_k$ is an $D$-by-$D$ symmetric positive-definite matrix) sampled from an NIW distribution whose probability density function (pdf) is denoted by $\mathrm{NIW}(\cdot;\lambda)$ where $\lambda$ represent the hyperparameters of the prior. 
Each of the $N$ \iid  observations $(\bx_i)_{i=1}^N\subset \RD$ is generated 
by first drawing a label, $z_i\in \Zplus$, from $\bpi$ (\ie, $ \mathrm{Cat}(\cdot;\bpi)$ is
the categorical distribution), and then drawing $\bx_i$ from 
Gaussian ${z_i}$ where $\Ncal(\cdot;\theta_{z_i})$ is a $D$-dimensional Gaussian pdf
parameterized by $\theta_{z_i}=(\bmu_{z_i},\bSigma_{z_i})$. 
Note that the latent variables in Equations~\eqref{Eqn:DPLatentVar:pi}--\eqref{Eqn:DPLatentVar:xi} above are $(\theta_k)_{k=1}^\infty$, $\bpi$, and $(z_i)_{k=1}^N$. For more details and alternative constructions, 
see~\cite{Sudderth:PhD:2006:GraphicalModels}.

\subsection{The SubC Sampler}
Due to space limits, and since most of the many details of the fairly-elaborated SubC sampler~\citep{Chang:NIPS:2013:ParallelSamplerDP} are inessential for understanding our work, their entire algorithm 
(in the context of the DPGMM) appears in our appendix. 
Below we provide only a high-level review while also focusing on the part most relevant to us: the splits. 
The SubC sampler consists of a \emph{restricted} Gibbs sampler and a split/merge framework, which together form an ergodic Markov chain. The operations in each step of the sampler are highly parallelizable while the splits/merges enable the sampler to perform \emph{large moves} (on the optimization surface). 

\begin{figure}[h!]
\centering
  \includegraphics[width=0.9\linewidth,trim={2mm 4mm 6mm 2mm},clip]{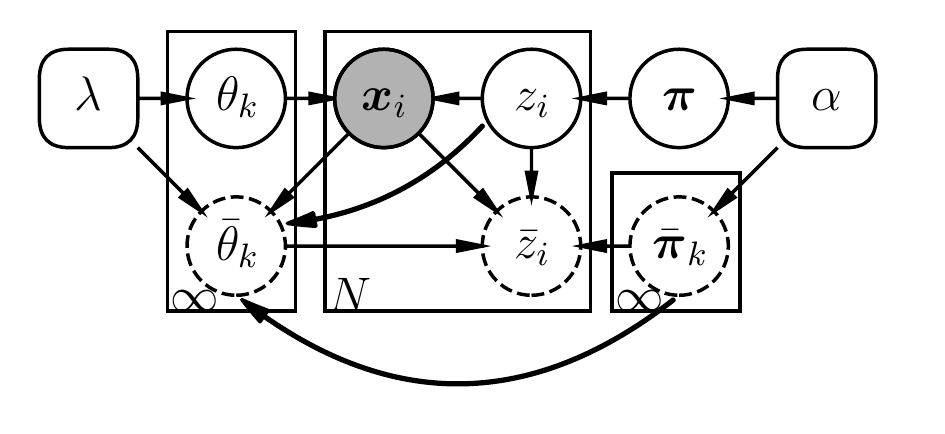}
  \caption{The DPGMM with the augmented space
  depicted as a graphical model. This figure is based on the one in~\cite{Chang:Thesis:2014sampling}.}
  \label{fig:dpgmm:aug}
\end{figure}

\textbf{The augmented space.} 
The latent variables, $(\theta_k)_{k=1}^\infty$, $\bpi$, and $(z_i)_{k=1}^N$, are augmented with auxiliary variables as follows. To each $z_i$, an additional \emph{subcluster label}, $\bar{z}_i\in\set{l,r}$, is added, where ``$l$'' and ``$r$'' conceptually stand for ``left'' and ``right'', respectively. 
To each component $\theta_k$ two subcomponents are added, $\bar{\theta}_{k,l},\bar{\theta}_{k,r}$, with weights $\bar{\bpi}_k$. See~\autoref{fig:dpgmm:aug} for a graphical model of the DPGMM with the augmented space.

\textbf{The restricted Gibbs sampler.} 
This restricted sampler is not allowed to change $K$, the current number of estimated clusters; rather, it can update only the parameters of the existing clusters, and, when sampling the labels, it can assign an observation only to an existing cluster. 
During the iterations of the restricted sampler, the subclusters evolve together with the clusters. Concretely, in each pair of subclusters the latter evolve using Gibbs sampling in a 2-component GMM (except that each iteration is also conditioned on the cluster label).

\textbf{The split/merge framework.} 
Splits and merges allow the sampler to change the current number of instantiated components, using the Metropolis-Hastings framework~\citep{Hastings:1970:MC}. 
Let us focus here on the splits. Every certain (user-defined) amount of iterations, the sampler proposes splitting an existing cluster into its subclusters.
The split of cluster $k$ is accepted with probability $\text{min}(1,H)$ where
\begin{align}
\hspace{-.2cm}        H = \frac{\alpha \Gamma(N_{k,l}) f_\bx(\bx_{\Ical_{k,l}};\lambda)  \Gamma(N_{k,r}) f_\bx(\bx_{\Ical_{k,r}};\lambda)}{\Gamma(N_k) f_\bx(\bx_{\Ical_k};\lambda)} 
        \label{eqn:HastingRatioSplit}
\end{align}
is the Hastings ratio, $\Gamma$ is the Gamma function, $N_{k,l}$
and $N_{k,r}$ are the number of points in subclusters $l$ or $r$, respectively, 
$\bx_{\Ical_k}$ denotes the points in cluster $k$,  $\bx_{\Ical_{k,s}}\subset \bx_{\Ical_k}$
denotes the points in subcluster $s\in\set{l,r}$,
and $f_\bx(\cdot;\lambda)$ is the \emph{marginal} data likelihood (see the appendix for the concrete expression). 
Upon a split acceptance, each of the new clusters is augmented with two subclusters (that must be initialized somehow). 

\section{METHOD}
\begin{figure}[t]
     \centering        
      % \subcaptionbox{
        % A single cluster (blue) that needs to be split
        % \label{fig:clusters_random_init.original}}
      % [.95\linewidth]{\includegraphics[width=0.9\linewidth]{../figures/two_clusters.pdf}}\\
      % \subcaptionbox{
        %  A random initialization of the labels (shown in 2 colors)
                % \label{fig:clusters_random_init.random}
  % [.95\linewidth]  
  \includegraphics[width=0.7\linewidth]{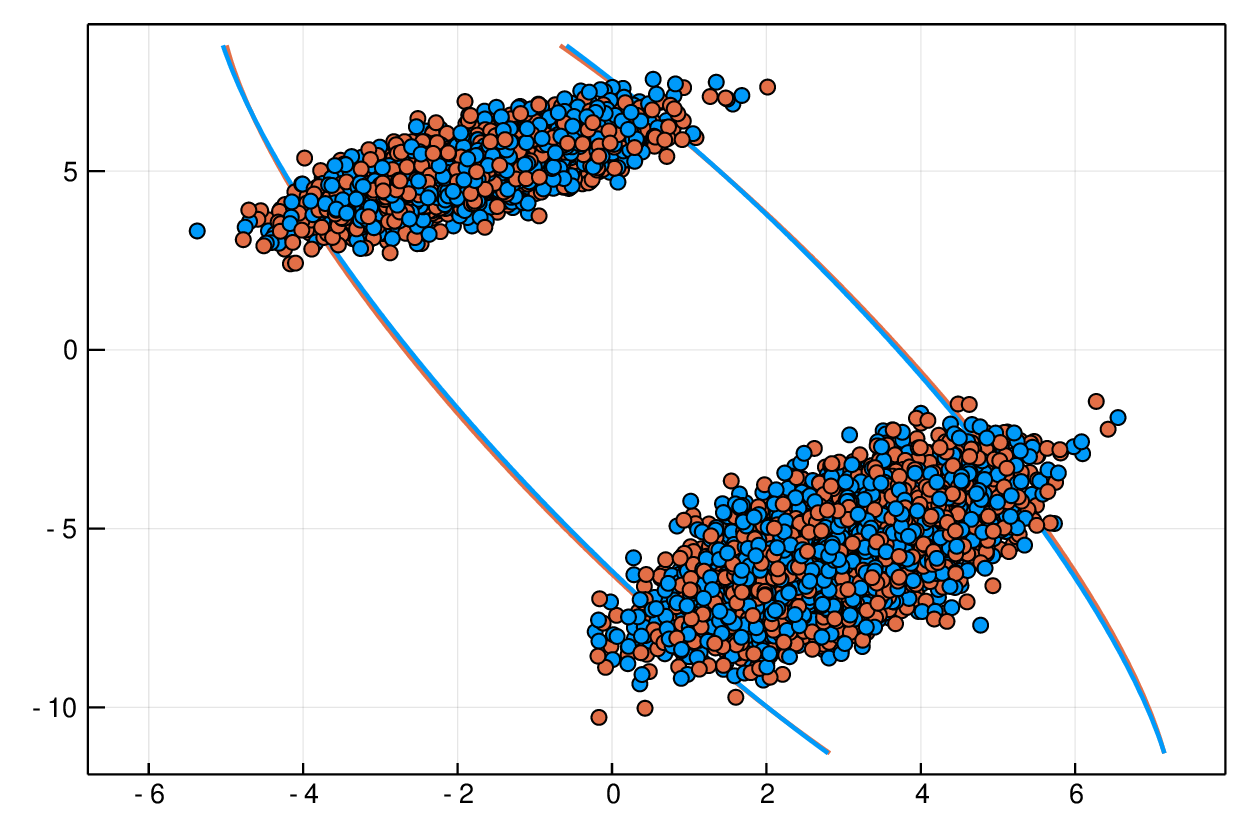}
  \vspace{-.3cm}
  \captionsetup{justification=justified, singlelinecheck=false}
  \caption[Random Initialization Example]{A random initialization~\citep{Chang:NIPS:2013:ParallelSamplerDP} yields 2 almost-identical Gaussians (visualized by the almost-identical orange and blue ellipses). At this state, the proposed split will be rejected with a very high probability (\EQN\eqref{eqn:HastingRatioSplit}). It will take at least several iterations before the 2-component GMM will imply a split that is likely to be accepted. }
\label{fig:clusters_random_init}
\end{figure}

Let us start with the failure modes of the SubC sampler.
In~\cite{Chang:NIPS:2013:ParallelSamplerDP}, the subcluster initialization is done by randomly partitioning the cluster of interest into two subclusters. 
\autoref{fig:clusters_random_init} shows an example. Although it is evident
that there are, in fact, two distinct clusters, the random initialization
 of course gives rise to initial subclusters (indicated in the figure in different colors)
 that are nearly identical (in terms of their estimated 2-component GMM parameters). While simple, this incurs a heavy price in the subsequent iterations of the algorithm, as the restricted Gibbs sampler will relatively-slowly move points between the two subclusters until a sufficiently-good split proposal is reached; \ie, until  the resulting 2-component GMM will give rise to a high Hastings Ratio (\EQN\eqref{eqn:HastingRatioSplit}).

\begin{figure*}[h]
\centering
  \subcaptionbox{Inferred $K$ vs. \# GT Clusters \label{fig:subc_failure:K}}
  {\includegraphics[width=0.32\linewidth]{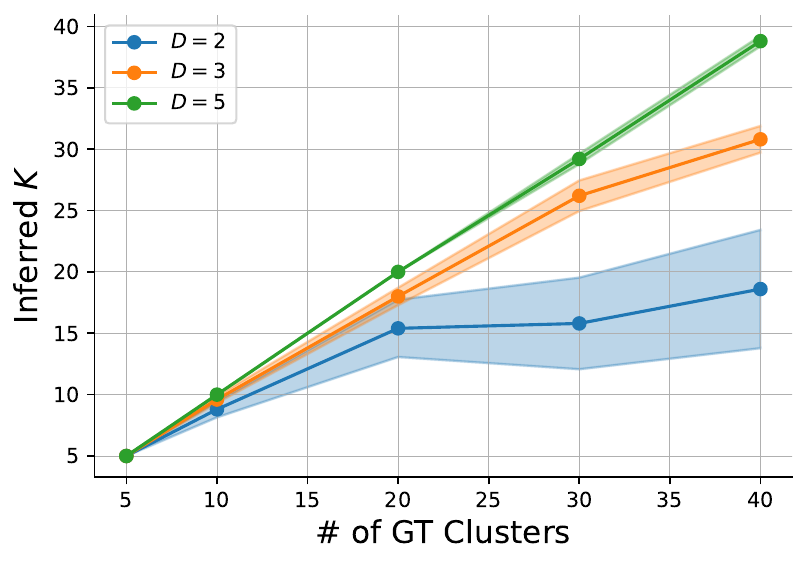}} 
  \subcaptionbox{NMI vs. \# GT Clusters \label{fig:subc_failure:NMI}}
  {\includegraphics[width=0.32\linewidth]{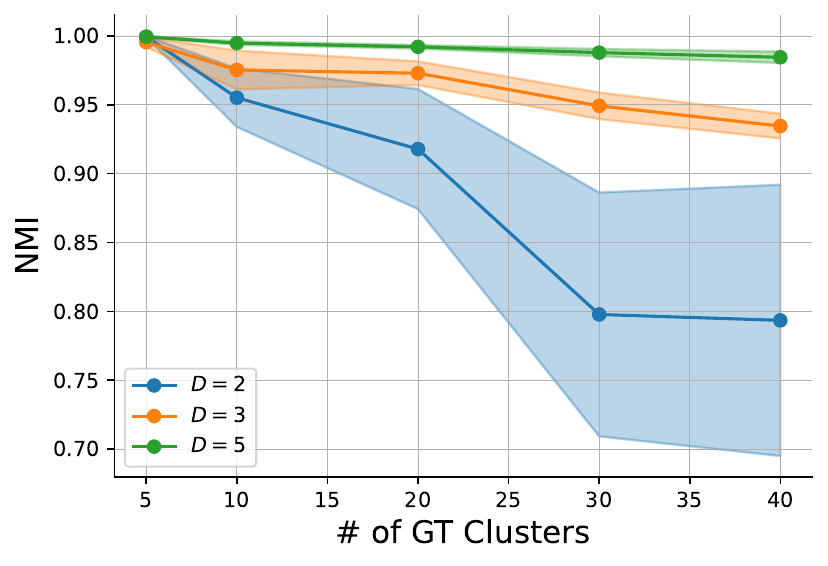}}
  \subcaptionbox{ARI vs. \# GT Clusters \label{fig:subc_failure:ARI}}
  {\includegraphics[width=0.32\linewidth]{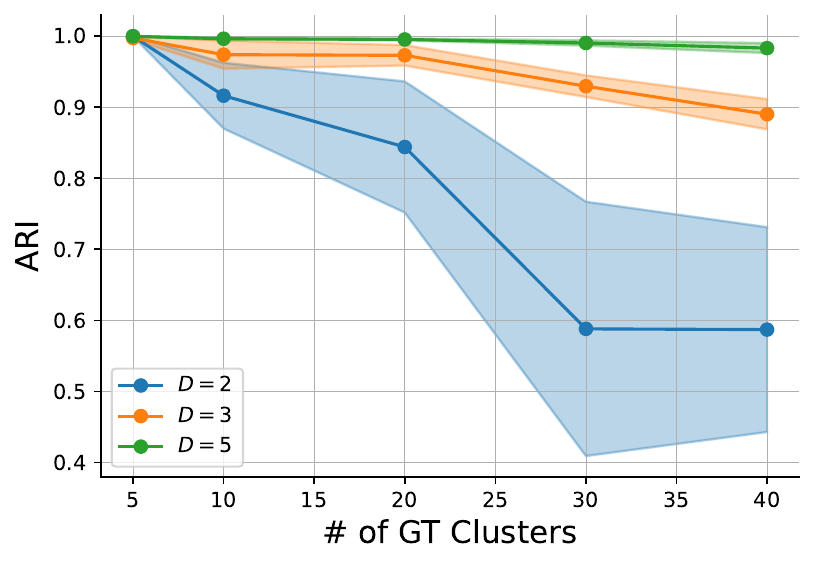}}  
\caption{Deterioration analysis of the SubC sampler~\citep{Chang:NIPS:2013:ParallelSamplerDP}. Results were obtained by running the SubC sampler on increasingly-denser synthetic GMM datasets.
%  (10 random datasets per configuration, displaying mean and std. dev.).
  Note that it is the \emph{lower} dimensions where the sampler struggles the most (as there it is harder to discover that a cluster needs to be split).} 
\label{fig:subc_failure}
\end{figure*}

Moreover, and as we will show, slow convergence is not the only problem with random initializations. 
In~\autoref{fig:clusters_random_init} we saw that even when clearly-distinct clusters exist, the two randomly-initialized subclusters are almost indistinguishable.  Eventually, the restricted Gibbs sampler will give rise to the correct subclusters, but it will take time to get there.
A more severe problem is shown in~\autoref{fig:intro:theirs}: on this challenging synthetic dataset, using random subcluster initializations the SubC sampler failed to achieve good clustering results, despite having converged; \ie, it missed many desired splits, and by converging to a poor local maximum, grossly underestimated $K$.
As demonstrated in \autoref{fig:subc_failure}, this phenomenon notably occurs in challenging datasets characterized by densely-packed and partially-overlapping clusters and \emph{especially in lower dimensions} 
as there it is often hard to distinguish between clusters. 

While random initialization is often adequate for clustering algorithms in general, we argue that it is a poor choice in the context of the SubC sampler. Recall that the whole point of maintaining the subcluster mixtures is to facilitate desired splits. It follows that this goal must be taken into consideration when initializing the subclusters. \emph{We suspect this insight was missed because the original sampler~\citep{Chang:NIPS:2013:ParallelSamplerDP} was already, overall,  highly effective, producing at-the-time state-of-the-art results.}  

\subsection{The Proposed Solutions}\label{Sec:Method}
In this section we propose two alternative subcluster initialization methods. 
The rationale behind proposing ``smart splits'' is based on the sampler's strong reliance on $H$ (\EQN\eqref{eqn:HastingRatioSplit}): a split proposal is accepted mostly when the (marginal) likelihood according to the two subclusters sufficiently improves upon the (marginal) likelihood according to the original cluster whose possible split is under consideration. Particularly, when the subclusters are almost indistinguishable (\eg, as in~\autoref{fig:clusters_random_init}), $H$ tends to be low. When the subclusters are initialized randomly, then, by construction, the region ``covered'' by one subcluster is roughly the same as the other. Thus, \emph{fast convergence} from such an initial state into a state where the two subclusters cover sufficiently-different regions to justify a split is unlikely. 
Given the above, it is preferable to initialize the subclusters in a way such that the covered regions are sufficiently distinct, resulting in an immediately higher $H$ or at least one that will be achieved within a few iterations. 
Importantly, the initialization scheme, as it is called many times during the sampler's run, cannot be too slow or computationally expensive. 
Let us first consider a simple solution. 

\subsubsection{Subcluster initialization via $K$-means.} 
The first solution we propose is based on a simple heuristic: 
create the subclusters by running on the cluster the $K$-means algorithm, with $K=2$.
Other classical clustering algorithms such as $K$-medoids, EM-GMM, \etc. are also optional, 
but this option is simple, intuitive and computationally cheap.  
Also note that one obvious advantage of $K$-means (over a random initialization) is that it yields two \textbf{non-overlapping} subclusters. 
Another related advantage, verified empirically, is that usually, in situations where the cluster needs to be split but the initial Hastings Ratio is still too low, 
a SubC sampler that uses $K$-means subcluster iterations requires fewer iterations, until the ratio is high enough.
In practice, $K$-means is fast, adding a relatively-insignificant computational overhead. However, as data becomes more densely packed and of higher dimension, the effectiveness of $K$-means initialization will decrease. %Indeed, we validate this in~\autoref{Sec:Results}.

\subsubsection{Subcluster initialization via SplitNet}\label{Sec:Method:Subsec:SplitNet}
Our second proposed solution is based on Deep-Learning (DL). The latter is attractive here as its feedforward pass is fast, and since it excels in learning 
complex data patterns.
Specifically, given a cluster we want the deep net to predict useful subclusters. 
Below we explain the learning-task formulation, the choice of architecture, loss function, data generation, and training procedure. Importantly, no manual labeling of data is needed.

The first step is to formulate the task at hand in terms suitable for DL. 
Intuitively, we would like our so-called SplitNet model, denoted by $f$, to learn to distinguish between the two
regions of the highest density. 
Let $X$ be a cluster of $N$ points in $\RD$. 
SplitNet's \textbf{input} is $X$, represented as a matrix, $X\in \RR^{N\times D}$, where each row corresponds to a a different point: $X = [\bx_1, \ldots , \bx_n]^T$.
Its \textbf{output} (thought of as subcluster assignments) is 
a binary-valued $N$-length column vector: \ie, 
$\widehat{Z}= (\widehat{z}_i)_{i=1}^N
= \MATRIX{\widehat{z}_1,\ldots,\widehat{z}_N}$
where $\widehat{z}_i \in\set{0,1}$.

Some inherent characteristics of the input and output must be considered when formulating the neural model and its accompanying training procedure.
\textit{First}, the loss should be invariant to a permutation of the rows of $X$
and $\widehat{Z}$, provided the \emph{same} permutation applies to both. 
In other words, we should consider the input/output as a set, $\set{(x_i,\hat{z_i})}_{i=1}^N$. 
\textit{Second}, we cannot know in advance the number of points,  $N$.
\textit{Third}, the label ordering (\ie, $(0,1)$ or $(1,0)$)  is arbitrary and interchangeable;
what is essential here is only the partition itself, not the ``names'' of the labels.
The loss for predicting $\widehat{Z}$ should be the same if we replace \emph{all} the $(\widehat{z}_i)_{i=1}^N$ with their 1-complement; namely, $(1-\widehat{z}_i)_{i=1}^N$.  

Let us summarize these considerations: 
1) The model must be able to work with any number of points at the input (varying $N$);
2) The model needs to be invariant to permutations of the rows of $X$ (as $\set{(x_i,\hat{z_i})}_{i=1}^N$ is a set);
3) The model's output (subcluster assignments) must be interchangeable ($(1,0)\leftrightarrow(0,1)$).

{\textbf{Model Architecture}:}
To fall in with the above considerations, we chose to base the model on a deep architecture that can process sets, and particularly, the Set Transformer (ST) architecture~\citep{lee2019set}. Another option
could be based on PointNet~\citep{qi2017pointnet}; however, we chose ST due to its relative simplicity, smaller model size and better performance (in our context). The ST architecture is permutation invariant, can work with any number of data points $N$ (for a fixed $d$), and learns the relations and interactions between the data points through an attention mechanism.

We formulated the model prediction as a binary ``segmentation'' task: each point in the input is assigned to either of the two subclusters via a binary label ``0'' or ``1''.  To allow this, we need the model to output a neuron per data-point in the input, corresponding to the subcluster assignment. 
To that end, we construct the ST in a particular way. Using known basic building blocks of the ST such as Induced Set Attention Block (ISAB), row Feed Forward (rFF), and Pooling by Multihead Attention (PMA), 
we use the following encoder and decoder: 
The \textbf{Encoder}, denoted by $H_X = \mathrm{ISAB}_{L} (X)$, is a stack of $L$ components of $\mathrm{ISAB}()$.
The \textbf{Decoder}, denoted by $Z = \mathrm{rFF}(\mathrm{PMA}_{M}(H_X))$, is a stack of $M$ components of $\mathrm{PMA}()$ followed by a single row-Fully-Connected layer.
For more details on the ST architecture, these building blocks,
and the $L$ and $M$ hyperparameters, see the appendix.

\begin{figure}[t!]
\centering
  % \subcaptionbox{Easy Data Samples \label{fig:data_samples:easy}}
  % {\includegraphics[width=0.9\linewidth]{../figures/data_samples_easy.pdf}} 
  
  % \subcaptionbox{Medium Data Samples \label{fig:data_samples:medium}}
  % {\includegraphics[width=0.9\linewidth]{../figures/data_samples_med.pdf}}
  
  % \subcaptionbox{Hard Data Samples \label{fig:data_samples:hard}}
  % {\includegraphics[width=0.9\linewidth]{../figures/data_samples_hard.pdf}}
  
\includegraphics[width=0.98\linewidth]{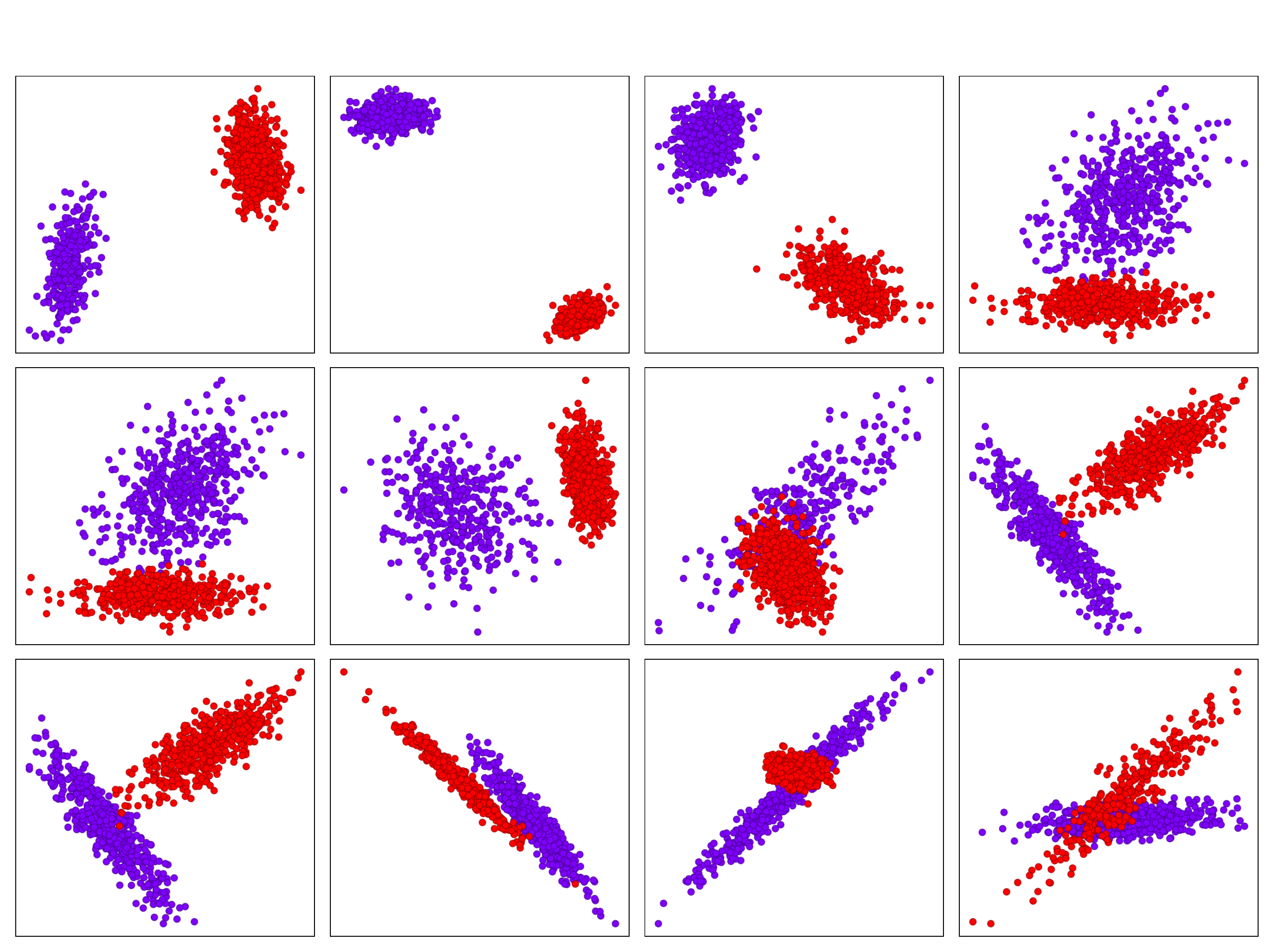}

\captionsetup{justification=justified, singlelinecheck=false}
\caption[Examples of Generated Training Data]{Examples of generated training data. Each panel represents a single training example; \ie, a cluster that needs to be split into two, according to the (ground-truth) colors. First row: ``easy'' data, middle row: ``medium-difficult'' data, last row: ``hard'' data.}
\label{fig:data_samples}
\end{figure}

\textbf{Generating Data for Training}\label{Sec:Method:Subsec:Data}
As is usual in DL, the training data is one of the most critical factors for model performance. Therefore, it must be of high quality and provided with enough variability so that the model would be able to generalize well to unseen and challenging examples (recall that here each example is an entire cluster). Fortunately, in our case, where the overarching goal is improving a DPGMM method, we can assume the training data comes from GMM. Thus, it is relatively simple to define a data-generation process with full control over various parameters such as the Gaussians' proximity to each other, the size and shape of the covariance matrices, how many points are drawn from each Gaussian, the number of Gaussians, how overlapping they are, and more. 
Recall the intuition that we would like to learn how to distinguish between the two most highly-dense regions of the data. Through experimentation, we found that training the model on exactly 2-component GMM's was better than training it on more than 2 components. This is also where the model can learn difficult splits, where other methods (\eg $K$-means) are likely to fail. We generate a dataset with specific characteristics: cluster imbalance (\ie, clusters of different sizes), non-spherical covariance matrices, and the degree of overlap between the clusters. Another critical factor when generating the data, is ensuring that the data is ``splittable'', \ie, that there exists a split with a relatively high Hastings ratio. Any sample with a low ratio is thus discarded. This mitigates the introduction of noisy data into the training.
Generated examples of samples of varying difficulty are shown in~\autoref{fig:data_samples}.
Note that each training example is an entire ``dataset'', \ie, a set of points drawn from a different 2-component GMM, where the number of points vary across the examples. 

\textbf{Loss Function.}
In each example, we know for each point from which Gaussian it was drawn. This provides synthetic Ground Truth (GT) labels: $Z=(z_i)_{i=1}^N$.  
Our supervised loss function, which uses these GT labels,  is based on the standard Binary Cross-Entropy (BCE) used for binary classification, which in our case is per data point each in example dataset. 
Recall that for a GT label $z_i \in \set{0,1}$ and a prediction $\hat{z}_i \in [0,1]$, the regular BCE loss is:
\begin{align}
     \ell_{\mathrm{BCE}}(z_i, \hat{z_i}) 
      &=- (
      z_{i}  \log \hat{z}_{i}      +
     \left(1-z_{i}\right)  \log \left(1-\hat{z}_{i}\right)) \, .
\end{align}
It is thus tempting to define (for an example dataset):
\begin{align}
      \Lcal_{\mathrm{BCE}}(Z, \hat{Z}) &=\tfrac{1}{N}{\sum\nolimits_{i=1}^{N}} \ell_{\mathrm{BCE}}(z_i, \hat{z_i}) 
      \, . 
\end{align}
However, recall that label ordering is arbitrary. As we would not want to penalize the model
for predicting, \eg, a perfect split which is the exact mirror of the GT labels,
we extend the BCE loss to ensure its invariance to such switching. 
First, we convert the GT labels to one-hot encoding vectors, $Z^{\mathrm{oh}}=(z_i^{\mathrm{oh}})_{i=1}^N$ where  
\begin{align}
    z_i^{\mathrm{oh}} = 
        \begin{cases}
            (1, 0), & \text{if } z = 0\\
            (0, 1),& \text{if } z = 1
        \end{cases}\, . 
\end{align}
With this new matrix $Z^{\mathrm{oh}}$, we compute the BCE loss per the two possible cases and then take the minimum, 
making the loss invariant to label-switching: 
\begin{align}
    & \Lcal_{\text{SplitNet}}(Z^{\mathrm{oh}}, \hat{Z}) = \min_{j\in \set{0,1}}  \tfrac{1}{N} \sum\nolimits_{i=1}^{N} \ell_{\mathrm{BCE}}(z_{i,j}^{\mathrm{oh}}, \hat{z}_i)% \right)
\end{align}
where $z_{i,j}^{\mathrm{oh}}$ is, in zero-based indexing,  the $j$ entry of $z_{i}^{\mathrm{oh}}$.

\textbf{Training.} We trained three models for dimensions $D=2,10,20$.
The full training details (most of which are standard), as well as the 
curriculum learning that we used, appear in the appendix.

% \clearpage

\section{EXPERIMENTS AND RESULTS}\label{Sec:Results}

We test the proposed approach in three different modes: 
1) SplitNet's performance on test data similar to its training data; 
2) qualitative comparisons between three variants of the SubC sampler (each with a different type of subcluster initializations); 
3) quantitative comparisons between those variants as well as other key methods on synthetic and real datasets. 

 \textbf{Performance Metrics.} We use several metrics: 
Log-posterior (LP) (when applicable); Normalized Mutual Information (NMI); Adjusted Rand Index (ARI); the inferred $K$ (or, equivalently, $K$-MAE -- the mean absolute error between true and the estimated $K$). In all of the experiments, the metrics reported (boxplots, mean+std.~dev.~values) correspond to 10 runs. 

% \begin{figure}[h!]
% \centering
%   {\includegraphics[width=0.9\linewidth]{../figures/training_compare_splits_easy.pdf}
% \caption[``Smart Splits'' on Easy-difficulty Datasets]{``Smart Splits'' on Easy-difficulty Datasets. Left column: ground truth data, Middle column: $K$-means splits, Right column: SplitNet splits}
% \label{fig:qual_comp:easy}}
% \end{figure}  

% \begin{figure}[h!]
% \centeringf
%   {\includegraphics[width=0.9\linewidth]{../figures/training_compare_splits_med.pdf}
% \caption[``Smart Splits'' on Medium-difficulty Datasets]{``Smart Splits'' on Medium-difficulty Datasets. Left column: ground truth data, Middle column: $K$-means splits, Right column: SplitNet splits}
% \label{fig:qual_comp:med}}
% \end{figure}  

\begin{figure}[t!]
\centering
  {\includegraphics[width=0.98\linewidth]{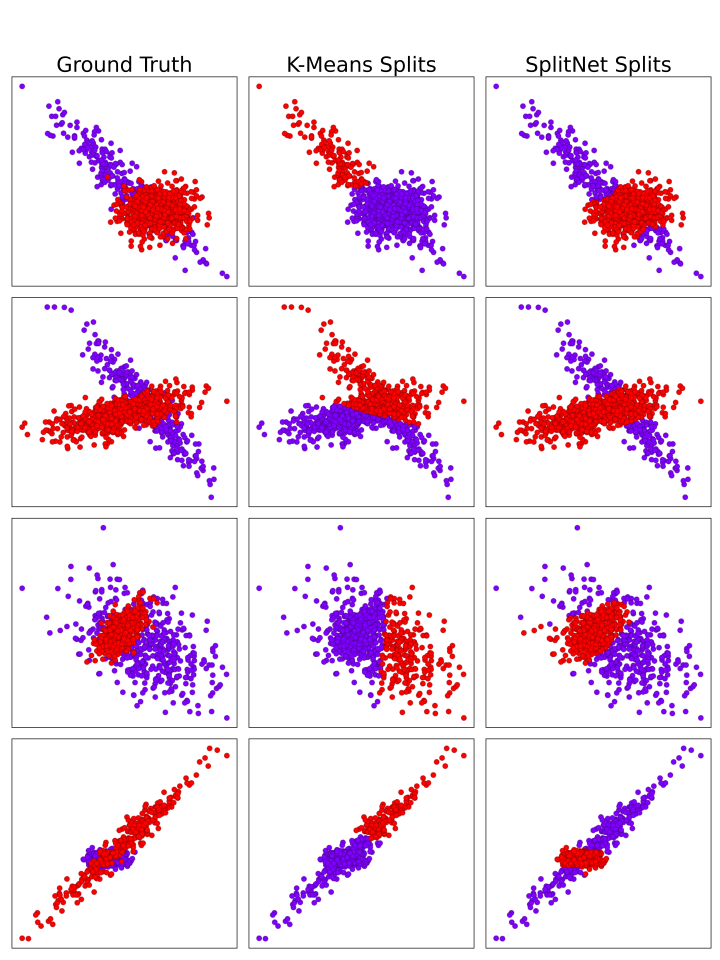}
  % {\includegraphics[width=0.98\linewidth,trim=0 0.5cm 0 3.2cm, clip]{/training_compare_splits_hard.png}
  \captionsetup{justification=justified, singlelinecheck=false}
  \caption{Splits on difficult test datasets. \textbf{Left column}: GT, \textbf{Middle}: $K$-means, \textbf{Right}: SplitNet
  (the last example is a success too, despite the reversed colors
  caused by label switching).}
\label{fig:qual_comp:hard}}
\end{figure}

\textbf{Choice of Test data.} 
We divide the experiments into synthetic data consisting of GMM data of various dimensions and numbers of clusters, and real data derived from several computer-vision datasets. 
\begin{figure*}[h!]
  \centering
  \newcommand{\myW}{0.23}
  \newcommand{\myWidth}{40mm}
  \newcommand{\myV}{-0.2cm}
  \subcaptionbox{
    Inferred $K$
    \label{fig:res:2d_k20:k}}
  [\myW\linewidth]{\includegraphics[width=\myWidth]{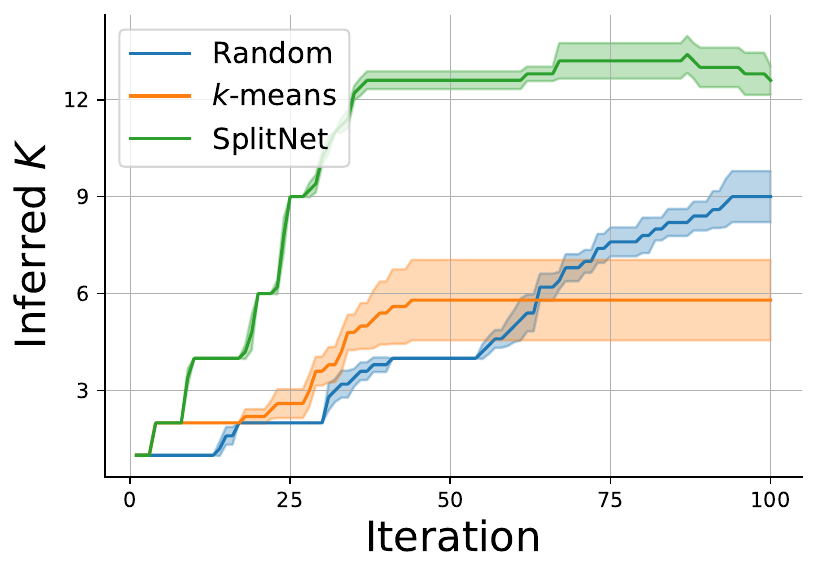}\vspace{\myV}} 
 \subcaptionbox{
     Log-Posterior
    \label{fig:res:2d_k20:ll}}
  [\myW\linewidth]{\includegraphics[width=\myWidth]{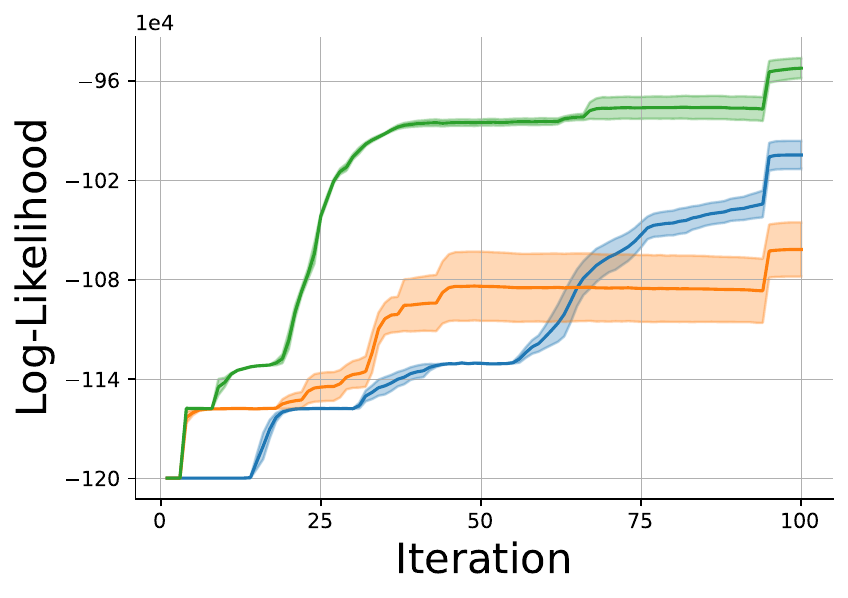}\vspace{\myV}}
  \subcaptionbox{
    NMI
    \label{fig:res:2d_k20:nmi}}    
  [\myW\linewidth]{\includegraphics[width=\myWidth]{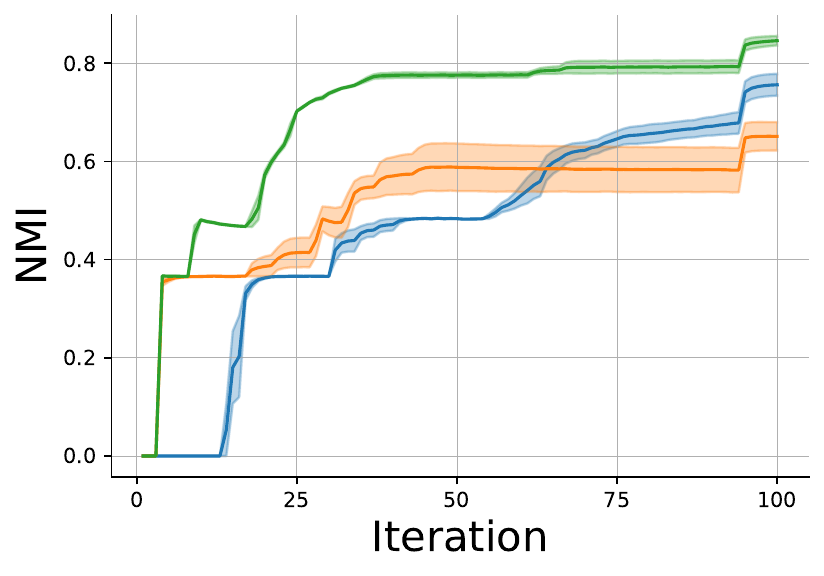}\vspace{\myV}} 
  \subcaptionbox{
    ARI
    \label{fig:res:2d_k20:ari}}
  [\myW\linewidth]{\includegraphics[width=\myWidth]{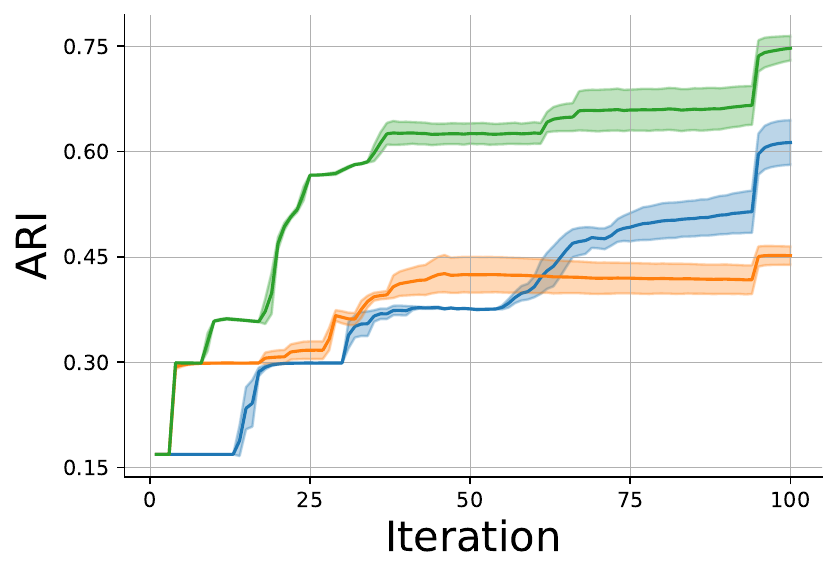}\vspace{\myV}} \\
  
  \subcaptionbox{
    GT
    \label{fig:res:2d_k20:gt_data}}
  [\myW\linewidth]{\includegraphics[width=\myWidth]{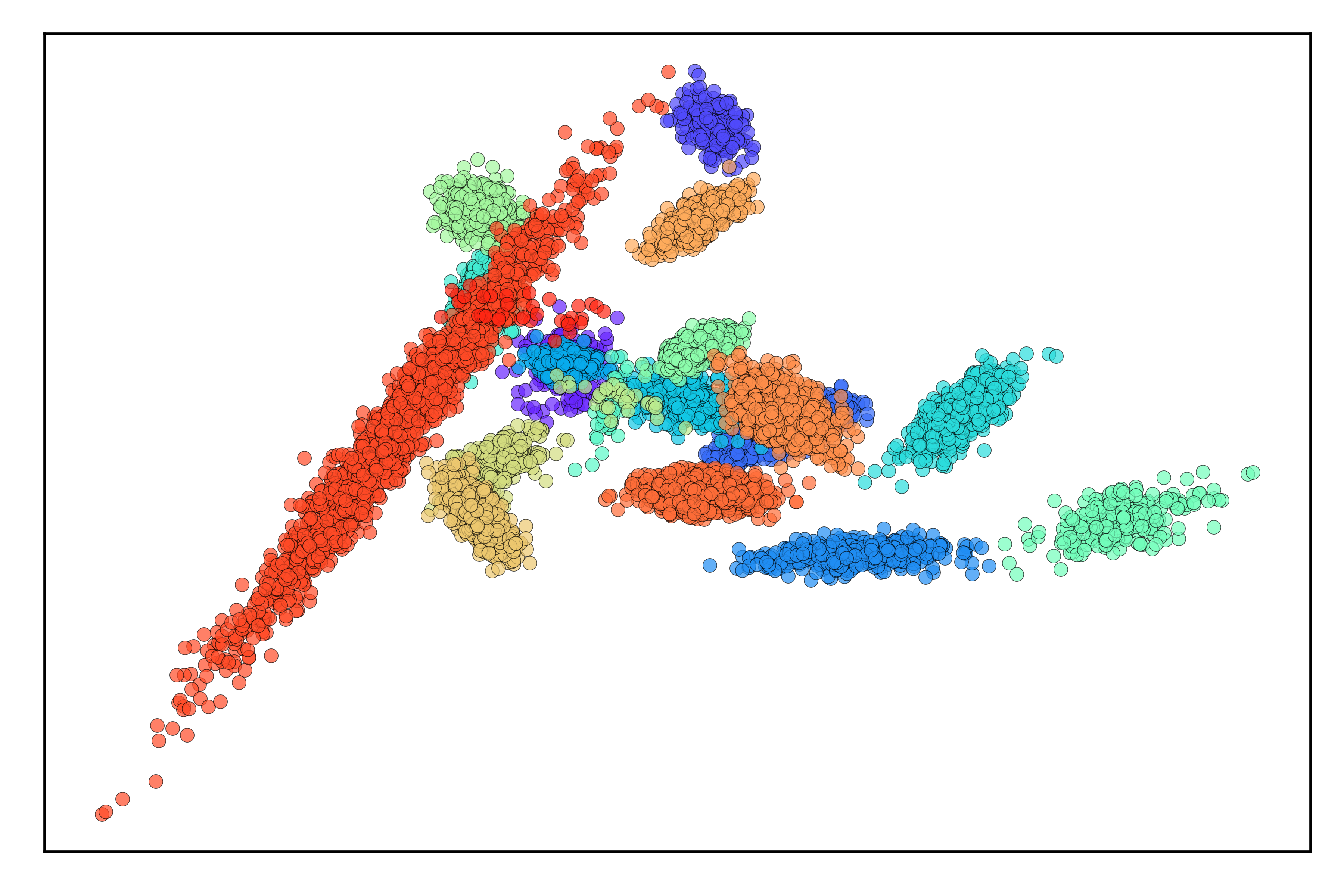}\vspace{\myV}} 
  \subcaptionbox{
    SubC-Random
    \label{fig:res:2d_k20:pred_random}}
  [\myW\linewidth]{\includegraphics[width=\myWidth]{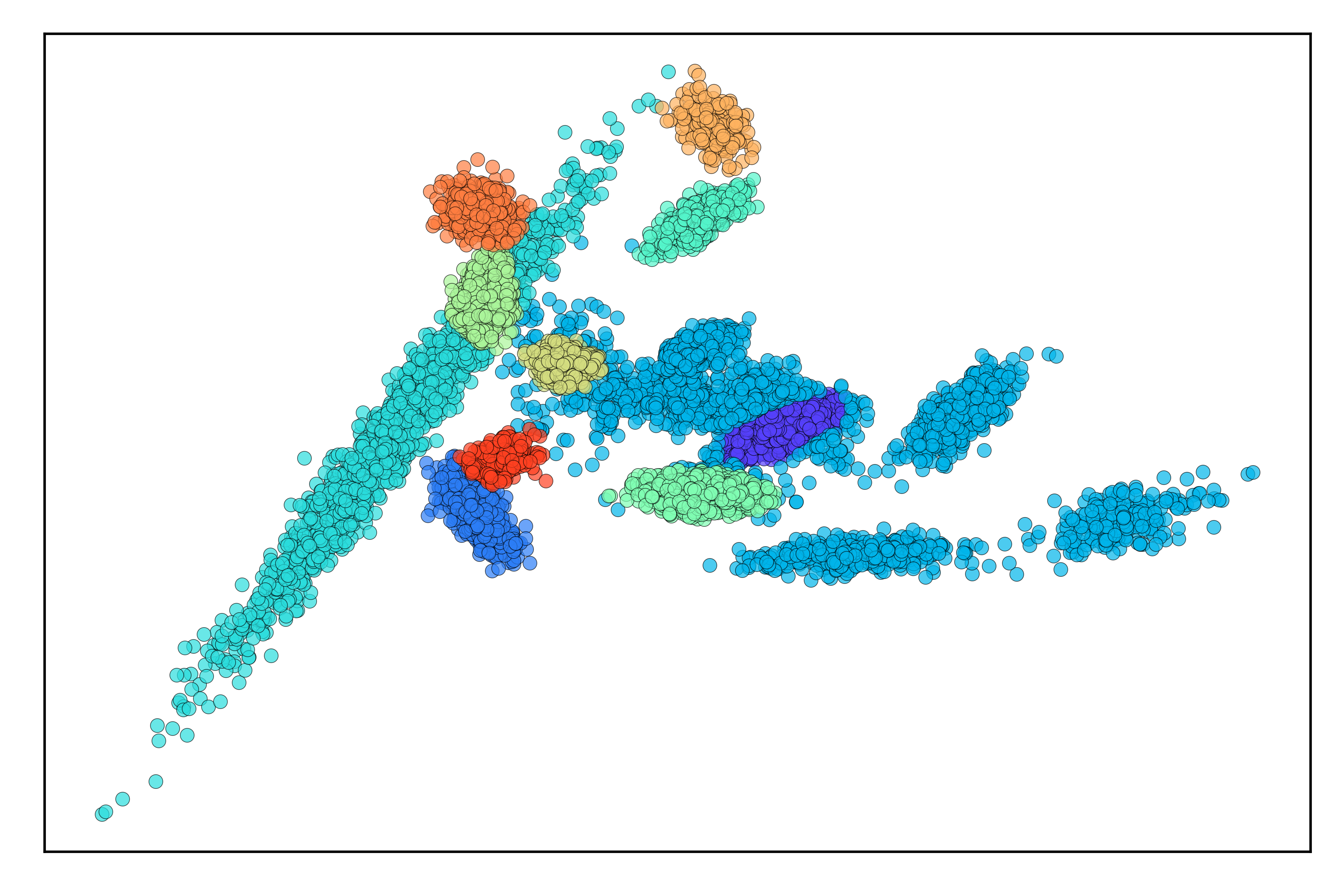}\vspace{\myV}} 
  \subcaptionbox{
     SubC-$K$-Means
    \label{fig:res:2d_k20:pred_kmeans}}
  [\myW\linewidth]{\includegraphics[width=\myWidth]{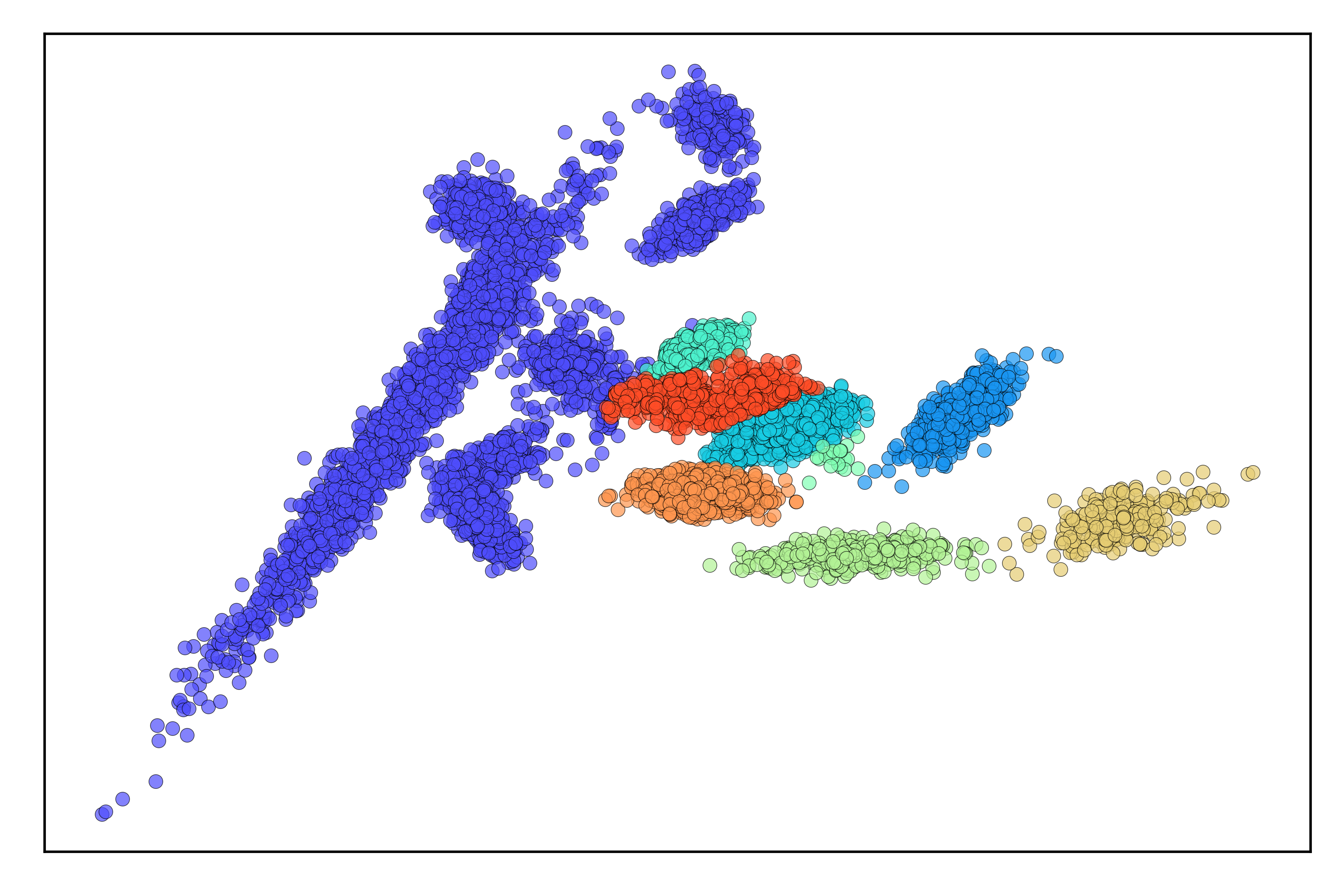}\vspace{\myV}} 
  \subcaptionbox{
     SubC-SplitNet
    \label{fig:res:2d_k20:pred_splitnet}}
  [\myW\linewidth]{\includegraphics[width=\myWidth]{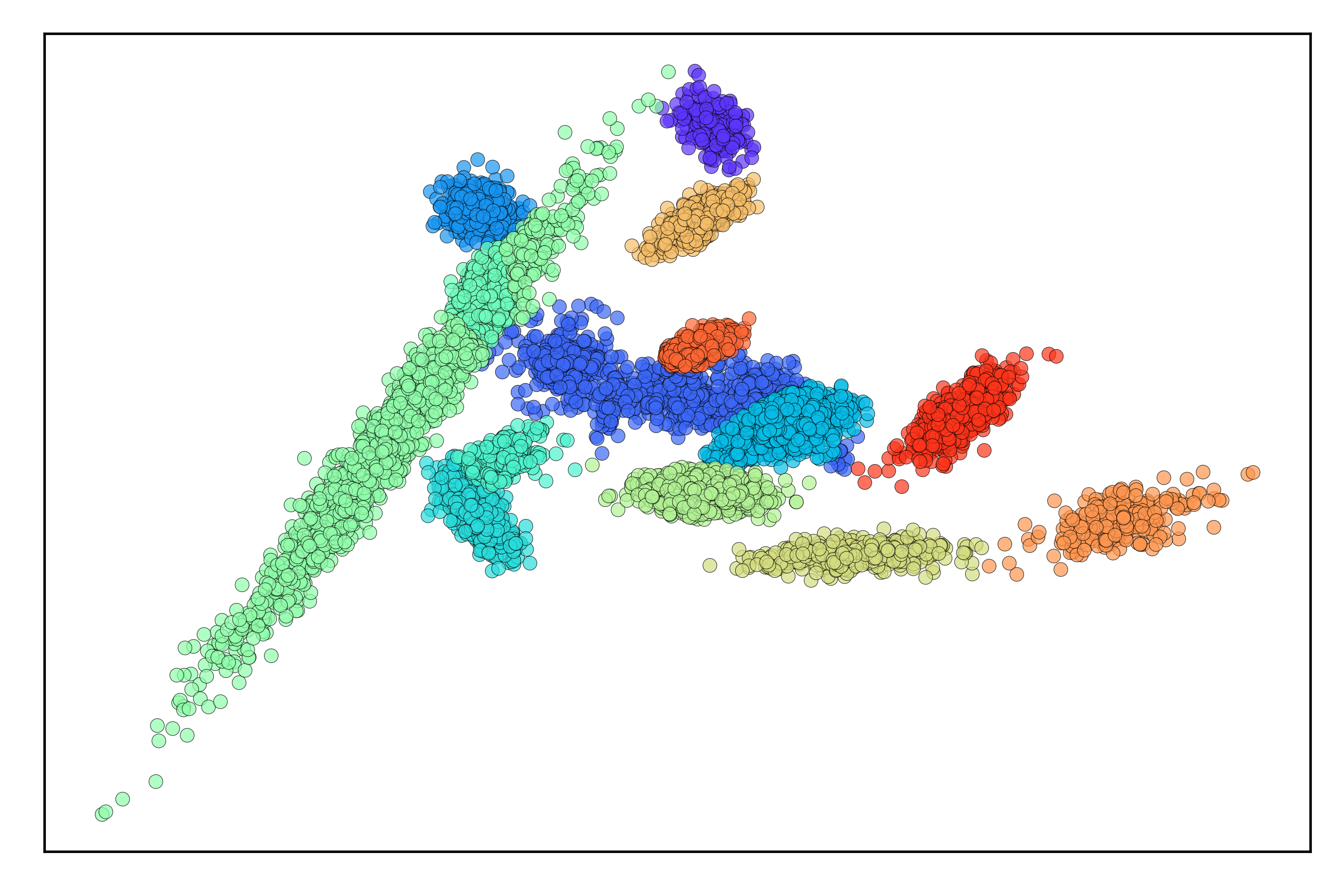}\vspace{\myV}}  
  \vspace{-0.15cm}
\caption{Subcluster initialization types: Comparison on 2D Data with $K$=20.}
\label{fig:res:2d_k20}
\end{figure*}

\begin{figure*}[h]
\centering
  \subcaptionbox{Inferred $K$ per Methods\label{fig:results:2D:boxplot:K}}
  {\includegraphics[width=0.32\linewidth,trim={4mm 4mm 5mm 3mm},clip]{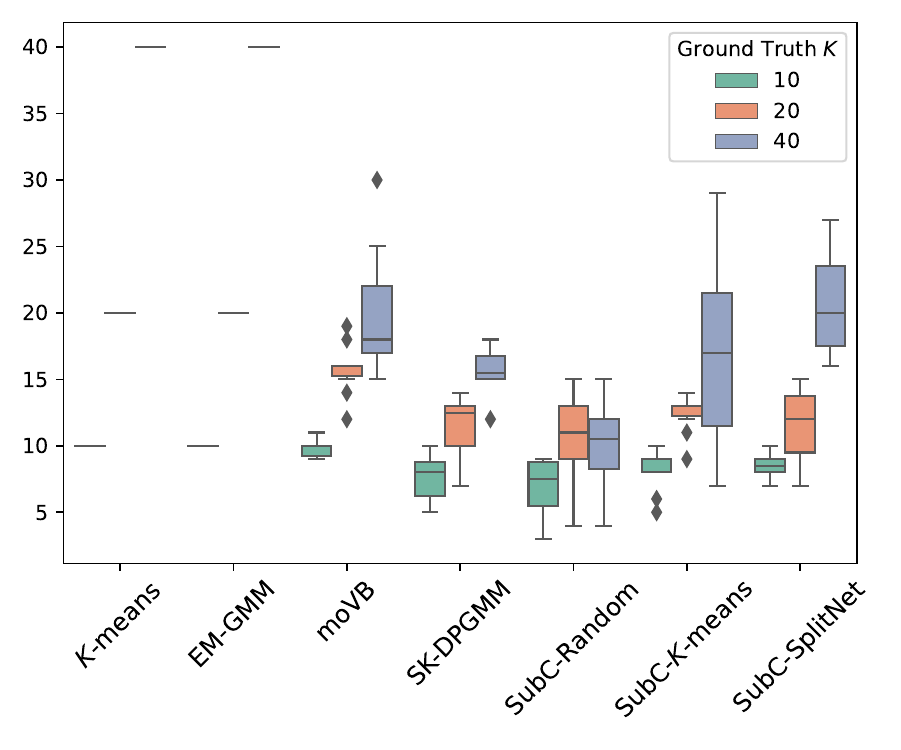}\vspace{-0.1cm}} 
  \subcaptionbox{NMI per Methods\label{fig:subc_failure:NMI}}
  {\includegraphics[width=0.32\linewidth,trim={4mm 4mm 5mm 3mm},clip]{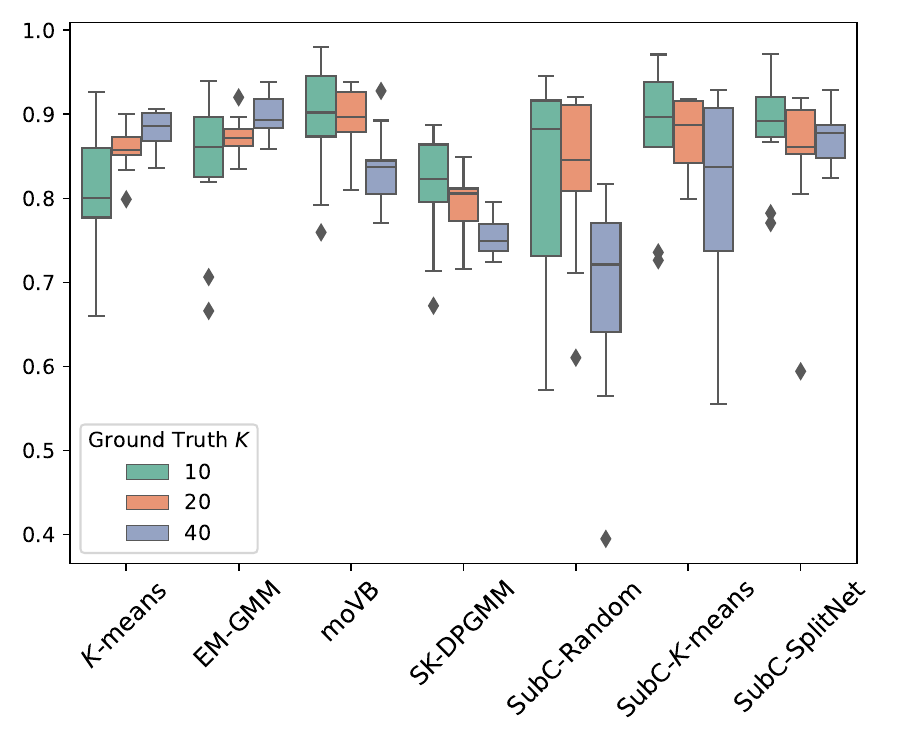}\vspace{-0.1cm}}
  \subcaptionbox{ARI per Methods\label{fig:subc_failure:ARI}}
  {\includegraphics[width=0.32\linewidth,trim={4mm 4mm 5mm 3mm},clip]{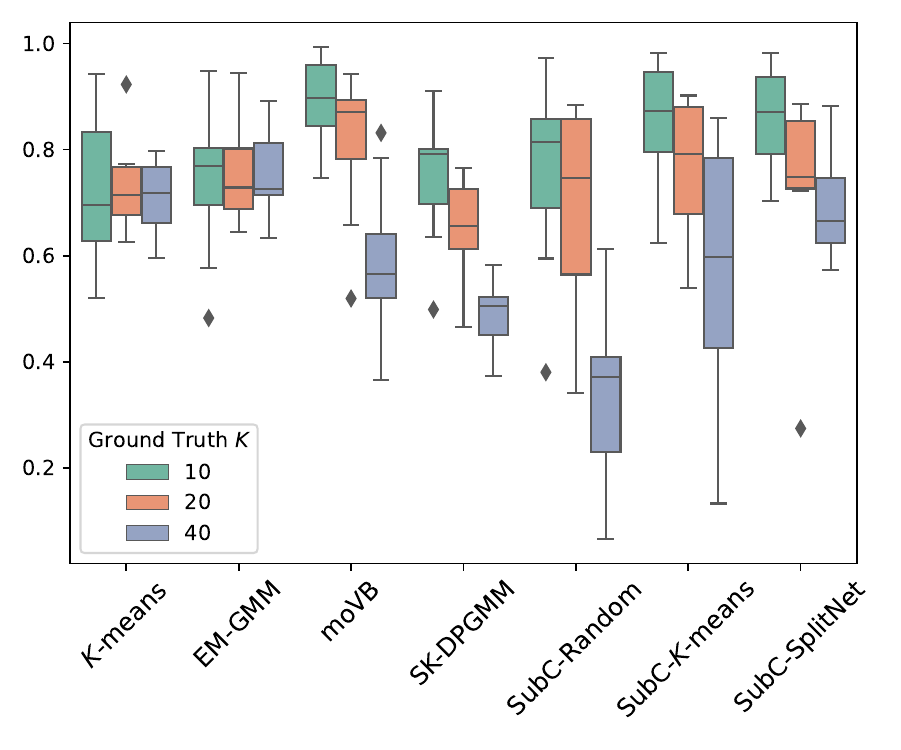}\vspace{-0.1cm}}
%   \subcaptionbox{Time per Methods\label{fig:subc_failure:time}}
%   {\includegraphics[width=0.24\linewidth,trim={4mm 4mm 5mm 3mm},clip]{../figures/boxplots/2D_Time.pdf}}  
\vspace{-.2cm} 
\caption{Performance on 2D Datasets.}
\label{fig:results:2D:boxplot}
\end{figure*}

\textbf{Algorithms.}
Chiefly, we seek to study the effect of the subcluster-initialization type on the SubC sampler. 
To that aim, we compare the same SubC sampler from~\cite{Chang:NIPS:2013:ParallelSamplerDP} (using its more recent
implementation~\citep{Dinari:CCGRID:2019:distributed}) but with three different subcluster-initialization types:
1) Random (\ie, the baseline), denoted as SubC-Random.
2) 2-means (using the $K$-means implementation from Julia's Clustering package), denoted as SubC-$K$-means.
3) SplitNet (based on the prediction of our trained deep net), denoted as SubC-SplitNet.
We also included two other DPGMM methods: 
4) A DPGMM implementation from scikit-learn~\citep{scikit-learn}, denoted as SK-DPGMM. It is a Bayesian GMM with a Dirichlet process prior fitted with variational inference. That implementation requires the specification 
of an upper bound on $K$ (set to $1000$) 
and 
5) The Memoized Online Variational Bayes~\citep{Hughes:NIPS:2013:memoizedDP}, denoted as moVB. 
Lastly, for completeness, we include the commonly-used clustering methods 7) $K$-means and 8) EM-GMM. 
(using scikit-learn's implementations), where to these methods we had to provide the GT $K$, giving them an unfair advantage.

\subsection{Synthetic Data} 
We start with a sanity check. We visually compared SplitNet and $K$-means on the test sets generated from the same generative process used to create SplitNet's training data. 
\autoref{fig:qual_comp:hard}, showing select examples of ``hard-to-split'' datasets, suggests the SplitNet predicts better splits than those found by $K$-means. For more results see the appendix. 
We now return to the original task of clustering with an unknown $K$. 

\textbf{Qualitative Comparisons.}
First, we compare SubC-Random, SubC-$K$-means, and SubC-SplitNet on multiple 2D synthetic GMM datasets, and run each method 10 times on the \textit{same} dataset. 
In~\autoref{fig:res:2d_k20}, which presents a fairly-difficult example of such a dataset, we see that the SubC-SplitNet achieves the fastest and better convergence and was also most stable (\ie, low variance). 
Note that (in this particular example) SubC-$K$-means was better than SubC-Random at the early steps of the runs but was eventually outperformed by it.  Additional examples appear in the
appendix.
\begin{table*}[h!]
    % \captionsetup{justification=centering, singlelinecheck=false}
    \caption{Results on Real Datasets}
%     \small
    \resizebox{\columnwidth*2}{!}{
    \setlength\tabcolsep{9pt}
    
    \begin{tabular}{@{}lllllll|| ll@{}}
    \toprule
                            &                                                                                                       &\begin{tabular}[c]{@{}l@{}}SK-\\DPGMM\end{tabular}                                                     & MoVB                                                                                                                                          &\begin{tabular}[c]{@{}l@{}}SubC-\\Random\end{tabular}                                                  &\begin{tabular}[c]{@{}l@{}}SubC-\\$K$-means\end{tabular}                                                                        &\begin{tabular}[c]{@{}l@{}}SubC-\\SplitNet\end{tabular}                                                                                                           &$K$-means                                                                                                                  & EM-GMM                                                                                                        \\ \midrule
        MNIST               & \begin{tabular}[c]{@{}l@{}}$K$-MAE:\\ NMI:\\ ARI:\\ \end{tabular}                          &\begin{tabular}[c]{@{}l@{}}$10.0\pm.00$\\$0.68\pm0.01$\\$0.47\pm0.02$\\   \end{tabular}                           &\begin{tabular}[c]{@{}l@{}}$1.00\pm 0.00 $\\$0.64\pm 0.00$\\$0.46\pm 0.00$\\                            \end{tabular}                          &\begin{tabular}[c]{@{}l@{}}$0.80\pm 0.44$\\$0.66\pm 0.04$\\$0.48 \pm 0.04$\\ \end{tabular}             &\begin{tabular}[c]{@{}l@{}}$\mathbf{0.00 \pm 0.00}$\\$0.68 \pm 0.01$\\$0.51\pm 0.00$\\          \end{tabular}                   &\begin{tabular}[c]{@{}l@{}}$\mathbf{0.00 \pm 0.00}$\\$\mathbf{0.69 \pm 0.01} $\\$\mathbf{0.52 \pm 0.01} $\\       \end{tabular}                      &\begin{tabular}[c]{@{}l@{}}$ - $\\$0.51 \pm 0.01$\\$0.37 \pm 0.01$\\ \end{tabular}                          &\begin{tabular}[c]{@{}l@{}}$ - $\\$0.67 \pm 0.01 $\\$0.53 \pm 0.04 $\\ \end{tabular}              \\ \midrule
        FMNIST              & \begin{tabular}[c]{@{}l@{}}$K$-MAE:\\ NMI:\\ ARI:\\ \end{tabular}                          &\begin{tabular}[c]{@{}l@{}}$10.0\pm.00$\\$0.57\pm0.01$\\$0.35\pm0.01$\\   \end{tabular}                           &\begin{tabular}[c]{@{}l@{}}$\mathbf{0.00\pm 0.00} $\\$\mathbf{0.58\pm 0.00}$\\$\mathbf{0.39\pm 0.00}$\\ \end{tabular}                          &\begin{tabular}[c]{@{}l@{}}$1.40\pm 0.54$\\$0.56\pm 0.01$\\$0.36 \pm 0.01$\\ \end{tabular}             &\begin{tabular}[c]{@{}l@{}}$1.69 \pm 0.89$\\$0.57 \pm 0.00$\\$0.36\pm 0.01 $\\                  \end{tabular}                   &\begin{tabular}[c]{@{}l@{}}$1.20 \pm 2.16$\\$\mathbf{0.58 \pm 0.01} $\\$0.38 \pm 0.01 $\\                         \end{tabular}                      &\begin{tabular}[c]{@{}l@{}}$ - $\\$0.51 \pm 0.01$\\$0.53 \pm 0.00$\\ \end{tabular}                          &\begin{tabular}[c]{@{}l@{}}$ - $\\$0.57 \pm 0.02 $\\$0.38 \pm 0.01 $\\ \end{tabular}                 \\ \midrule
        CIFAR10             & \begin{tabular}[c]{@{}l@{}}$K$-MAE:\\ NMI:\\ ARI:\\ \end{tabular}                          &\begin{tabular}[c]{@{}l@{}}$10.0\pm.00$\\$0.71\pm0.01$\\$0.58\pm0.01$\\   \end{tabular}                           &\begin{tabular}[c]{@{}l@{}}$14.00\pm 0.00$\\$0.68\pm 0.00$\\$0.53\pm 0.00$\\                            \end{tabular}                          &\begin{tabular}[c]{@{}l@{}}$1.30\pm 0.68$\\$0.68\pm 0.02$\\$0.55 \pm 0.04$\\ \end{tabular}             &\begin{tabular}[c]{@{}l@{}}$\mathbf{1.00 \pm 0.82}$\\$0.72 \pm 0.02$\\$0.61\pm 0.05$\\          \end{tabular}                   &\begin{tabular}[c]{@{}l@{}}$2.00 \pm 0.00$\\$\mathbf{0.74 \pm 0.00} $\\$\mathbf{0.64 \pm 0.00} $\\                \end{tabular}                      &\begin{tabular}[c]{@{}l@{}}$ - $\\$0.79 \pm 0.00$\\$0.78 \pm 0.00$\\ \end{tabular}                          &\begin{tabular}[c]{@{}l@{}}$ - $\\$0.72 \pm 0.00 $\\$0.60 \pm 0.00 $\\ \end{tabular}              \\ \midrule
        CIFAR20             & \begin{tabular}[c]{@{}l@{}}$K$-MAE:\\ NMI:\\ ARI:\\ \end{tabular}                          &\begin{tabular}[c]{@{}l@{}}$20.0\pm.00$\\$0.44\pm0.01$\\$0.20\pm0.01$\\   \end{tabular}                           &\begin{tabular}[c]{@{}l@{}}$\mathbf{2.00\pm 0.00} $\\$0.41\pm 0.00$\\$0.19\pm 0.00$\\                   \end{tabular}                          &\begin{tabular}[c]{@{}l@{}}$6.90\pm 1.85$\\$0.43\pm 0.01$\\$0.25 \pm 0.01$\\ \end{tabular}             &\begin{tabular}[c]{@{}l@{}}$8.20 \pm 1.47$\\$\mathbf{0.56 \pm 0.01}$\\$\mathbf{0.28\pm 0.01}$\\ \end{tabular}                   &\begin{tabular}[c]{@{}l@{}}$8.50 \pm 1.95$\\$0.46 \pm 0.01 $\\$0.27 \pm 0.01 $\\                                  \end{tabular}                      &\begin{tabular}[c]{@{}l@{}}$ - $\\$0.49 \pm 0.00$\\$0.33 \pm 0.00$\\ \end{tabular}                          &\begin{tabular}[c]{@{}l@{}}$ - $\\$0.44 \pm 0.00 $\\$0.24 \pm 0.00 $\\ \end{tabular}              \\ \midrule
        STL-10              & \begin{tabular}[c]{@{}l@{}}$K$-MAE:\\ NMI:\\ ARI:\\ \end{tabular}                          &\begin{tabular}[c]{@{}l@{}}$10.0\pm.00$\\$0.63\pm0.00$\\$0.51\pm0.01$\\   \end{tabular}                           &\begin{tabular}[c]{@{}l@{}}$10.00\pm 0.00$\\$0.62\pm 0.00$\\$0.51\pm 0.00$\\                            \end{tabular}                          &\begin{tabular}[c]{@{}l@{}}$1.80\pm 1.09$\\$0.60\pm 0.03$\\$0.47 \pm 0.05$\\ \end{tabular}             &\begin{tabular}[c]{@{}l@{}}$1.20 \pm 0.83$\\$0.63 \pm 0.02$\\$0.52\pm 0.04$\\                   \end{tabular}                   &\begin{tabular}[c]{@{}l@{}}$\mathbf{1.00 \pm 0.00}$\\$\mathbf{0.67 \pm 0.00} $\\$\mathbf{0.56 \pm 0.00} $\\       \end{tabular}                      &\begin{tabular}[c]{@{}l@{}}$ - $\\$0.70 \pm 0.00$\\$0.67 \pm 0.00$\\ \end{tabular}                          &\begin{tabular}[c]{@{}l@{}}$ - $\\$0.61 \pm 0.00 $\\$0.46 \pm 0.00 $\\ \end{tabular}              \\ \midrule
    \end{tabular}
    }
    \label{tab:results:real_datasets}    
\end{table*}

\textbf{Quantitative Comparisons.}
In these comparisons we also add the other methods. 
We considered multiple GMM configurations. 
Each configuration consists of a choice of $D$, $K$, and $n$
where in each configuration we generated 10 GMMs, drawing $n$ points from each, to generate 10 different datasets. 
On each such dataset, all methods ran for a fixed number of 200 iterations (\ie no early-stopping for convergence). 
\autoref{fig:results:2D:boxplot} presents the results for $D=2$, $K \in\set{10,20,40}$, and $n=20,000$ -- computed for each method across the 10 datasets -- in boxplot form. Note that the SplitNet-based sampler degrades the least as the number of clusters increases, while achieving the best metrics, with the least variance. Additional results, in higher dimensions and higher $K$ values, are in the
appendix. 
Of note, the slowest methods, by far, are the SK-DPGMM and moVB, as they were consistently slower by about 2 and 1 orders of magnitude, respectively, compared with all the SubC samplers (whose times were similar to each other -- each was $\sim5$ [sec] -- as the number of iterations was fixed). We emphasize that all algorithms were run on a single thread (despite the fact that the SubC-* samplers support multiple threads). 
\begin{figure*}[t]
    % \vspace{-.35cm}
    \centering
          \subcaptionbox{2D Dataset}
          {\includegraphics[width=0.32\linewidth,trim={4mm 4mm 5mm 3mm},clip]{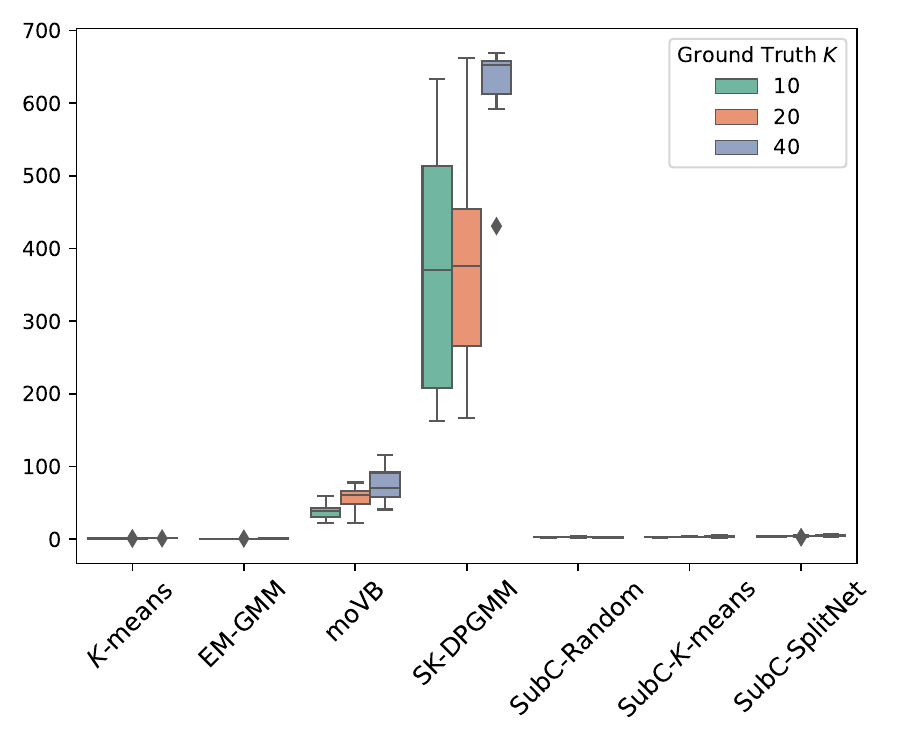}\vspace{-0.1cm}} 
          \subcaptionbox{3D Dataset}
          {\includegraphics[width=0.32\linewidth,trim={4mm 4mm 5mm 3mm},clip]{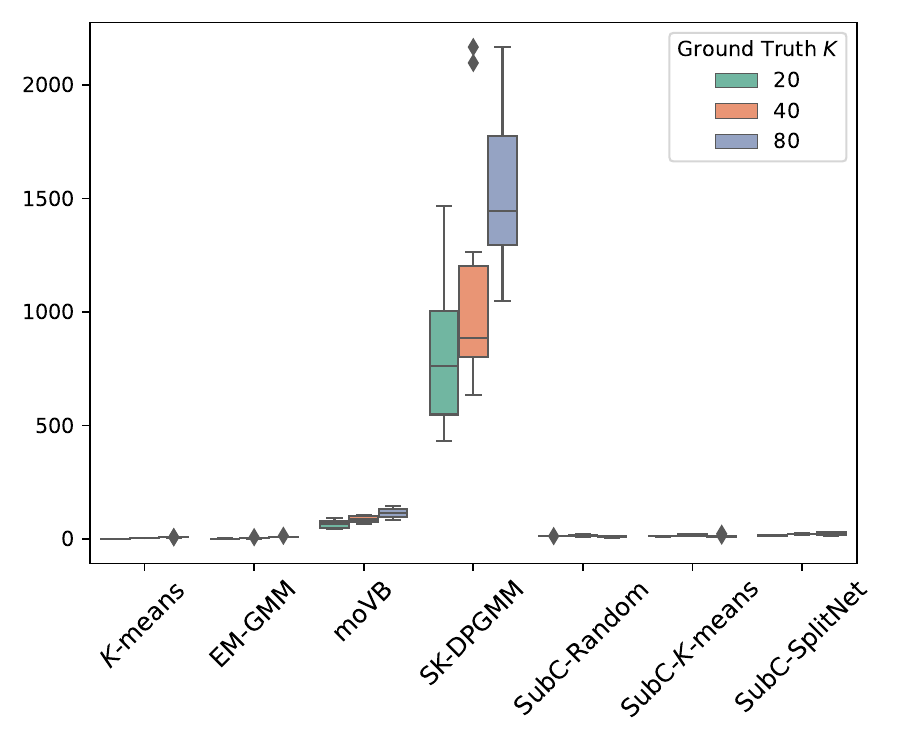}\vspace{-0.1cm}}
          \subcaptionbox{5D Dataset}
          {\includegraphics[width=0.32\linewidth,trim={4mm 4mm 5mm 3mm},clip]{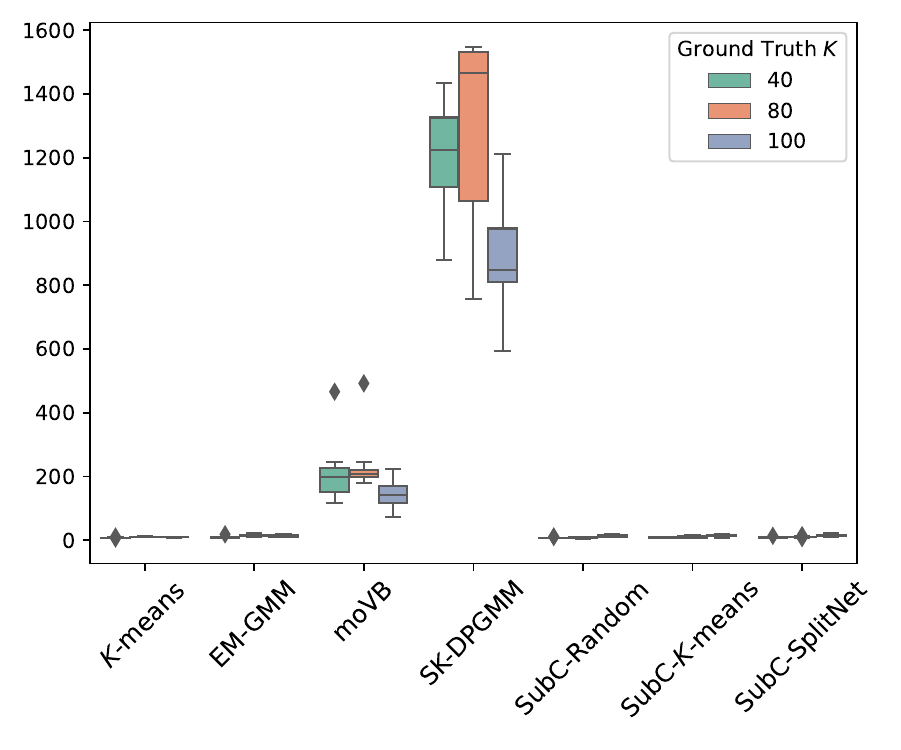}\vspace{-0.1cm}}
    % \vspace{-.2cm} 
    \caption{Running times (in [sec]) on different datasets.}
    \label{fig:results:time}
    % \vspace{-.35cm}
    \end{figure*}

\subsection{Real Data}
We now turn to the real datasets.  As these are high dimensional (\ie, images), we first use dimensionality reduction. In the MNIST and Fashion MNIST datasets, we project the  images themselves via Principal Component Analysis (PCA) to a 20-dimensional space. In CIFAR10, CIFAR20, and STL10, we compute deep features using a recent unsupervised method~\citep{vangansbeke2020scan}. The deep features are 512-dimensional, so we apply PCA on them, projecting the data to
$\RR^{20}$. Full details about the dataset can be found in the appendix. In each dataset, following the PCA, each method was run 10 times. \autoref{tab:results:real_datasets}~summarizes the results. We observe that in  most cases, the SubC-$K$-Means and SubC-SplitNet outperform the others. In fact, they sometimes even beat the parametric methods ($K$-means; EM-GMM) that were provided with the GT $K$.

\subsection{Analysis of the Results}

\textbf{Performance:} In relatively-easy datasets with a low $K$ (\eg, $K=10$ in~\autoref{fig:results:2D:boxplot}), most methods perform similarly. However, when the dataset become harder (higher $K$; more densely packed), 
the proposed SubC-$K$-Means and (more so) SubC-SplitNet usually start outperforming the others.

\textbf{Convergence Rate, Speed, and Stability:} 
As~\autoref{fig:res:2d_k20} and similar figures in the appendix show, 
SubC-$K$-Means and SubC-SplitNet converge faster than SubC-Random.  
Those figures, together with the std.~dev. values in~\autoref{tab:results:real_datasets} also demonstrate that SubC-SplitNet is the most stable (\ie, has the lowest variance) among the three methods. 

When compared with other methods (moVB; SK-DPGMM), the SubC-* methods are 1 or 2 orders of magnitude faster, as can be seen in \autoref{fig:results:time}: moVB and (especially) SK-DPGMM are much slower than the SubC samplers. Of note, the fact that the figure might suggest that SubC-SplitNet sampler took (\emph{very slightly!}) more time than the other two SubC methods is misleading: 1)
Recall we fixed the number of iterations to 200, which sufficed for convergence for all methods. However, the SubC-SplitNet needed far fewer iterations as it converged faster. 2) the implementation of SplitNet-based method combines software from Julia (from ~\cite{Dinari:CCGRID:2019:distributed}) 
and PyTorch (SplitNet). Currently, the cross talk between the languages costs a little overhead which we believe could be eliminated by some software engineering. We also emphasize that all algorithms were run on the same single machine using a single thread despite the fact that the SubC-* methods support both multiple threads and multiple machines.

{
\textbf{Performance and Generalization on Real Datasets.}
Despite the fact that SplitNet was trained on synthetic 2-component GMM data, is it is able to perform well and generalize to real datasets, outperforming all other methods on most datasets and metrics. 
}

\section{CONCLUSION}\label{Sec:Conclusion}
In this paper we have identified failure modes of the SubC sampler~\citep{Chang:NIPS:2013:ParallelSamplerDP}. These failures are tied to the random subcluster initializations.  As a remedy, we proposed two alternative subcluster initializations: $K$-means and a new DL method, called SplitNet. We showed that the SubC-SplitNet sampler usually outperforms both the baseline SubC sampler and the SubC-$K$-means sampler in performance, convergence speed, and stability. The added value is especially apparent in challenging datasets. 
SplitNet's limitation is that it is hard to scale it to very high dimensions; training on such data is expensive in both time and memory. However, in high dimensions, clusters are usually more easily separable and the failure modes of the baseline sampler are less frequent, thus better initializations are less needed there anyway. 

Future work may explore similar ideas for non-Gaussian mixtures: while it is reasonable to believe that the proposed method will also work well for many other continuous distributions, extending it to discrete distributions (\eg, multinomials) will require new ideas. Another interesting idea is to train SplitNet to directly maximize Hasting Ratio of the split.  

\clearpage
\small
\bibliographystyle{abbrvnat}
\bibliography{./ms}

% %\runningtitle{I use this title instead because the last one was very long}
% \runningtitle{Common Failure Modes of Subcluster-based Sampling in DPGMMs -- and a DL Solution (Appendix)}

% % If your paper is accepted and the number of authors is large, the
% % style will print as headings an error message. Use the following
% % command to supply a shorter version of the authors names so that
% % they can be used as headings (for example, use only the surnames)
% %
% %\runningauthor{Surname 1, Surname 2, Surname 3, ...., Surname n}
%  \runningauthor{Winter*, Dinari*, and Freifeld $\qquad$ (* = equal contribution)}
 
% Supplementary material: To improve readability, you must use a single-column format for the supplementary material.

\clearpage
\appendix
\onecolumn \makesupplementtitle
\thispagestyle{empty}

% \aistatstitle{Common Failure Modes of Subcluster-based Sampling in Dirichlet Process Gaussian Mixture Models -- and a Deep-learning Solution
% \\ -- \\
% Appendix}
% \aistatsauthor{Vlad Winter* \And Or Dinari* \And Oren Freifeld
% \newline }
% \aistatsaddress{ winterv@post.bgu.ac.il \\ Ben-Gurion University 
% \And dinari@post.bgu.ac.il\\ Ben-Gurion University 
% \And  orenfr@cs.bgu.ac.il \\ Ben-Gurion University } 

\vspace*{1cm}
% \begin{abstract}
    This appendix contains the following:
1) additional results and some more technical details about the experiments;
2) the SubC sampler's full algorithm;
3) the expressions for the posterior hyperparameters and Marginal Data Likelihood in a Gaussian Model
with an NIW prior; 
4) additional training details;
5) additional training data details;
6) details about the Set Transformer architecture. 

% \end{abstract}
% \clearpage

\section{Additional Results and Technical Details about Experiments}

For completeness, below we provide the additional results that were alluded do in the paper
but were omitted there due to space limits.

\subsection{Synthetic-data Experiments}

\subsubsection{SplitNet's Performance on Test Dataset.}

\begin{figure}[h!]
\centering
  {\includegraphics[width=0.7\linewidth]{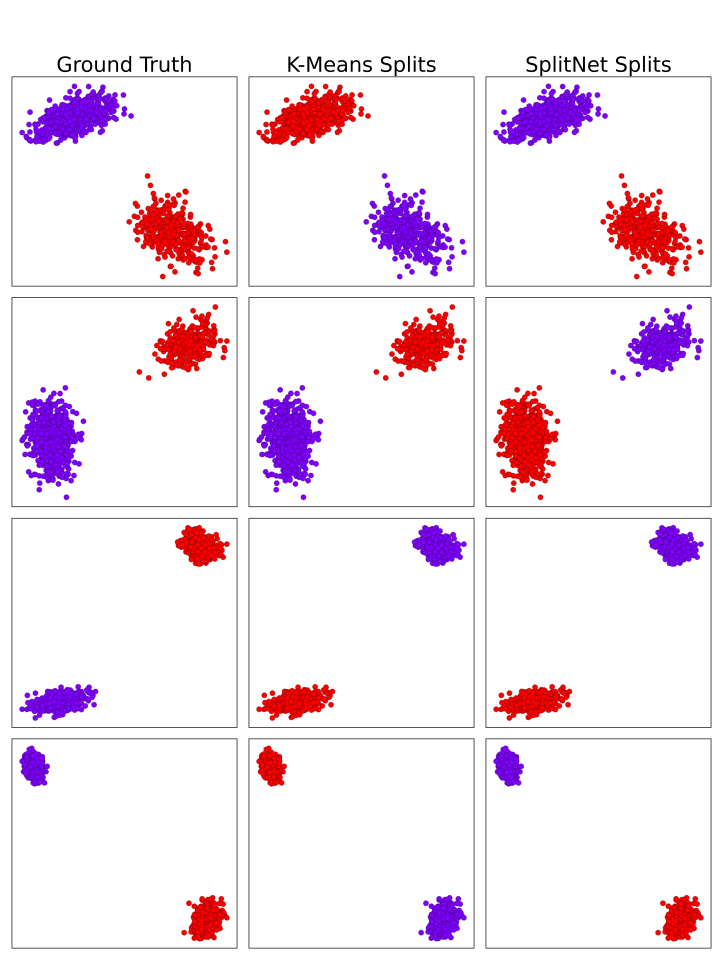}
  \caption{Splits on easy test datasets. \textbf{Left column}: GT, \textbf{Middle}: $K$-means, \textbf{Right}: SplitNet
  (Note that in some of the examples there is label switching \wrt the GT, but that by itself does not indicate a failure of either of the methods). }
\label{fig:qual_comp:easy}}
\end{figure}

\begin{figure}[h!]
\centering
  {\includegraphics[width=0.7\linewidth]{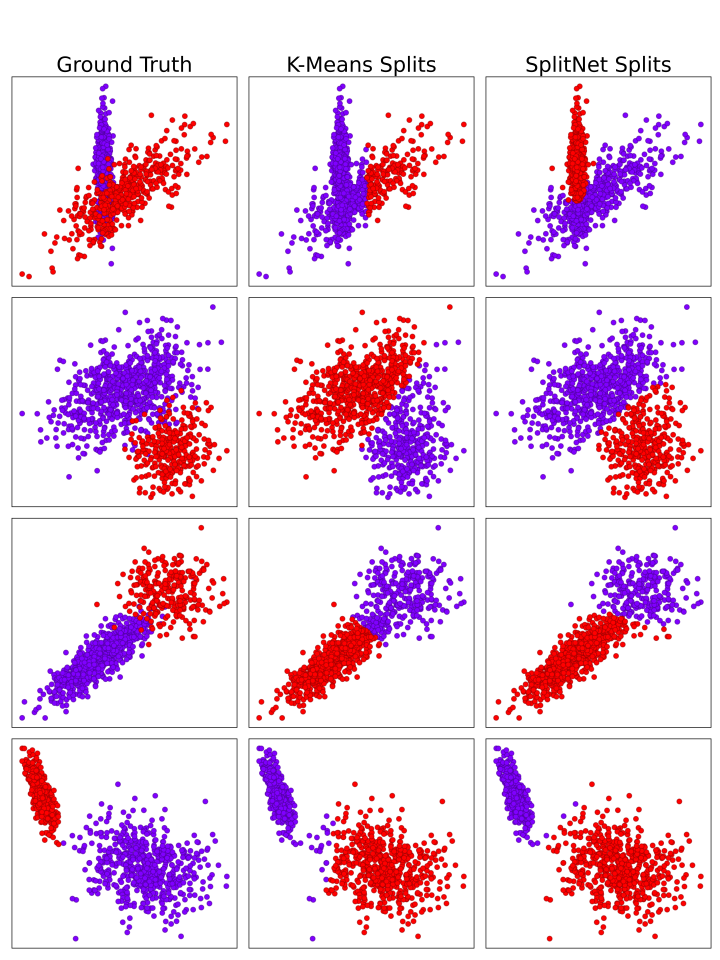}
  \caption{Splits on medium-difficulty test datasets. \textbf{Left column}: GT, \textbf{Middle}: $K$-means, \textbf{Right}: SplitNet
  (Note that in some of the examples there is label switching \wrt the GT, but that by itself does not indicate a failure of either of the methods). }
  \label{fig:qual_comp:med}}
\end{figure}

Here we visually compare SplitNet and $K$-means splits on a few test sets generated from the same generative process used to create SplitNet's training data. To that aim,  We created 3 test-sets, conceptually tagged as 
``easy'' (\autoref{fig:qual_comp:easy}),
``medium''(\autoref{fig:qual_comp:med})
and ``hard'' (\autoref{fig:qual_comp:hard}).

\begin{Remark}
 Note that in some of the examples (for either $K$-means or SplitNet) there exists a label switching \wrt the Ground Truth (GT), but that by itself does not indicate a failure of either of the methods
 since the ordering of the labels arbitrary. 
\end{Remark}

% \begin{figure}[h]
% \centering
%   {\includegraphics[width=0.75\linewidth]{/training_compare_splits_hard.png}
%   \captionsetup{justification=justified, singlelinecheck=false}
%   \caption{Splits on difficult test datasets. \textbf{Left column}: GT, \textbf{Middle}: $K$-means, \textbf{Right}: SplitNet
% (Note that in some of the examples there is label switching \wrt the GT, but that by itself does not indicate a failure of either of the methods). }\label{fig:qual_comp:hard}}
% \end{figure}  
% \clearpage

As the figures show, both methods succeed in the easy example. Both methods still do well with the medium-level examples, though it is fair to say that SplitNet slightly wins by a small margin. 
However, in most difficult example it is evident that SplitNet's significantly outperforms $K$-means.

\pagebreak

\subsubsection{Qualitative Comparisons.}

\begin{figure}[h!]
  \centering
  \subcaptionbox{Inferred $K$ 
    \label{fig:res:2d_k10:k}}
  [.24\linewidth]{\includegraphics[width=0.24\textwidth]{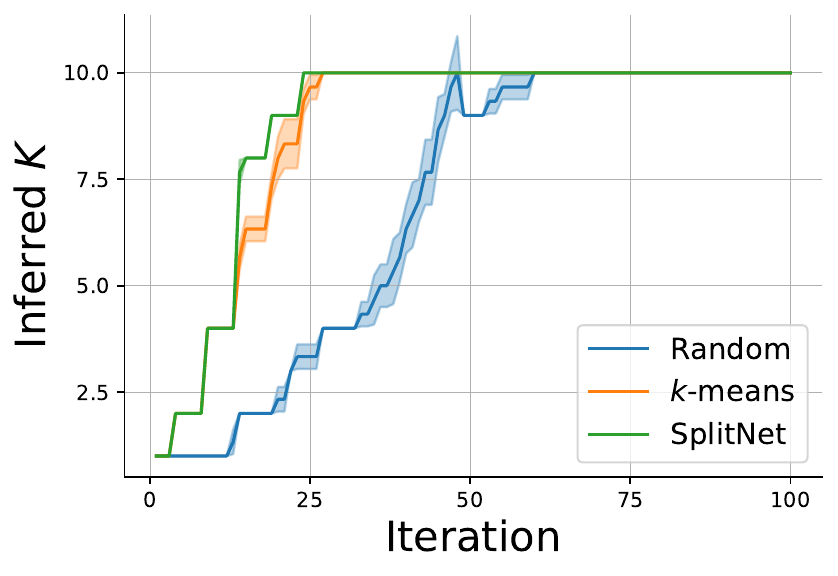}}
  \subcaptionbox{Log-Posterior 
    \label{fig:res:2d_k10:ll}}
  [.24\linewidth]{\includegraphics[width=0.24\textwidth]{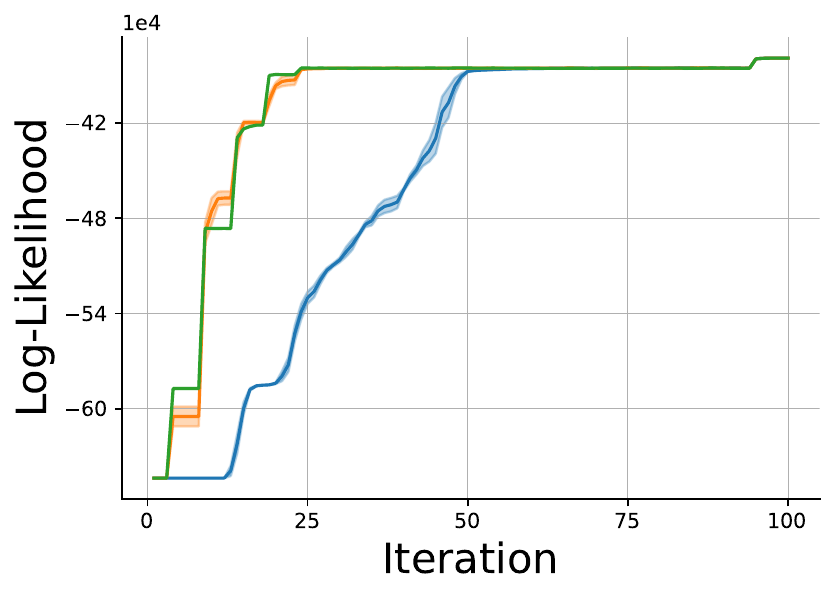}}
  \subcaptionbox{NMI 
    \label{fig:res:2d_k10:nmi}}
  [.24\linewidth]{\includegraphics[width=0.24\textwidth]{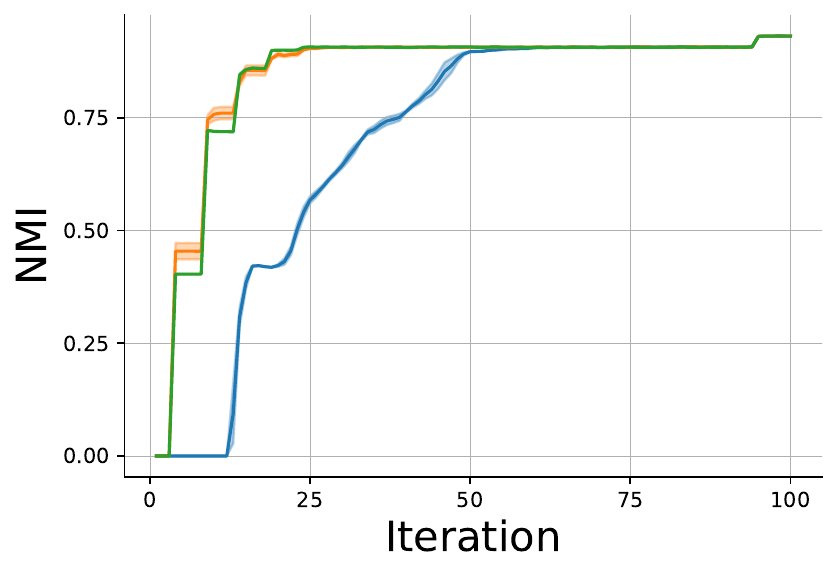}}
  \subcaptionbox{ARI 
    \label{fig:res:2d_k10:ari}}
  [.24\linewidth]{\includegraphics[width=0.24\textwidth]{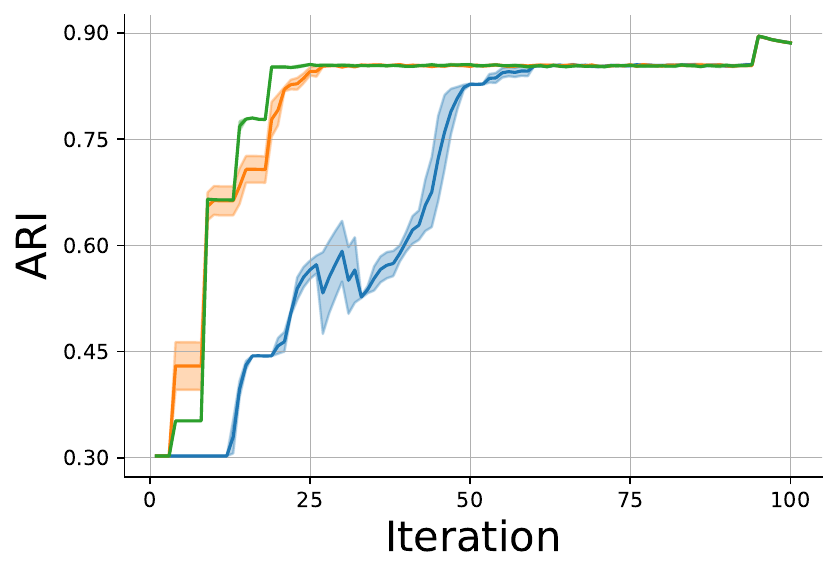}}
  
  \subcaptionbox{Ground Truth Labels
    \label{fig:res:2d_k10:gt_data}}
  [.24\linewidth]{\includegraphics[width=0.24\textwidth]{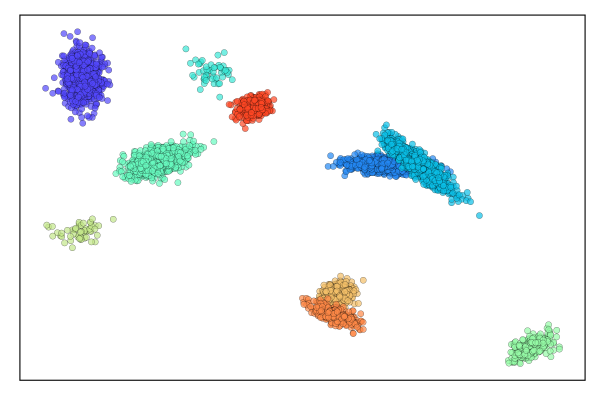}}
  \subcaptionbox{
    SubC-Random
    \label{fig:res:2d_k10:pred_random}}
  [.24\linewidth]{\includegraphics[width=0.24\textwidth]{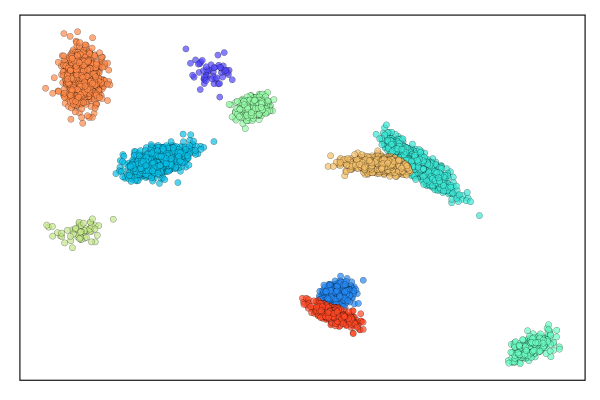}}
  \subcaptionbox{
    SubC-$K$-means
    \label{fig:res:2d_k10:pred_kmeans}}
  [.24\linewidth]{\includegraphics[width=0.24\textwidth]{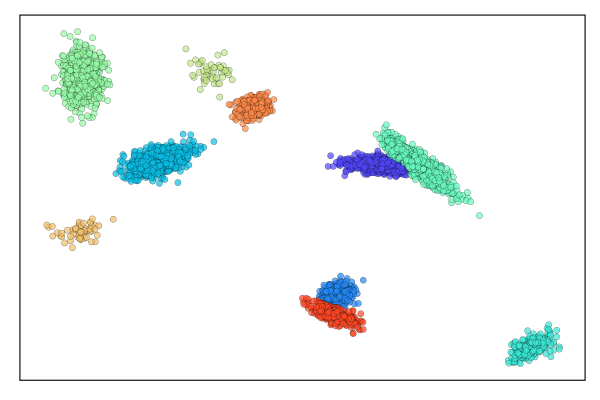}}
  \subcaptionbox{
    SubC-SplitNet
    \label{fig:res:2d_k10:pred_splitnet}}
  [.24\linewidth]{\includegraphics[width=0.24\textwidth]{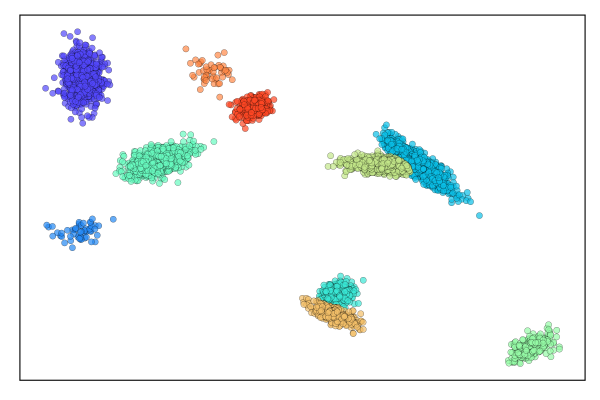}}
  \caption{Comparison on 2D Data with $K=10$.}
\label{fig:res:2d_k10}
\end{figure}

\begin{figure}[h!]
  \centering
  
  \subcaptionbox{
    Inferred $K$
    \label{fig:res:2d_k40_:k}}
  [.24\linewidth]{\includegraphics[width=0.24\textwidth]{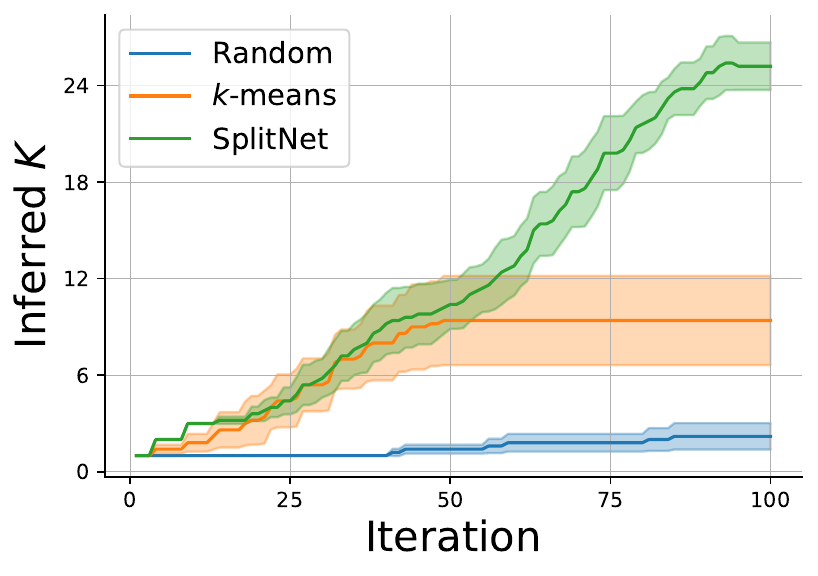}}
  \subcaptionbox{
     Log-Posterior
    \label{fig:res:2d_k40_:ll}}
  [.24\linewidth]{\includegraphics[width=0.24\textwidth]{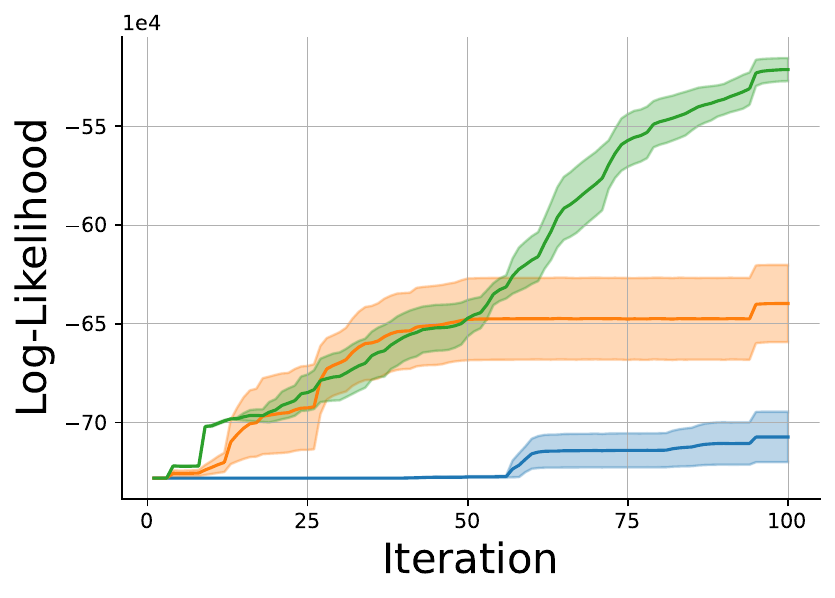}}
  \subcaptionbox{
    NMI
    \label{fig:res:2d_k40_:nmi}}
  [.24\linewidth]{\includegraphics[width=0.24\textwidth]{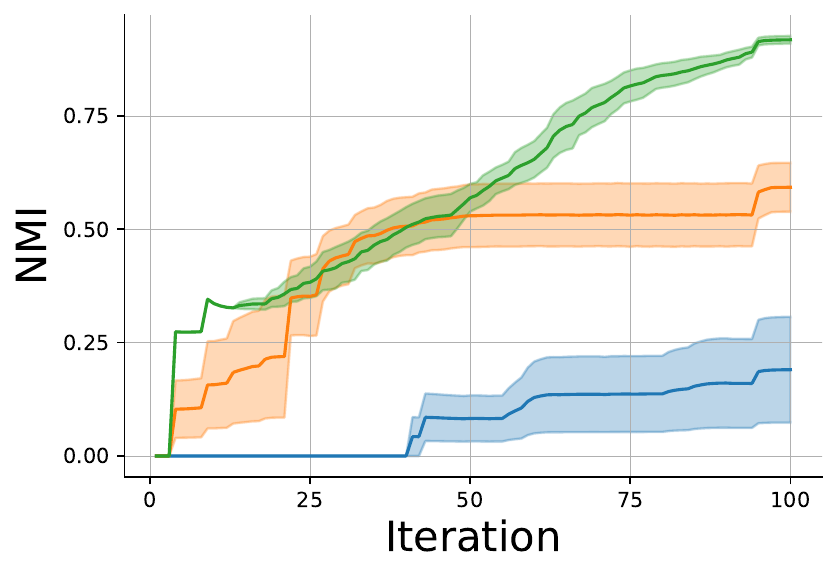}}
  \subcaptionbox{
    ARI
    \label{fig:res:2d_k40_:ari}}
  [.24\linewidth]{\includegraphics[width=0.24\textwidth]{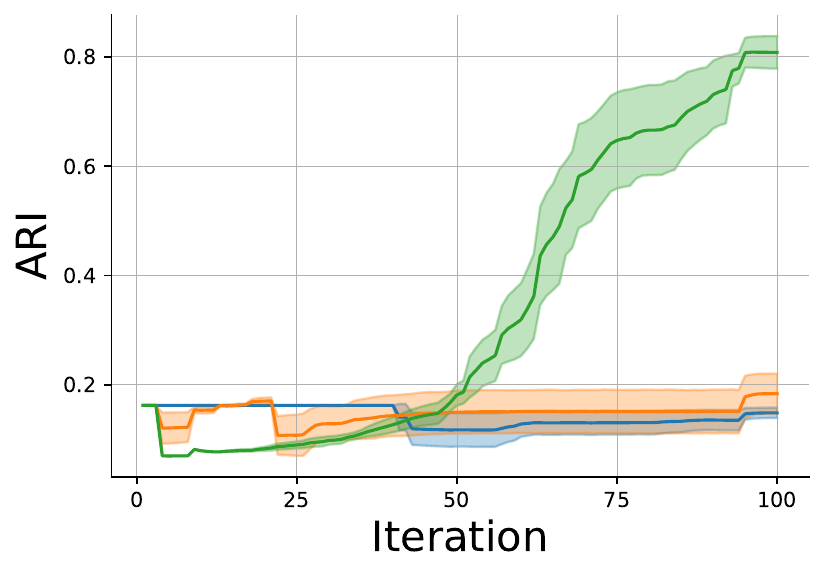}}
  
  \subcaptionbox{
    Ground Truth Labels
    \label{fig:res:2d_k40_:gt_data}}
  [.24\linewidth]{\includegraphics[width=0.24\textwidth]{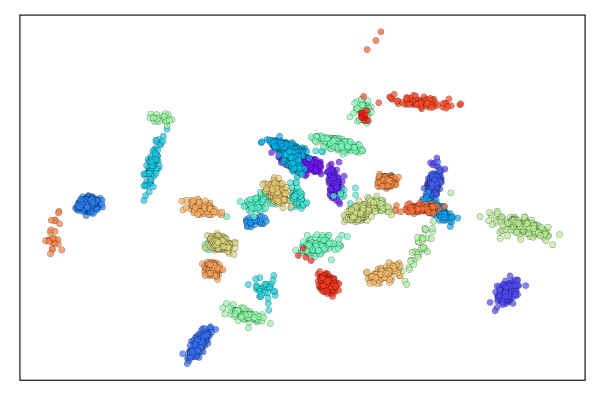}}
  \subcaptionbox{
    SubC-Random
    \label{fig:res:2d_k40_:pred_random}}
  [.24\linewidth]{\includegraphics[width=0.24\textwidth]{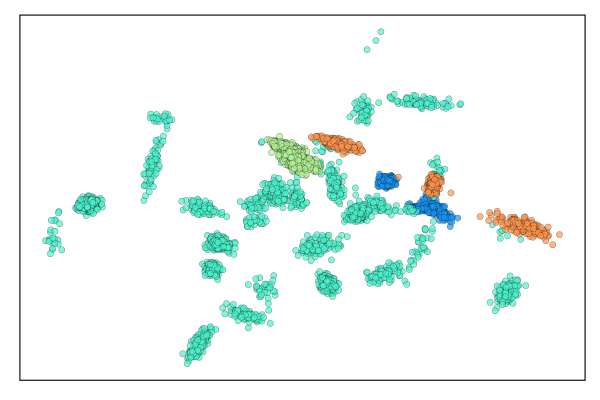}}
  \subcaptionbox{
    SubC-$K$-means
    \label{fig:res:2d_k40_:pred_kmeans}}
  [.24\linewidth]{\includegraphics[width=0.24\textwidth]{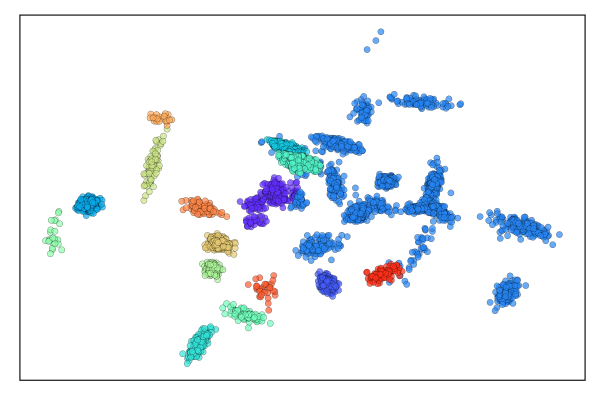}}
  \subcaptionbox{
    SubC-SplitNet
    \label{fig:res:2d_k40_:pred_splitnet}}
  [.24\linewidth]{\includegraphics[width=0.24\textwidth]{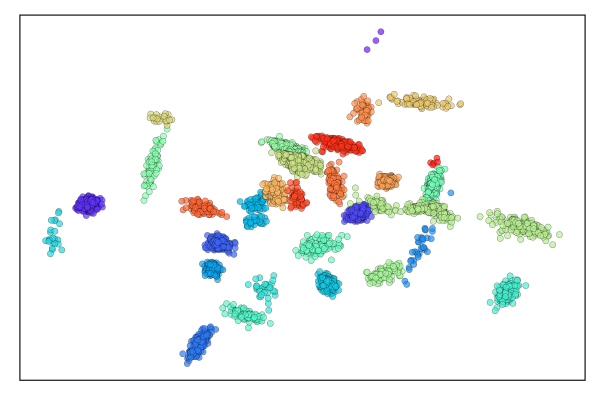}}
    
\caption{Comparison on 2D Data with $K=40$.}
\label{fig:res:2d_k40_}
\end{figure}

Here we present two more quantitative comparisons (in addition to the one in the paper) between the SubC-Random, SubC-$K$-means, and SubC-SplitNet methods on two different 2D synthetic GMM datasets. 
The first is for data drawn from a GMM with $K=10$ components,
while in the second we used a GMM with $K=40$ components.
On each of the two datasets we run each method 10 times. 
The results are shown in~\autoref{fig:res:2d_k10} ($K=10$) and~\autoref{fig:res:2d_k40_} ($K=40$). 
The results are consistent with ones shown in the paper, suggesting that SubC-SplitNet outperforms, overall,  
the other two in terms of performance, convergence speed and low variance.

\clearpage

\subsubsection{Quantitative Comparisons.}

\begin{figure*}[h!]
\centering
  \subcaptionbox{Inferred $K$ per Method\label{fig:results:3D:boxplot:K}}
  {\includegraphics[width=0.32\linewidth]{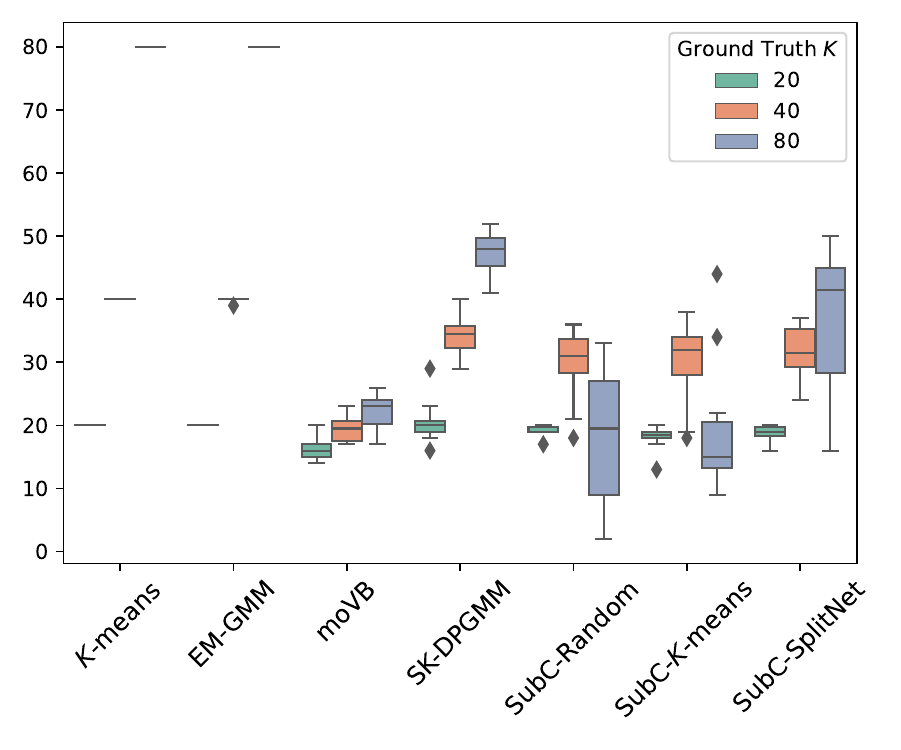}} 
  \subcaptionbox{NMI per Method\label{fig:results:3D:boxplot:NMI}}
  {\includegraphics[width=0.32\linewidth]{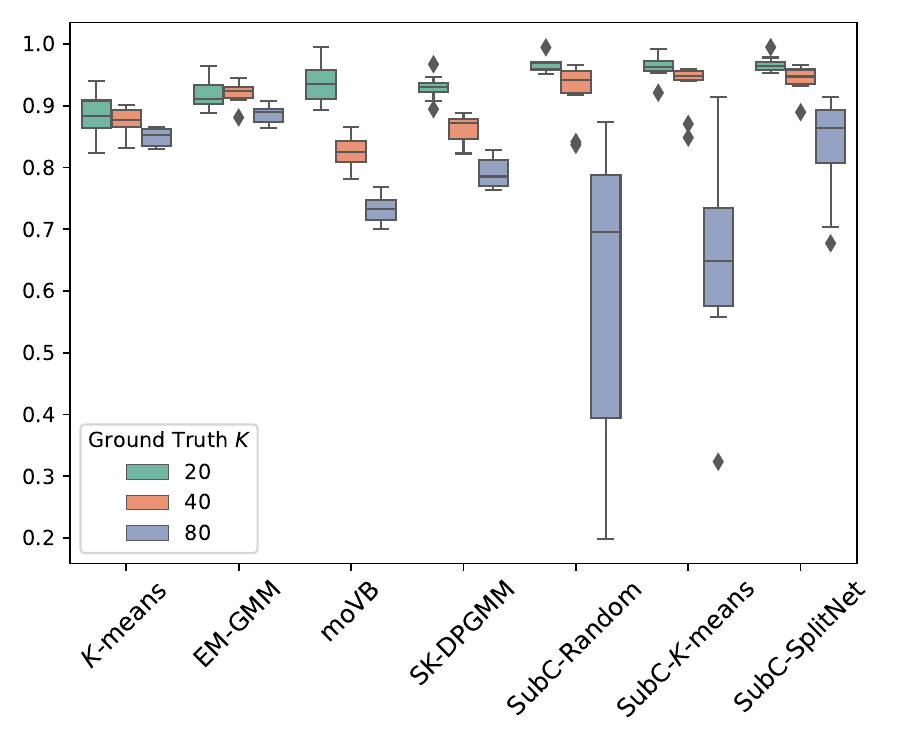}}
  \subcaptionbox{ARI per Method\label{fig:results:3D:boxplot:ARI}}
  {\includegraphics[width=0.32\linewidth]{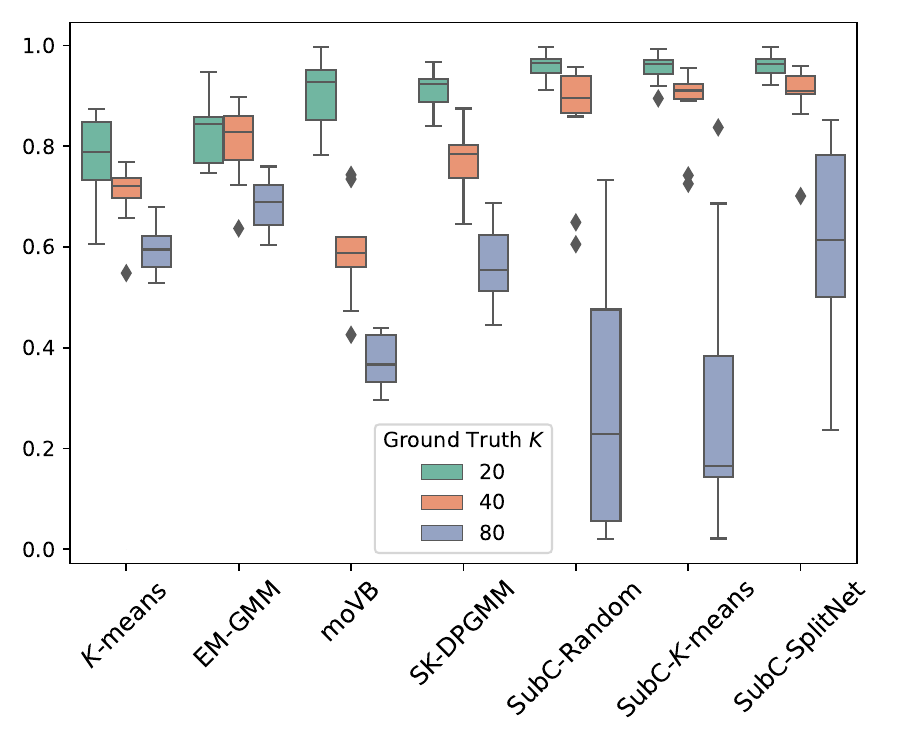}}
%   \subcaptionbox{Time per Method\label{fig:results:3D:boxplot:time}}
%   {\includegraphics[width=0.24\linewidth]{../figures/boxplots/3D_Time.pdf}}  

\caption{Performance on 3D Datasets.}
\label{fig:results:3D:boxplot}
\end{figure*}

\begin{figure*}[h!]
\centering
  \subcaptionbox{Inferred $K$ per Method\label{fig:results:5D:boxplot:K}}
  {\includegraphics[width=0.32\linewidth]{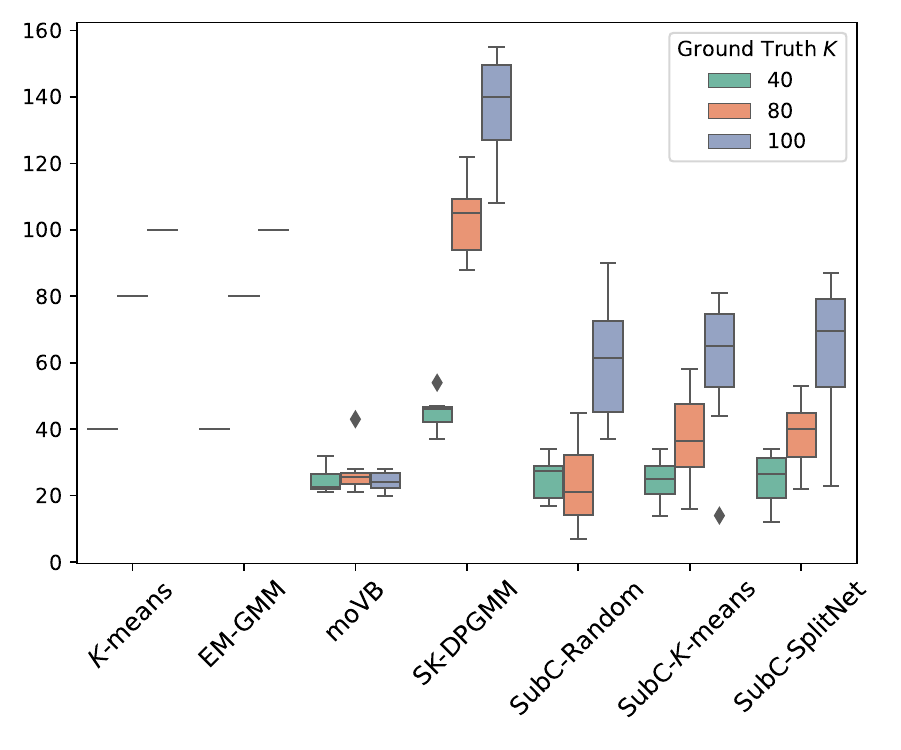}} 
  \subcaptionbox{NMI per Method\label{fig:results:5D:boxplot:NMI}}
  {\includegraphics[width=0.32\linewidth]{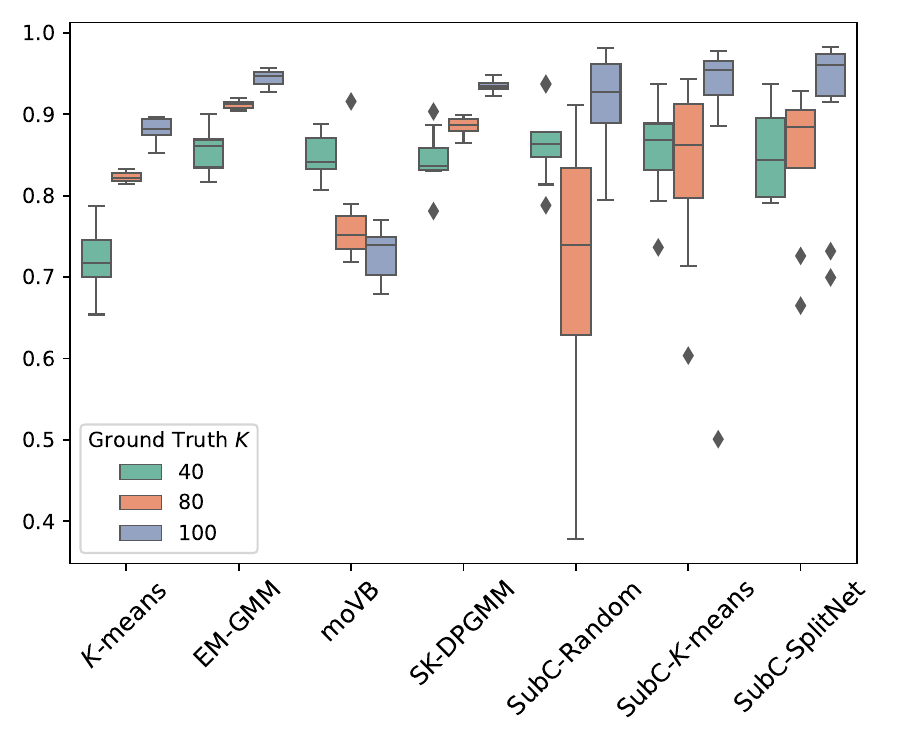}}
  \subcaptionbox{ARI per Method\label{fig:results:5D:boxplot:ARI}}
  {\includegraphics[width=0.32\linewidth]{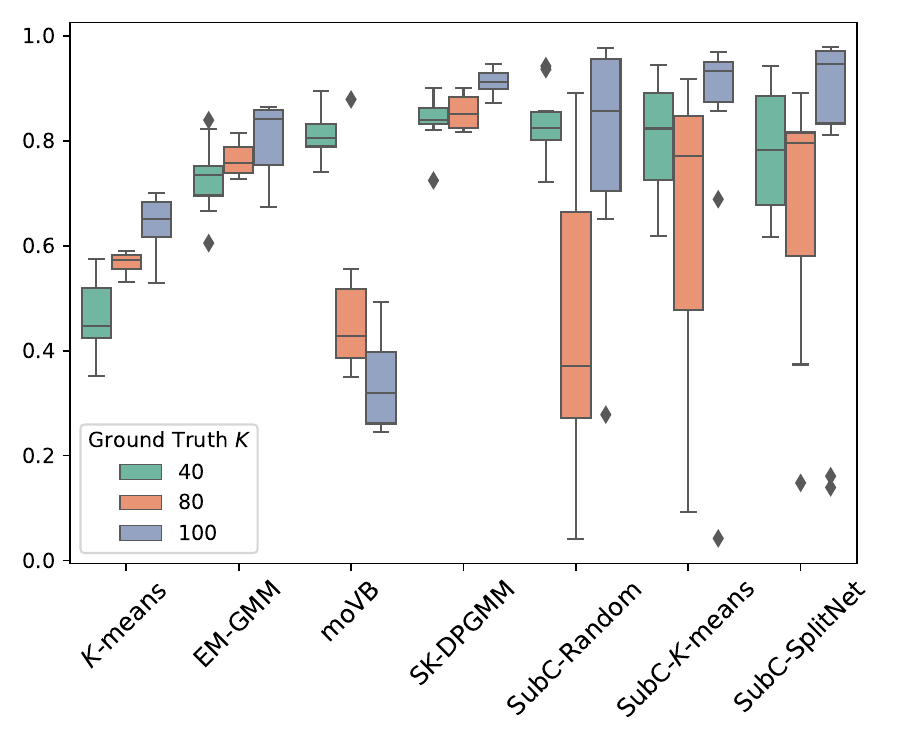}}
  % \subcaptionbox{Time per Method\label{fig:results:5D:boxplot:time}}
  % {\includegraphics[width=0.24\linewidth]{../figures/boxplots/5D_Time.pdf}}  

\caption{Performance on 5D Datasets.}
\label{fig:results:5D:boxplot}
\end{figure*}

\begin{figure*}[h!]
\centering
  \subcaptionbox{Inferred $K$ per Method\label{fig:results:10D:boxplot:K}}
  {\includegraphics[width=0.32\linewidth]{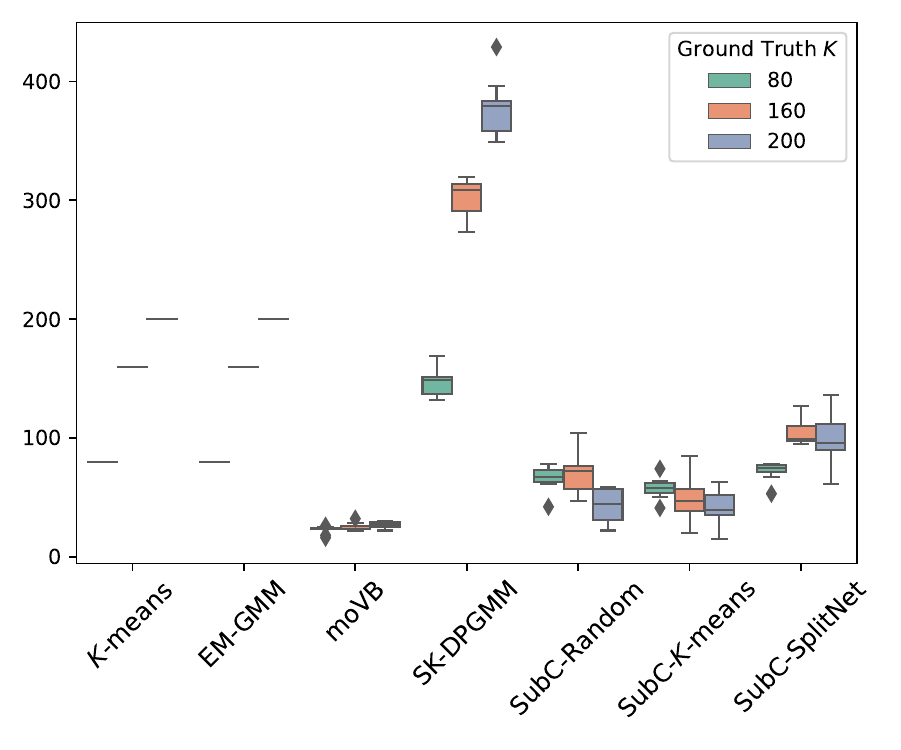}} 
  \subcaptionbox{NMI per Method\label{fig:results:10D:boxplot:NMI}}
  {\includegraphics[width=0.32\linewidth]{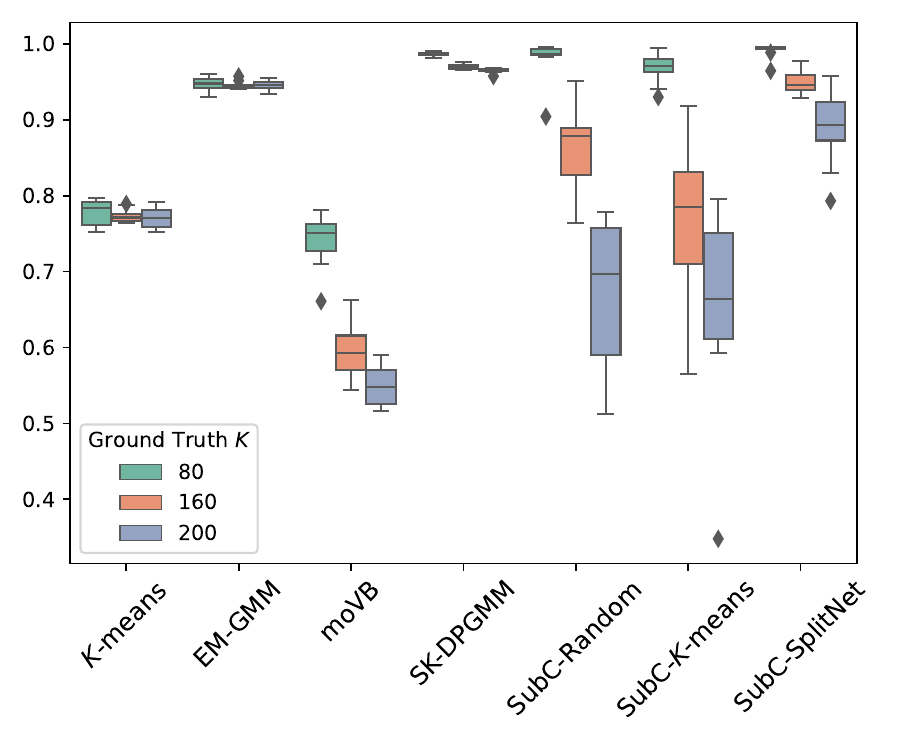}}
  \subcaptionbox{ARI per Method\label{fig:results:10D:boxplot:ARI}}
  {\includegraphics[width=0.32\linewidth]{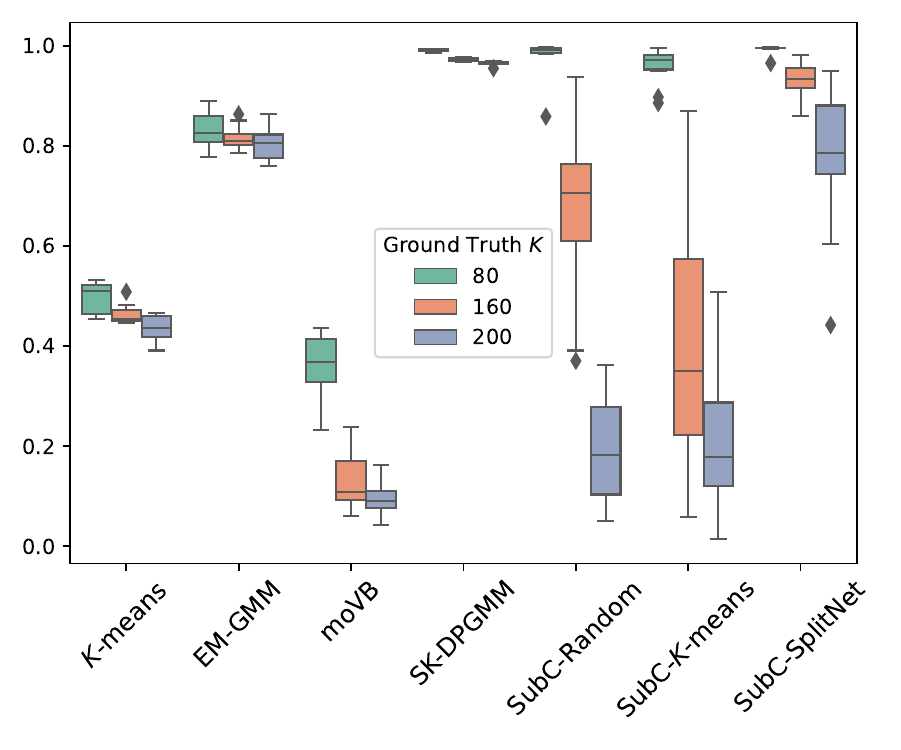}}
  % \subcaptionbox{Time per Method\label{fig:results:10D:boxplot:time}}
  % {\includegraphics[width=0.32\linewidth]{../figures/boxplots/10D_Time.pdf}}  

\caption{Performance on 10D Datasets.}
\label{fig:results:10D:boxplot}
\end{figure*}

Here we present additional GMM test configurations with a greater number of GT clusters and on higher dimensions: $D\in\set{3,5,10}$, and the results are shown in ~\autoref{fig:results:3D:boxplot}, ~\autoref{fig:results:5D:boxplot} and ~\autoref{fig:results:10D:boxplot} respectively. In each setting, we generate 10 different datasets. We observe that the SubC-SplitNet method generally outperforms the rest of the methods. Also, note that the SK-DPGMM method consistently over-estimates the number of GT clusters, which me explain why its NMI and ARI values are high. The performance of the moVB method deteriorates as the number of clusters increases. Again, we stress that the SK-DPGMM method is 2 orders of magnitude slower than the SubC-* methods (\eg, 1500 seconds vs. 15 seconds).

\clearpage

\subsection{Experiments Details}

All the experiments and model training were done on an Ubuntu 16.04 machine with Intel® Xeon(R) CPU E5-2630 @ 2.20GHz Processor, with an NVIDIA Tesla P100 (16 Gb VRAM) GPU.

The following SW language/packages versions were used: Julia: 1.4, Python 3.8, PyTorch 1.9.

\subsection{Details on Dataset used}

For the first two datasets, MNIST and Fashion MNIST, we use the raw features of the images themselves, and project them via PCA to a lower dimension. For the rest of the datasets (CIFAR10, CIFAR20, and STL10), we compute deep features produced by a recent unsupervised method SCAN~\citep{vangansbeke2020scan} to learn and classify images. The deep features are 512-dimensional vectors, so we apply PCA and project the data to a 20-dimensional space.

\textbf{MNIST}

The MNIST dataset~\citep{deng2012mnist} is a dataset of 60,000 small square 28×28 pixel grayscale (784 total pixels) images of handwritten single digits between 0 and 9. We apply PCA on the flattened images and project the test data to $D=20$.

\textbf{Fashion MNIST}

The Fashion-MNIST dataset~\citep{xiao2017fashion} contains 60,000 training images (and 10,000 test images) of fashion and clothing items, taken from 10 classes. Each image is a standardized 28×28 size in grayscale (784 total pixels). We apply PCA on the flattened images and project the test data to $D=20$. Results are reported on the testset.

\textbf{CIFAR10}

The CIFAR-10 dataset~\citep{Krizhevsky:TechReport:2009:cifar} consists of 60000 32x32 color images in 10 classes, with 6000 images per class. There are 50000 training images and 10000 test images. Instead of raw image features, we use the deep features (embeddings) produced by SCAN, which are 512-dimensional. We further reduced them 20 dimensions via PCA. Results are reported on the testset.

\textbf{CIFAR20}

CIFAR100~\citep{Krizhevsky:TechReport:2009:cifar} is an extension of CIFAR10, only with 100 classes. Each image comes with a "fine" label (the class to which it belongs) and a "coarse" label (the superclass to which it belongs). The 100 classes in the CIFAR-100 are grouped into 20 superclasses. Here, we treat the superclasses as the ground truth clusters. Features are processed similarly as in the CIFAR10 case. Results are reported on the test set images.

\textbf{STL10}

STL-10 is an image recognition dataset~\citep{coates2011analysis} consists of 96x96 color images, with a corpus of 100000 unlabeled images, 50000 training images, and 8000 test images. Features are processed similarly as in the CIFAR10 case. Results are reported on the testset.

\clearpage

\section{The SubC Sampler Algorithm}
\autoref{Alg:SubC} below is the full algorithm of the original SubC sampler (\ie, SubC-Random in our terminology), proposed by~\citep{Chang:NIPS:2013:ParallelSamplerDP}, (where $\appropto$ denotes sampling proportional to the right-hand side of the equation).
Note that $H_{\mathrm{merge}}$ from~\autoref{Alg:SubC} here was called $H$ in our paper. 
The rest of the notation is as in our paper.

\begin{algorithm}[h!]
    \begin{algorithmic}[1]
        \small
        \Procedure{Restricted Gibbs sampling}{$X$}
        \State Sample cluster weights $\pi_1,\pi_2,\dots,\pi_K,$:
                \begin{align}
                    (\pi_1,\dots,\pi_K,\tilde \pi_{K+1}) \sim \text{Dir}(N_1, \dots, N_K, \alpha).
                    \label{eqn:sample_weights}
                \end{align}
        \State Sample subcluster weights $\bar \pi_{k,l}, \bar \pi_{k,r}$ for each 
        cluster $k\in\{1,\dots,K\}$:
            \begin{align}
            (\bar \pi_{k,l}, \bar \pi_{k,r}) \sim \text{Dir}(N_{k,l} + \alpha/2, N_{k,r} + \alpha/2).
            \label{eqn:sample_sub_weights}
            \end{align}
        \State Sample cluster parameters $\theta_k$ for each cluster $k$:
            \begin{align}
            \theta_k \appropto f_\bx(\bx_{\Ical_k};\theta_k) f_\theta(\theta_k;\lambda)
            \label{eqn:sample_params}
            \end{align}
        \State Sample subcluster parameters $\bar \theta_{k,h}$ for each cluster $k\in\{1,\dots,K\}$ and $h\in\{l,r\}$:
            \begin{align}
            \bar\theta_{k,h} \appropto f_\bx(\bx_{\Ical_{k,h}};\bar\theta_{k,h}) 
            f_\theta(\bar\theta_{k,h};\lambda).
            \label{eqn:sample_sub_params}
            \end{align}
        \State Sample cluster assignments $z_i$ for each point $i\in \{1,\dots,N\}$:
            \begin{align}
            z_i \appropto \sum\nolimits_{k=1}^{K} \pi_k  f_\bx(\bx_i;\theta_{k}) \bone_{z_i=k}.
            \label{eqn:sample_labels}
            \end{align}
        \State Sample subcluster assignments $\bar z_i$ for each point $i\in \{1,\dots,N\}$:
            \begin{align}
            \bar z_i \appropto \sum\nolimits_{h\in\{l,r\}} \pi_{z_{i,h}} f_\bx(\bx_i;\bar\theta_{z_{i,h}}) 
            \bone_{\bar z_i=h}.
            \label{eqn:sample_sub_labels}
            \end{align}
        \EndProcedure
        
        \Procedure{Propose and Accept Splits}$(X)$
        \State \label{alg:chang:split} Propose to \textbf{randomly} split cluster $k$ into its 2 subclusters for all $k\in \{1,2,\dots, K\}$. 
        \State Calculate the Hastings ratio $H_{\text{split}}$ and accept the split with probability $\text{min}(1,H_{\text{split}})$
        where: 
                \begin{align}
            \hspace{-.2cm}        H_{\text{split}} = \frac{\alpha \Gamma(N_{k,l}) f_\bx(\bx_{\Ical_{k,l}};\lambda)  \Gamma(N_{k,r}) f_\bx(\bx_{\Ical_{k,r}};\lambda)}{\Gamma(N_k) f_\bx(\bx_{\Ical_k};\lambda)} 
                     \label{eqn:HastingRatioSplit}
                \end{align}
        \EndProcedure
            
        \Procedure{Propose and Accept Merges}$(X)$
        \State Propose to merge clusters $k_1, k_2$ for all pairs $k_1, k_2 \in \{1,2,\dots, K\}$.
        \State Calculate the Hastings ratio $H_{\text{merge}}$ and accept the merge with probability $\text{min}(1,H_{\text{merge}})$ where 
             \begin{align}\hspace{-.2cm} 
\hspace{-.2cm} H_{\text{merge}}  = \frac{\Gamma(N_{k_1}+N_{k_2}) f_\bx(\bx_{\Ical_{k_1}\cup \Ical_{k_2} };\lambda)} 
{\alpha \Gamma(N_{k_1}) f_\bx(\bx_{\Ical_{k_1}};\lambda)  \Gamma(N_{k_2}) f_\bx(\bx_{\Ical_{k_2}};\lambda)}
                 \label{eqn:sample_merges}
             \end{align}
        \EndProcedure
    \end{algorithmic}
        \caption{The SubC Sampler~\citep{Chang:NIPS:2013:ParallelSamplerDP}}
            \label{Alg:SubC}
\end{algorithm}

\pagebreak

\section{Posteriors and Marginal Data Likelihood in a Gaussian Model}
Here we provide the expressions, in a Gaussian model with an NIW prior, for the posterior hyperparameters and the marginal data likelihood. For more details
see~\citep{Gelman:Book:2013:Bayesian} or~\citep{Chang:Thesis:2014sampling}.

In what follows, we assume we have $m$ samples, each of dimension $D$, from a multivariate normal distribution: 
\begin{align}
    \bx_i | \bmu,\bSigma \sim \Ncal(\bmu,\bSigma)
\end{align}    
where $\bx$, an $m \times d$ matrix, collects all the data such that $\bx_i$ is $i$'th row of $\bx$. 
\subsection{Posterior Distribution of the Parameters}

When the mean and covariance matrix of the sampling distribution are unknown, one can place a Normal-Inverse-Wishart prior on the mean and covariance parameters jointly: $(\bmu,\bSigma)\sim \mathrm{NIW} (\bmu_{0},\kappa , \bPsi ,\nu)$ (where $(\bmu_{0},\kappa , \bPsi ,\nu)$, hyperparameters associated
with the NIW distribution, were collectively denoted by $\lambda$ in our paper). 

By conjugacy, the resulting posterior distribution for the mean and covariance matrix will also be a Normal-Inverse-Wishart distribution with closed-form updates:  
$(\bmu,\bSigma| \bx)\sim \mathrm{NIW} (\bmu_{m},\kappa _{m},\bPsi_{m},\nu _{m})$:
The posterior NIW four parameters are updated as follows:

\begin{align}
    & \bmu_{m} = \frac{\kappa \bmu_0 + m{\bar{ \bx }}}{\kappa + m} \\
    & \kappa_{m}=\kappa + m  \\
    & \nu_{m} = \nu + m  \\
    & \bPsi_{m}= \bPsi +S + \frac{\kappa m}{\kappa +m} (\bm{{\bar{\bx}} - \mu_0})^{T} (\bm{\bar{\bx}}-\mu_0) \\ 
    & \mathrm{with,} ~~{\b{S}}= \sum_{i=1}^{m} (\bm{\bx_{i}-\bar {\bx}})^{T}(\bm{\bx_i}-\bar{\bx}) 
\end{align}\label{Eqn:HR_posterior_update}

\subsection{The Marginal Data Likelihood}
When marginalizing over the parameters of a Gaussian (\ie, its mean and covariance),
one obtains the marginal data likelihood (given the hyperparameters of the NIW prior): 
\begin{align}
    &  f_\bx(\bx;\lambda)=f_\bx(\bx;\bmu_{0},\kappa , \bPsi ,\nu) =\frac{1}{\pi^{\frac{m d}{2}}} \frac{\Gamma_{D}\left(\nu_{m} / 2\right)}{\Gamma_{D}(\nu / 2)} \frac{|\nu \bPsi|^{\nu / 2}}{\left|\nu_{m} \bPsi_{m}\right|^{\nu_{m} / 2}}\left(\frac{\kappa}{\kappa_{m}}\right)^{d / 2} \label{Eqn:LL_post} \, 
\end{align}    
where $\Gamma_d$ is the $D$-dimensional Gamma function.

\clearpage

\section{Model Training Details}

% We trained the models on the MACHINE DETAILS with a GPU MODEL. 
The models, training code and data-generation related code, are all based on the PyTorch~\citep{NEURIPS2019_9015}, NumPy~\citep{2020NumPy-Array}, SciPy~\citep{2020SciPy-NMeth}, and SciKit-Learn~\citep{scikit-learn} Python packages. 

To build and compare the SplitNet models across a variety of scenarios, we trained three SplitNet models, for dimensions $D=2,10,20$ .

\subsection{Nuances for successful learning}

Here are a few important design choices to aid with the training of SplitNet:

\textbf{Input Normalization:} 

Each dataset $X=[x_1, \cdots, x_n]^T$ (during training and inference) is normalized by subtracting the mean and dividing by the standard deviation: $\hat{X} = \frac{X - \mathrm{mean}(X)}{std(X)}$. 
% \begin{align*}
    % \hat{X} = \frac{X - \mathrm{mean}(X)}{std(X)}
% \end{align*}
This standardizes the input space in terms of range, and helps the model to learn faster.

\textbf{Curriculum Learning:}

As we mentioned earlier, we can control the training dataset difficulty level. 
% While experimenting with different datasets compositions for training, we have found empirically that by including difficult examples too early in the training process, the model could not learn properly on them. So a solution to that is a simple version of curriculum learning - 
We have found that gradually increasing the dataset difficulty level during training, helps the training speed. Specifically, we choose an initial ``easy'' NIW prior (for data generation) for the beginning of the training, a final ``difficult'' NIW prior, and the update frequency $T$ (number of epochs). The choices for these priors are listed in \autoref{tab:training-settings}.

Then, we (linearly) interpolate the NIW prior between the initial and final values (in $\frac{\text{Total Epochs}}{T}$ points), generate a more difficult data subset with those interpolated values, and replacing some existing data. So that at the end of the training, the entire dataset consists of varying-difficulty clusters, from easy to hard, and the model can perform well on all of them. \autoref{fig:data_samples_sup} shows concrete examples of different levels of difficulty.

\textbf{Ensuring ``Splittable'' Clusters:}

It is essential to ensure that the generated dataset is ``splittable''\wrt the Hastings Ratio - \ie, its log Hastings Ratio is higher than 1. So any generated datasets with a low Hastings' Ratio are discarded to avoid training on ``noisy'' samples, which can only hinder the model's performance.

\subsection{Training Hyperparameters}

The training details, data-related and model's hyper-parameters are summarized in~\autoref{tab:training-settings}. 

\begin{table*}[h!]
    \centering
    \caption{Training Settings for SplitNet}
    \small
        \begin{tabular}{ccccc}
            \toprule
        \multirow{2}{*}{}                                                                    & {\textbf{Parameter}}     & \multicolumn{3}{c}{\textbf{Value}}                                   \\ \midrule
                                                                                            &                                         & \textbf{$D=2$ Model} & \textbf{$D=10$ Model} & \textbf{$D=20$ Model} \\ \midrule
        \multirow{3}{*}{\rotatebox{90}{\textbf{General}}}   & Optimizer                               & Adam                 & Adam                  & Adam                  \\
                                                            & Learning Rate                           & 0.01                 & 0.01                  & 0.01                  \\
                                                            & Total Epochs                            & 200                  & 300                   & 400                   \\ 
                                                            & Batch Size                              & 64                  & 64                    & 32                        \\ \midrule
        \multirow{6}{*}{\rotatebox{90}{\textbf{Model}}}     & Hidden Layer Size                      & 128                  & 128                   & 256                   \\
                                                            & \# of Encoders (ISAB)                   & 2                    & 2                     & 3                     \\
                                                            & \# of Decoders (PMA)                    & 2                    & 2                     & 3                     \\
                                                            & \# of Inducing Seeds                    & 64                   & 64                    & 128                   \\
                                                            & \# of PMA Seeds                         & 8                    & 16                    & 16                    \\ 
                                                            & \# of Heads                             & 4                    & 8                     & 8                     \\ \midrule
        \multirow{6}{*}{\rotatebox{90}{\textbf{Data}}}      & Size of Dataset (train)                 & 10000                & 10000                 & 10000                 \\
                                                            & Size of Dataset (validation)            & 1000                 & 1000                  & 1000                  \\
                                                            & NIW Prior  - Initial ($\bmu_0, \bPsi, \nu, \kappa$)             & $\bzero_{2\times 1}, \bI_{2\times 2}$, 10, 0.1  & $\bzero_{10\times 1}, \bI_{10\times 10}$, 20, 0.1 &  $\bzero_{20\times 1}, \bI_{20\times 20}$, 25, 0.1 \\
                                                            & NIW Prior  - Final ($\bmu_0, \bPsi, \nu, \kappa$)               & $\bzero_{2\times 1}, \bI_{2\times 2}$, 4, 2     & $\bzero_{10\times 1}, \bI_{10\times 10}$, 11, 5   &  $\bzero_{20\times 20}, \bI_{20\times 20}$, 21, 2.5 \\
                                                            & NIW Prior Update Frequency, $T$ (\# of epochs)           & 20                 & 30                    & 40     \\              
        % &
        % \begin{tabular}[c]{@{}c@{}} NIW Prior  - Initial Values\\ u\_0, Psi\_, nu, kappa \end{tabular}`' &  0, 1, 10, 0.1 &  , 20, 0.1 &  , 25, 0.1 \\
        % &
        % \begin{tabular}[c]{@{}c@{}} NIW Prior  - Final Values\\ mu_0, Psi\_, nu, kappa \end{tabular} &  , 4, 2 & , 11, 5 &  , 21, 2.5 \\ 
            % &
\bottomrule
\end{tabular}
\label{tab:training-settings}
\end{table*}

\pagebreak

\section{Training Data Details}

For deep learning in general, the data the model is trained on is one of the most critical factors for good model performance. Therefore, it must be of high quality and provided with enough variability so that the model would be able to generalize well on unseen and challenging datasets.

Fortunately, in the case of the DPGMM, we assume the data is a GMM, and it is relatively simple to define a generative process for producing an GMM with much control over various parameters such as the Gaussians' proximity to each other, the variance of their covariance matrices, the ratio of the number of points between the Gaussians, the number of clusters, how overlapping they are, and more. The generative algorithm we used is described in~\autoref{alg:data_sample} below. To facilitate the creation of a rich dataset space, we used the NIW distribution for sampling the Gaussian parameters. 

\begin{algorithm}[h]
% \LinesNumbered
% \SetAlgoLined
% \setstretch{1.2}

% \KwIn{$\alpha_{Dir}, N_{max}, \nu, \kappa, \Psi, \bmu_0 $}
$p_1, p_2 \sim \mathrm{Dir}([\alpha_{\mathrm{Dir}}, \alpha_{\mathrm{Dir}}])$

$n_1, n_2 =\lceil p_1 \cdot N_{max} \rceil , \lceil p_2 \cdot N_{max} \rceil$

$\bmu_1, \bSigma_1 \sim NIW( \nu, \kappa, \Psi, \bmu_0)$

$\bmu_2, \bSigma_2 \sim NIW( \nu, \kappa, \Psi, \bmu_0)$

$X_l \sim \mathcal{N}(\bmu_1, \bSigma_1)$ - sample $n_1$ points.

$X_r \sim \mathcal{N}(\bmu_2, \bSigma_2)$ - sample $n_2$ points.

$X \longleftarrow [X_l, X_r]$    

$Z \longleftarrow [\bzero_{n_1},  \bone_{n_2} ]$

% \KwOut{X, Z}
\caption{Training Data Generation Process}
\label{alg:data_sample}
\end{algorithm}

This is also where the model can learn ``difficult'' splits, where other methods (e.g., $K$-means splits) are most likely to fail. We generate a dataset with specific characteristics: clusters imbalance, non-spherical covariances, and overlapping clusters. The ``knobs'' with which we tune this are the following:

\begin{enumerate}
    \item $\kappa$ (one of the NIW hyperparameters) - the larger $\kappa$ is, the more separable the clusters are.
    \item $\nu$ (another NIW hyperparameter) - the smaller $\nu$ is, the more isotropic the cluster are.
    \item $\alpha_{\mathrm{Dir}}$ (a hyperparameter of the two-dimensional Dirichlet distribution) - the smaller $\alpha$ is, the more imbalanced the points allocation between the subclusters.
\end{enumerate}

\begin{figure}[h!]
\centering
  \subcaptionbox{Easy Data Samples \label{fig:data_samples:easy}}
  {\includegraphics[width=0.8\linewidth]{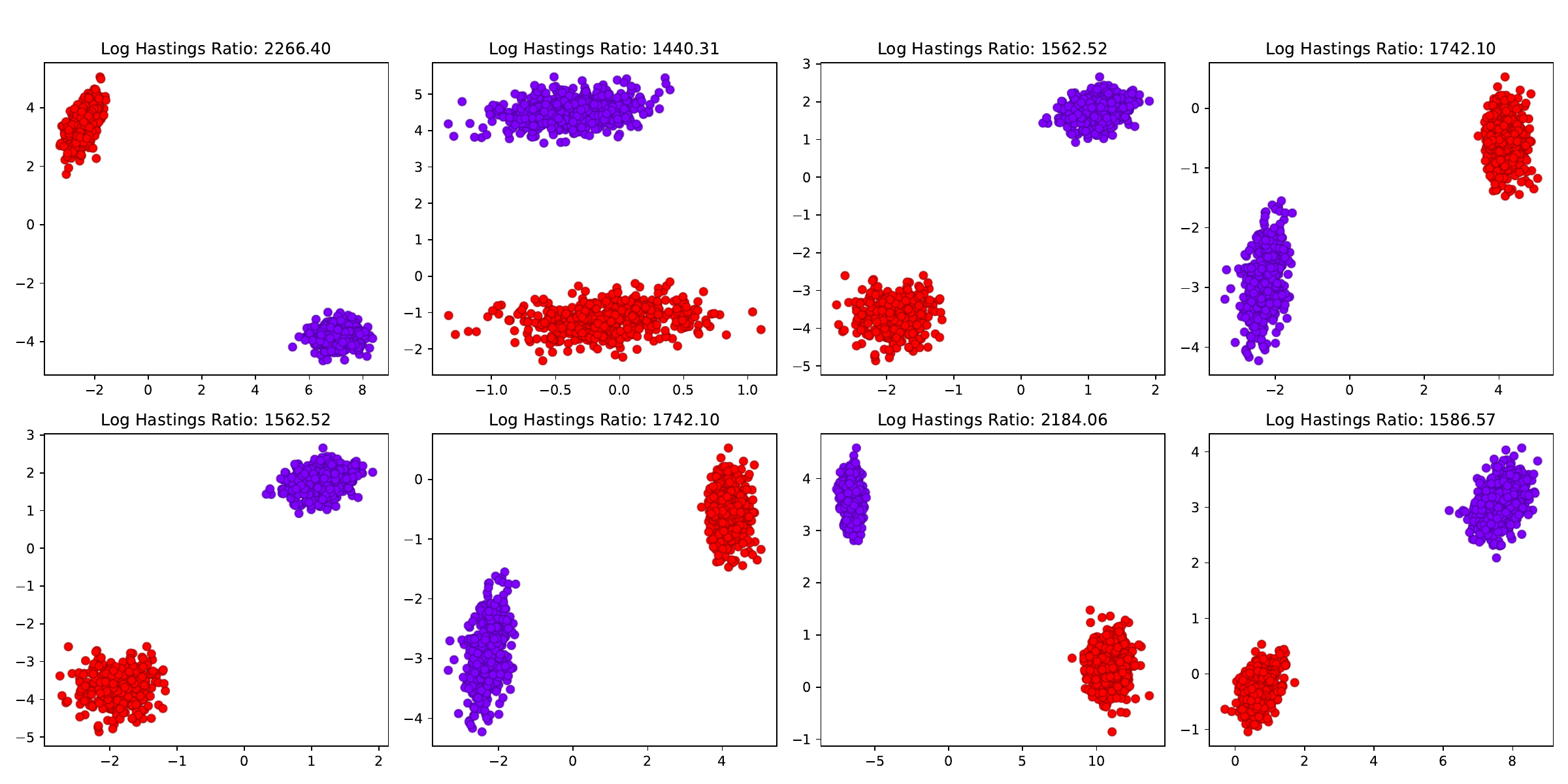}} 
  
  \subcaptionbox{Medium Data Samples \label{fig:data_samples:medium}}
  {\includegraphics[width=0.8\linewidth]{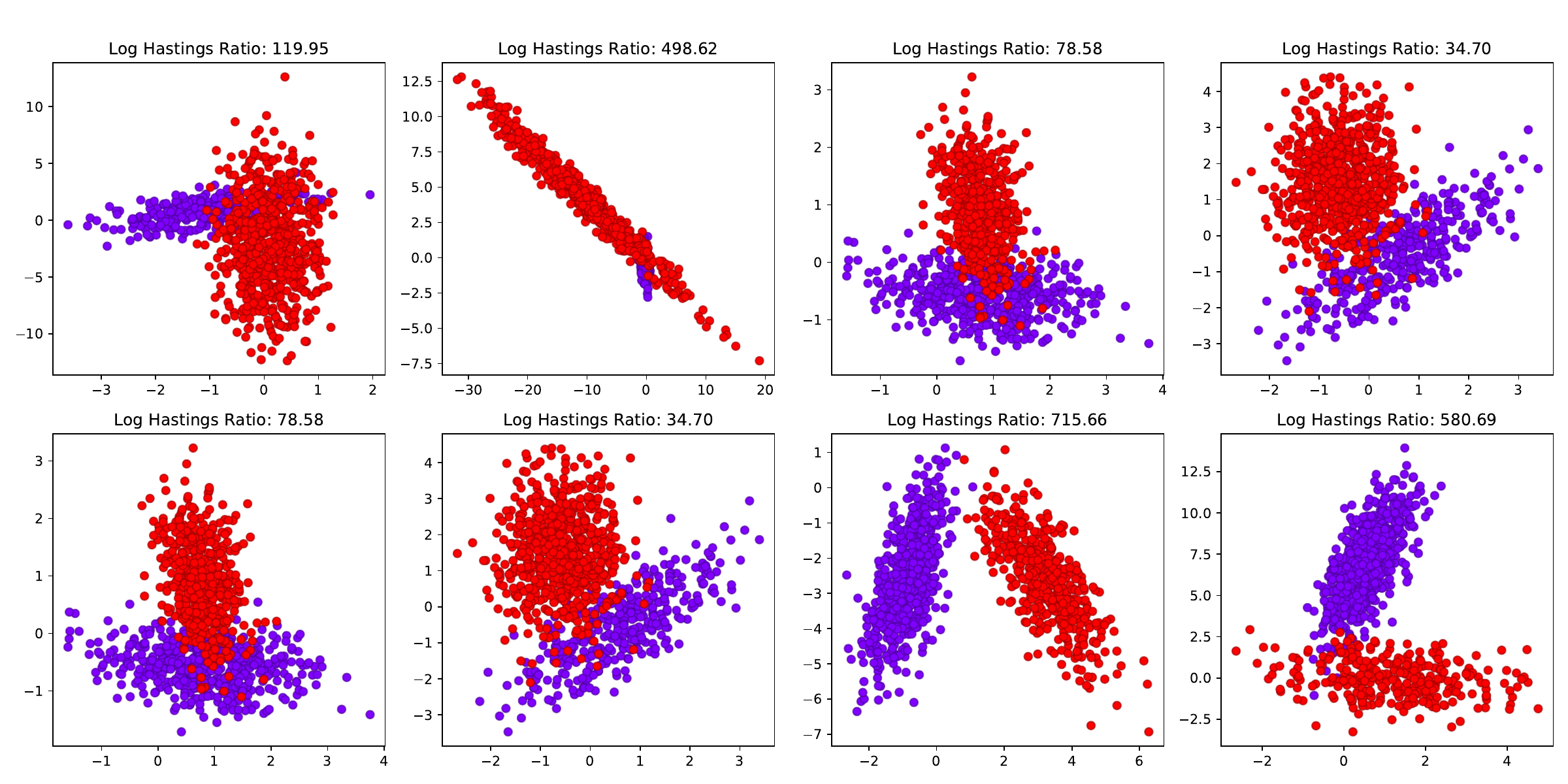}}
  
  \subcaptionbox{Hard Data Samples \label{fig:data_samples:hard}}
  {\includegraphics[width=0.8\linewidth]{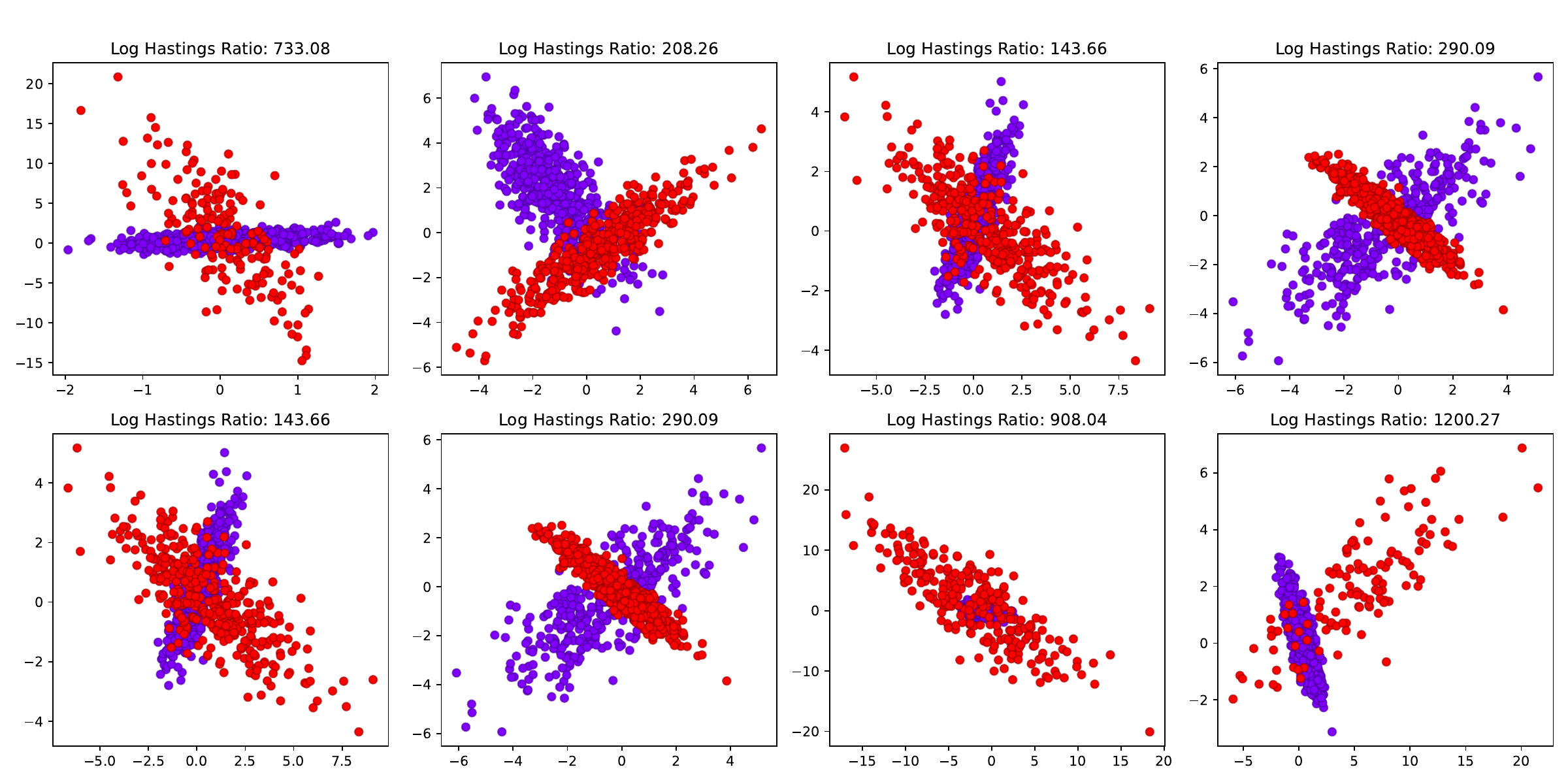}}
  
\captionsetup{justification=justified, singlelinecheck=false}
\caption{Examples of Generated Training Data of Various difficulty levels} %. First row: "easy" data, middle row: "medium" data, last row: "hard" data.}
\label{fig:data_samples_sup}
\end{figure}

Examples of samples of varying difficulty datasets are depicted in \autoref{fig:data_samples_sup}.

% Intuitively, what we are trying to teach the model is to identify and separate two distinct regions of high density. 
% We have also experimented with generating and training data with multiple clusters ($K>2$), but it didn't improve performance.

\clearpage

\clearpage

\section{The Set Transformer}

The Set Transformer (ST)~\citep{lee2019set} is a permutation-invariant set-input neural network that uses self-attention operations as building blocks. It utilizes multi-head attention~\citep{vaswani2017attention} for both encoding elements of a set and decoding encoded features into outputs.

The attention mechanism allows processing every element in an input set, which enables the ST to naturally encode pairwise or higher-order interactions between elements in the set.

Assume we have $n$ query vectors (corresponding to a set with $n$ elements) each with dimension $d_{q}: Q \in \RR^{n \times d_{q}}$. 
An attention function $\mathrm{Att}(Q, K, V)$ is a function that maps queries $Q$ to outputs using $n_{v}$ key-value pairs $K \in$ $\RR^{n_{v} \times d_{q}}, V \in \RR^{n_{v} \times d_{v}}$.

\begin{align}
    \mathrm{Att}(Q, K, V ; \omega)=\omega\left(Q K^{\top}\right) V
\end{align}

The pairwise dot product $QK^T \in \RR^{n \times n_v}$ measures how similar each pair of query and key vectors is, with weights computed with an activation function $\omega$. The output
$\omega(QK^T)V$ is a weighted sum of $V$ where a value gets more weight if its corresponding key has larger dot product with the query.

Multihead Attention is an extension of this mechanism, originally introduced by~\citep{vaswani2017attention}, in which $Q, K ,V$ are first projected into $h$ different $d_q^M, d_k^M, d_v^M$-dimensional vectors, respectively. Then the attention function is applied individually on these projections. The output is computed by concatenating the attention outputs and applying linear transformation on it:

\begin{align}
    &\mathrm{Multihead-Att}(Q, K, V ; \lambda, \omega)=\mathrm{concat}\left(O_{1}, \cdots, O_{h}\right) W^{O}\\
    &\text { where } O_{j}=\mathrm{Att}\left(Q W_{j}^{Q}, K W_{j}^{K}, V W_{j}^{V} ; \omega_{j}\right)
\end{align}

\begin{figure}[h!]
    \centering
      \subcaptionbox{MAB Block \label{fig:ST_arch:mab}}
      [.48\linewidth]{\includegraphics[width=0.45\textwidth]{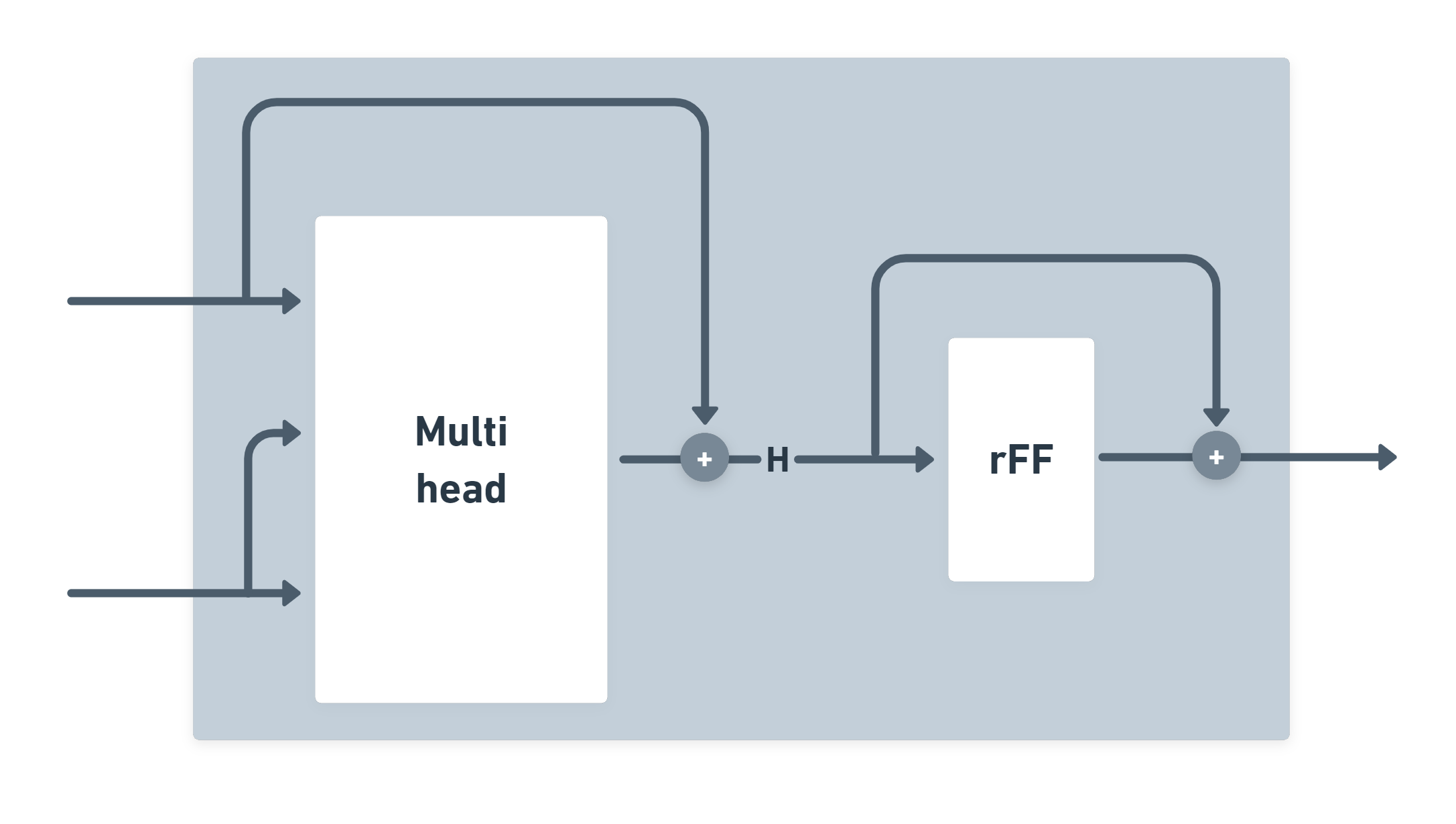}} 
      \subcaptionbox{SAB Block \label{fig:ST_arch:sab}}
      [.48\linewidth]{\includegraphics[width=0.45\textwidth]{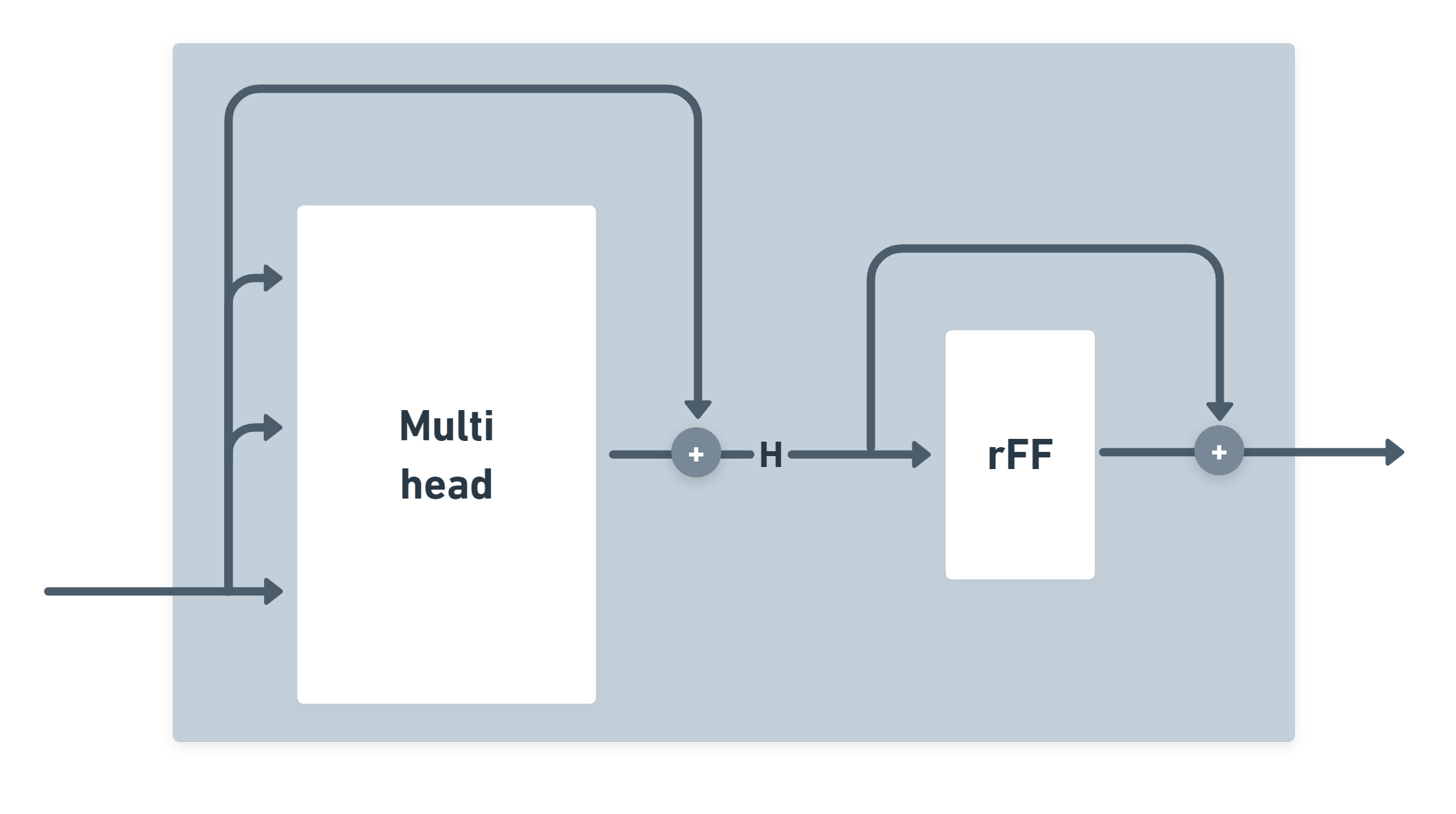}}
      
      \subcaptionbox{ISAB Block \label{fig:ST_arch:isab}}
      [.48\linewidth]{\includegraphics[width=0.45\textwidth]{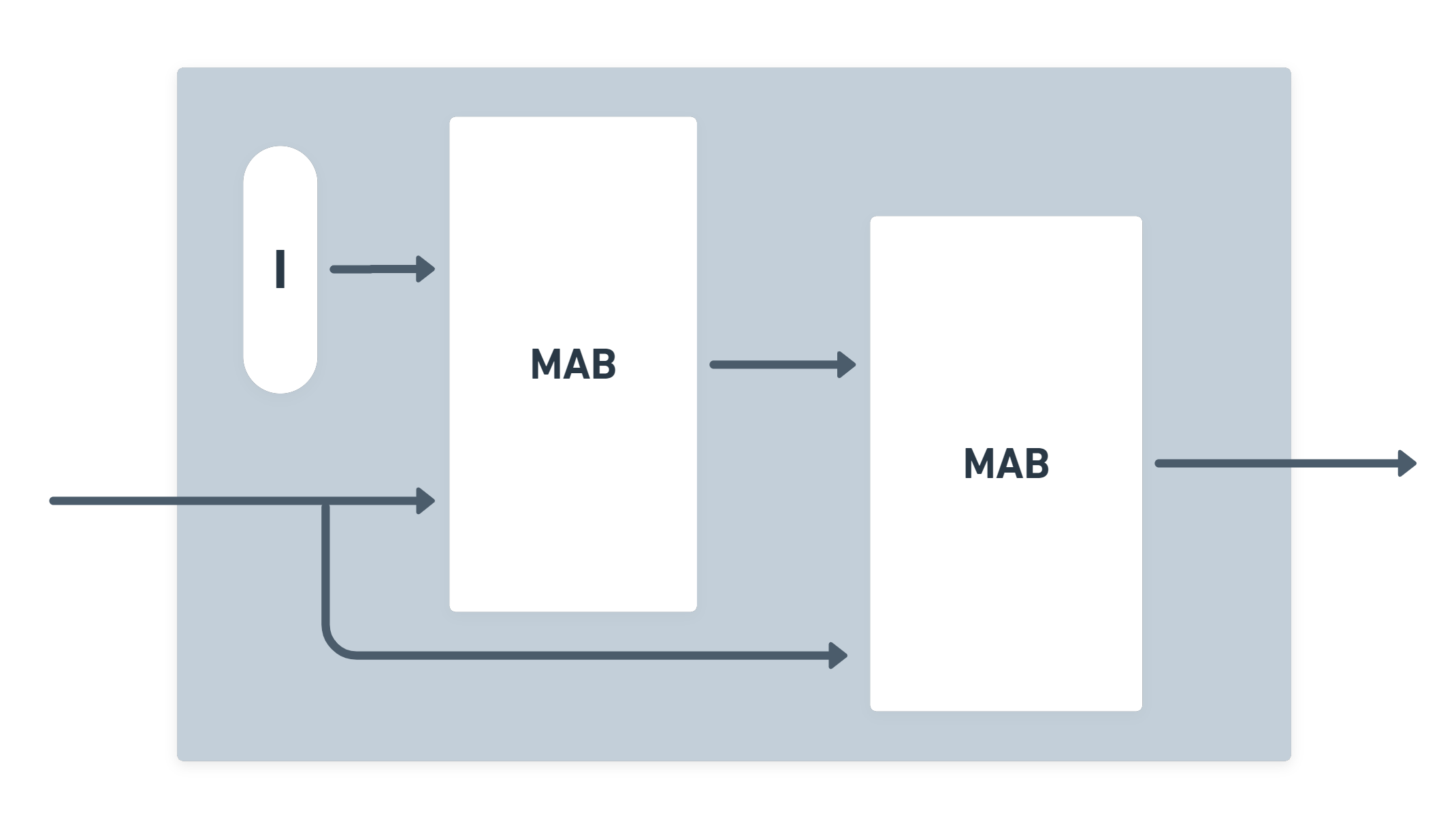}}
    \subcaptionbox{PMA Block \label{fig:ST_arch:pma}}
      [.48\linewidth]{\includegraphics[width=0.45\textwidth]{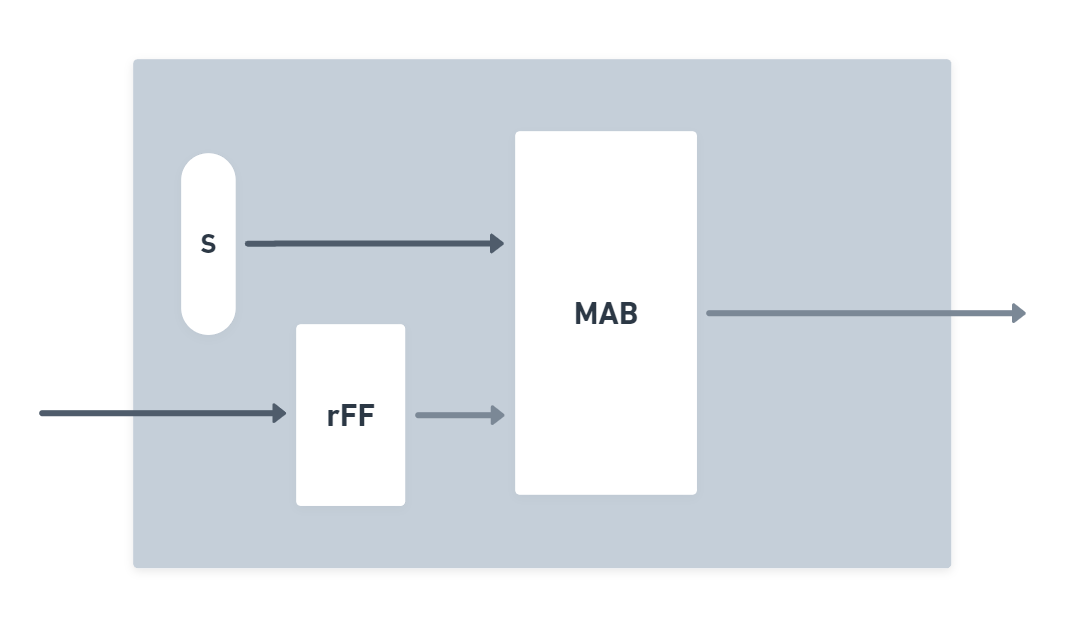}}
      
    \caption{Basic Attention-based Blocks of the Set Transformer}
    \label{fig:ST_arch}
\end{figure}  

The fundamental building block of a ST is the Multihead Attention Block (MAB), which takes two sets $X = [x_1, \ldots , x_n]$ and $Y = [y_1, ... , y_m]^T$ and outputs a set of the same size as $X$.  An MAB is defined as
\begin{align}
    \mathrm{MAB}(X, Y)=H+\mathrm{rFF}(H)
\end{align}

where $H=X+\mathrm{rFF}($ Multihead-$\mathrm{Att}(X, Y))$ and rFF(·) is a feed-forward layer applied row-wise (\ie, for each element). $\mathrm{MAB}(X, Y)$ computes the pairwise interactions between the elements in X and Y with sparse weights obtained from attention. 

A Self-Attention Block (SAB) is simply MAB applied to the set itself: 
\begin{align}
    \mathrm{SAB}(X) = \mathrm{MAB}(X, X).
\end{align}

We can model high-order interactions among the items in a set by stacking multiple SABs.
Note that the time-complexity of SAB is $O(n^2)$ because of pairwise computation. To reduce this, the authors~\citep{lee2019set} proposed to use Induced Self-Attention Block (ISAB) defined as:
\begin{align}
    \mathrm{ISAB}(X)=\mathrm{MAB}(X,\mathrm{MAB}(I, X))
\end{align}
where $I = [i_1, \ldots , i_m]^T$ are trainable inducing points. ISAB indirectly compares the elements of X through the inducing points, reducing the time-complexity to $O(nm)$.

To summarize a set into a fixed-length representation, ST uses an operation called Pooling by Multihead Attention (PMA). A PMA is defined as 
\begin{align}
    \mathrm{PMA}_k(X) = \mathrm{MAB}(S, X)
\end{align}
where $S = [s_1, \ldots , s_k]^T$ are trainable parameters.

A visual representation of these basic blocks is available in \autoref{fig:ST_arch}.

% \clearpage
% \small
% % % % % \bibliographystyle{ieee}
% % % % %\bibliographystyle{ieee_fullname}
% \bibliographystyle{abbrvnat}
% \bibliography{../refs}

\end{document}